\documentclass{IEEEtran}

\usepackage[T1]{fontenc}
\usepackage{lmodern}
\usepackage{newtxtext,newtxmath}
\usepackage{subfigure}
\usepackage{enumitem}
\usepackage{xspace}
\usepackage{color}
\usepackage{graphicx}
\usepackage{textcomp,gensymb}
\usepackage{multirow}
\usepackage{diagbox}
\usepackage{array}
\usepackage{algorithm}
\usepackage{algpseudocode}
\usepackage{url}
\usepackage{hyperref} 

\newcommand{\ALERT}{{\small \sf \mbox{\textit{ALERT}}}\xspace}
\newcommand{\mine}{{\small \sf \mbox{\textit{ISA-ViT}}}\xspace}
\newcommand{\rev}[1]{\textcolor{black}{#1}}
\newcommand{\re}[1]{\textcolor{black}{#1}}

\begin{document}

\title{ALERT Open Dataset and Input-Size-Agnostic Vision Transformer for Driver Activity Recognition using IR-UWB}
\author{
\IEEEauthorblockN{
Jeongjun Park\IEEEauthorrefmark{1}\IEEEauthorrefmark{4},
Sunwook Hwang\IEEEauthorrefmark{1}\IEEEauthorrefmark{4},
Hyeonho Noh\IEEEauthorrefmark{2}\IEEEauthorrefmark{4},
Jin Mo Yang\IEEEauthorrefmark{3},
Hyun Jong Yang\IEEEauthorrefmark{3},
Saewoong Bahk\IEEEauthorrefmark{3}\IEEEauthorrefmark{5}
}\\
\vspace{10pt}
\IEEEauthorblockA{\IEEEauthorrefmark{1}
\footnotesize
System LSI, Samsung Electronics, South Korea}\\
\IEEEauthorblockA{\IEEEauthorrefmark{2}
Department of ICE, Hanbat National University, Daejeon, Republic of Korea}\\
\IEEEauthorblockA{\IEEEauthorrefmark{3}
Department of ECE and INMC, Seoul National University, Seoul, Republic of Korea}\\
\IEEEauthorblockA{\IEEEauthorrefmark{4}
This work was done when the author was with Department of ECE and INMC, Seoul National University}\\
\IEEEauthorblockA{\IEEEauthorrefmark{5}
Senior Member, IEEE}
\thanks{This paper has been accepted for publication in IEEE Access.}
\thanks{
Corresponding authors: Sunwook Hwang (sunwookh@snu.ac.kr) and
Saewoong Bahk (sbahk@snu.ac.kr).
}
}

\maketitle
\begin{abstract}
Distracted driving contributes to fatal crashes worldwide. To address this, researchers are using Driver Activity Recognition~(DAR) with Impulse Radio Ultra-Wideband~(IR-UWB) radar, which offers advantages like interference resistance, low-power use, and privacy. However, two challenges limit its adoption: the lack of large-scale, real-world UWB datasets on diverse distracted driving behaviors, and the difficulty in adapting fixed-input Vision Transformers~(ViTs) to UWB radar data with non-standard dimensions.

This work tackles both challenges. We present the \ALERT dataset, containing 10,220 radar samples of seven distracted driving activities in real driving conditions. We also propose the Input-Size-Agnostic Vision Transformer~(\mine), a framework designed for radar-based DAR. \mine resizes UWB data to fit ViT input requirements while preserving radar-specific information like Doppler shifts and phase data. By adjusting patches and using pre-trained positional embedding vectors (PEVs), \mine avoids the limitations of simple resizing. Additionally, a domain fusion strategy combines range and frequency domain features, enhancing classification accuracy.

Comprehensive experiments demonstrate that \mine achieves a 22.68\% higher accuracy compared to the existing ViT method in UWB-based DAR. By publicly releasing the \ALERT dataset and detailing our input-size-agnostic strategy, this work paves the way for more robust and scalable distracted driving detection systems in real-world scenarios.
\end{abstract}


\section{Introduction}\label{sec:intro}
According to the World Health Organization (WHO), 1.19 million people die every year due to vehicle accidents~\cite{who}. Additionally, the National Highway Traffic Safety Administration (NHTSA)~\cite{NHTSA} in the United States estimates that approximately 6 million traffic accidents are reported to police each year, resulting in 38,000 to 40,000 fatalities annually. Among these vehicle accidents, distracted driving accounts for about 10\% of fatal incidents, often involving some form of driver distraction. Given its unavoidable nature, detecting and mitigating distracted driving is a crucial task.

To prevent driver distraction, many researchers have studied Driver Activity Recognition~(DAR) using various approaches: camera vision~\cite{ohn2014head, kavi2016multiview}, acoustic signals~\cite{xie2021hearsmoking, xie2019d}, and Radio Frequency~(RF) signals~\cite{wifiheadtracking, jia2018wifind, bai2019widrive, bai2020carin, RaDA, blinkradar, wificamera}. However, these approaches have several limitations.
Camera vision struggles with poor lighting and raises privacy concerns regarding the potential restoration in intelligent vehicle systems~\cite{Hwang_2023_ICCV, concretizer}.
Acoustic signal-based methods can be disrupted by ambient noise and also pose privacy risks due to continuous microphone activation. 
RF signal-based approaches have predominantly been studied using WiFi. However, utilizing WiFi, primarily operating within the busy 2.4 GHz industrial, scientific, and medical (ISM) band, within a vehicle can potentially cause interference with other vehicle devices using Bluetooth or similar busy ISM band communications. 

In this paper, we exploit the Impulse Radio Ultra-Wideband (IR-UWB) radar for DAR while driving to tackle these problems. UWB offers several advantages: i) robustness against interference and multipath fading due to its wide bandwidth ($\geq$ 500 MHz), ii) operation within the 6.5--9 GHz range with low transmit power (-41.3 dBm/MHz), ensuring coexistence with popular wireless formats such as satellite navigation, WiFi, Bluetooth, and ZigBee~\cite{fira}. iii) inherent privacy protection, as it avoids capturing visual or audio data. 
\re{Owing to these characteristics, UWB sensing has been widely adopted for human activity recognition in indoor environments, where stable and robust sensing performance has been consistently reported~\cite{Alarifi2016UWB}.}
Consequently, UWB-based Human Activity Recognition~(HAR) has garnered significant attention~\cite{HARSANET, HAR_2017, SleepPoseNet, RaDA, blinkradar, Uhead, han2020ir, maitre2021recognizing, noori2021ultra, maitre2020fall, vit1,vit2}.

\textbf{Challenges.} 
\re{Despite the potential of UWB, applying it to DAR involves two key challenges.
The first is the lack of comprehensive studies that address diverse and complex distracted driving activities using UWB sensing data collected in real-driving environments. Building such datasets is time-consuming and labor-intensive, and the effort grows with the number of labels and the complexity of driving conditions. This burden may partly explain why many studies~\cite{xie2021hearsmoking, xie2019d, Uhead, blinkradar} focus on a single distracted activity. However, given the wide variety of distracted behaviors, monitoring only one activity is insufficient for preventing traffic accidents.}

\re{Although an existing DAR dataset~\cite{RaDA} includes multiple distracted driving activities, it is collected in simulated environments. As a result, it does not fully reflect real-world factors such as road conditions, vehicle-induced vibrations, and other environmental influences, which can limit generalizability to real driving scenarios.}

\re{The second challenge is the mismatch between UWB data and state-of-the-art DAR models. Vision Transformers (ViTs)~\cite{vit, han2022survey, khan2022transformers, ruan2022vision} have achieved strong performance across vision benchmarks and have therefore been adopted in UWB sensing. However, aligning UWB representations with fixed ViT input sizes is non-trivial: naive interpolation can distort radar-specific cues (e.g., Doppler shifts, phase, and attenuation), and pre-trained positional embedding vectors (PEVs) optimized for natural images may no longer match the UWB geometry, degrading performance.}

\re{Nevertheless, many ViT-based UWB studies~\cite{li2023human, zhao2022distributed, lai2023vision, li2023advancing, kim2023study} treat resizing as a minor implementation detail and do not provide domain-specific handling of UWB input geometry. This omission can cause substantial performance loss, motivating an input-size-agnostic strategy that preserves radar-domain features while remaining compatible with pre-trained ViTs.}

\re{
\textbf{Research questions.}
To address these challenges, this study investigates the following research questions: 
\begin{itemize}[leftmargin=*]
  \item Can IR-UWB radar sensing reliably recognize diverse distracted driving activities in real-world driving environments? 
  \item How can Vision Transformers be effectively adapted to UWB radar data with varying input sizes without distorting domain-specific information? 
  \item Does fusing range and frequency domain representations improve the performance of UWB-based driver activity recognition?
\end{itemize}
}

\textbf{Approaches.}
To address the first challenge, the lack of real-driving datasets, we collect the \ALERT dataset, comprising 10,220 samples of 7 distracted driving activities captured in real-driving environments.
We present benchmarking performance of widely used 8 learning algorithms—CNN-based~\cite{googlenet, resnet, densenet, mobilenet}, RNN-based~\cite{maitre2020fall, maitre2021recognizing, noori2021ultra}, and transformer-based approaches~\cite{deit, vit}—that are applicable in UWB DAR. 
Additionally, we design experiments to provide the effective utilization of the \ALERT dataset across these 8 algorithms such as analyzing observation time, multipath effects, and frequency area. Consequently, we make the \ALERT dataset publicly available to promote its adoption and foster further advancements in UWB DAR research.

To overcome these compatibility issues, we propose the Input-Size-Agnostic Vision Transformer~(\mine), which introduces a customized resizing scheme for UWB data, preserving domain-specific features such as range and frequency information. By strategically adjusting patch dimensions and leveraging the pre-trained PEVs, \mine maintains spatial coherence and avoids the signal corruption often caused by naive resizing.
This design also modifies the linear projection layers to accommodate altered patch sizes, ensuring seamless integration with pre-trained ViT weights. Our experiments reveal that this tailored approach significantly improves performance while retaining straightforward implementation.

Furthermore, we develop a domain fusion strategy that leverages both range and frequency domain representations, which preserve distinct yet complementary signal characteristics. By applying our input-size-agnostic resizing and then fusing the extracted features, we capitalize on the synergy between the two domains, boosting classification accuracy and reinforcing the benefits of an appropriately resized radar input.

The contributions of this work are summarized as follows:
\begin{itemize}[leftmargin=*]
  \item We present the \ALERT dataset, the first UWB dataset capturing comprehensive driving activities in real-driving environments. With 10,220 samples across 7 activities, we evaluate its performance using 8 learning algorithms, spanning CNN-based, RNN-based, and transformer-based methods. By making the dataset publicly available, we aim to promote its adoption and support diverse experimentation tailored to various research needs.
  
  \item We propose \mine, a novel model addressing challenges in adapting pre-trained ViTs to UWB data with varying input sizes. \mine introduces a resizing method that avoids information loss and incorporates a domain fusion technique combining range and frequency domain features, significantly enhancing classification accuracy.

  \item Through extensive experiments, we demonstrate \mine's superior performance, achieving a classification accuracy of 76.28\% and distracted driving detection accuracy of 97.35\%. Additionally, we present various benchmarking results and trade-offs for selecting suitable models in UWB DAR. Our work provides insights for future studies through benchmarking and the provision of the open \ALERT dataset.
\end{itemize}

\re{The rest of this paper is organized as follows. Sec.~\ref{sec:related} reviews related works on IR-UWB-based activity recognition and vision-transformer-based modeling. Sec.~\ref{sec:dataset} introduces the ALERT dataset and the preprocessing to construct range–time and frequency–time representations. Sec.~\ref{sec:benchmark} describes the experimental protocol, baseline configurations, and evaluation metrics for fair benchmarking. Sec.~\ref{sec:ISA-ViT} presents the proposed input-size-agnostic ViT with domain fusion. Sec.~\ref{sec:evaluation} reports the experimental results and analyses. Sec.~\ref{sec:discussion} discusses key observations and limitations, and Sec.~\ref{sec:conclusion} concludes the paper.}

\section{Related Works}\label{sec:related}
\textbf{DAR using UWB radar.} Recently, several DAR systems utilizing UWB technology have emerged~\cite{blinkradar, RaDA, Uhead, khan2022ir}. 
\rev{\mbox{Xu~\emph{et al}.~\cite{Uhead}}} present UHead, a driver attention monitoring system based on UWB radar. Unlike traditional methods that rely on specialized sensors or cameras, UHead tracks the driver's head motion to infer attention. It extracts time-frequency data from UWB radar signals to estimate the direction and angle of head rotations. However, detecting a driver's distracted behavior solely based on head rotation has its limitations.
Drivers can exhibit distracted behaviors without rotating their heads significantly such as using a mobile phone or smoking while driving.

Similarly, \rev{Hu~\emph{et al.}~\cite{blinkradar}} design a system for fine-grained eye-blink monitoring in driving scenarios using customized IR-UWB radar. The system extracts eye-blink signals while filtering out interference from body movements and environmental factors. However, this approach is limited as it only targets drowsiness, overlooking other potential distractions that may occur during driving.

\rev{Khan~\emph{et al.}~\cite{khan2022ir}} present a radar-based system for robust heart rate detection intended for in-car monitoring. They propose a novel time-series-based algorithm instead of traditional frequency-domain methods, using a deep learning classifier to detect heart rate patterns. Additionally, they optimize the radar sensor's location inside the car to reduce interference from random body motions.

However, the above studies have focused solely on a single distracted driving activity. Given the diverse causes of traffic accidents resulting from distracted driving, targeting only one type of activity is insufficient to effectively prevent serious traffic incidents. To address this issue, there is a need for studies that encompass a comprehensive range of distracted driving activities.
 
The study in~\cite{RaDA} introduces an IR-UWB radar-based monitoring system capable of detecting a wide range of in-car activities. It also provides an open dataset for DAR in a simulated driving environment, covering 6 distracted driving activities: driving, autopilot, sleeping, driving while using a smartphone, smartphone use, and talking to a passenger. To the best of our knowledge, this is the first open dataset for UWB DAR in a simulated driving environment addressing six distracted driving activities. 

\rev{However, simulated-driving environments fail to accurately represent real-world driving conditions, which involve factors such as ground vibrations, vehicle vibrations, and driving realism. For example, Morales-Alvarez \emph{et al}.~\cite{morales} report that the recognition accuracy of driver observation models decreased from 85.7\% to 46.6\% when transferred from simulation to real-world scenarios. Similarly, Stocco \emph{et al.}~\cite{stocco} emphasize the “reality gap” in simulation-based testing for autonomous systems, noting that critical cues like vibration, cabin constraints, and natural driver reactions are often missing in simulated settings.}

Additionally, imposing specific restrictions on driver activities (e.g., requiring the use of the right hand while talking or mandating two-handed smartphone use) may make the dataset more distinguishable but less realistic. Consequently, a realistic \textit{de facto} standard open dataset is essential to advance UWB DAR studies. To address this need, we introduce the \ALERT dataset in the following section, designed with realistic properties in mind.

Compared to the dataset from~\cite{RaDA}, the \ALERT dataset focuses on more detailed distracted driving activities in \emph{real-world driving environments}. It includes data for 7 activities: relaxation (autopilot), driving, nodding, smoking, drinking, panel control, and smartphone use. Additionally, the \ALERT dataset provides both range and frequency domain data, enabling a variety of experimental approaches to be explored using the data.

\textbf{The use of ViT with UWB.}
Various works~\cite{li2023human, zhao2022distributed, lai2023vision, li2023advancing, kim2023study} introduce the use of ViT with UWB. \rev{Li~\emph{et al}.~\cite{li2023human}} introduce MobileViTX, an improved version of the MobileViT architecture, specifically designed for human activity recognition~(HAR) using IR-UWB radar. The Zhao \emph{et al}.~\cite{zhao2022distributed} explore the feasibility of classifying human activities using data captured by a distributed UWB radar system. They compare ViT with CNN-based architectures and validate the robustness of these models, demonstrating ViT's ability to generalize to unseen participants.

Moreover, \rev{Lai~\emph{et al.}~\cite{lai2023vision}} focus on developing a non-invasive sleep posture recognition system using multiple UWB radars. They employ machine learning models for classification, including CNN-based networks and ViT. \rev{Kim~\emph{et al}.~\cite{kim2023study}} introduce a new approach for 3D human pose estimation using IR-UWB radar, leveraging a transformer-based deep learning model.

The above studies have attempted to apply UWB data to ViT. However, none of them provide sufficient discussion on how to adapt UWB data, which has a different input size from images, for use in ViT. The unique study~\cite{ast} addresses how to apply audio data to ViT when the input shapes differ. \rev{Gong~\emph{et al}.~\cite{ast}} introduce the audio spectrogram transformer~(AST) model, the first purely attention-based model for audio classification. They cut and interpolate the PEVs for height or width in the AST model. By doing so, they transfer 2D spatial knowledge from the pre-trained ViT to the AST model, even when the input shapes are different. 
\re{Beyond vision and audio, transformer-based models have also been explored for non-visual spatio-temporal and time-series datasets, including 2D transformer formulations for financial time-series prediction and temporal graph-based modeling that captures structural and temporal dependencies~\cite{Celik2025MotifCrypto, Tuncer2022Deep2DTransformer, Uygun2025TemporalGNN}.}
However, excessive cutting or interpolation of PEVs may incur the ruin of spatial information in pre-trained ViT~\cite{ruan2022vision}. Therefore, we need to explore a method that is not affected by input size while still utilizing the pre-trained parameters of the ViT.

\section{ALERT Open Dataset}\label{sec:dataset}
To collect comprehensive activities related to distracted driving in real-driving environments, we introduce the \ALERT dataset in this section.

\subsection{Sensor Specification}
\begin{table}[t]
    \centering
    \begin{minipage}{0.45\textwidth}
        \centering
        \small
        \caption{UWB radar parameters for data collection}
        \vspace{3pt}
        \begin{tabular}{|c|c|}
            \hline
            \textbf{Parameters} & \textbf{Values} \\ 
            \hline
            Tx center frequency & 8.748~GHz \\
            \hline
            Tx bandwidth ($-10$ dBm) & 1.5~GHz \\
            \hline
            Energy per pulse & 0.3--1.7~pJ \\
            \hline
            Frame per second & 100~Hz \\
            \hline
            Range (bins) & 0--9~m (178) \\
            \hline
            Sampling frequency & 23.328~GHz \\
            \hline
            Pulse repetition frequency & 15.1875~MHz \\
            \hline
        \end{tabular}
        \label{table1_1}
    \end{minipage}%
    \hfill
    \begin{minipage}{0.45\textwidth}
        \centering
        \includegraphics[width=\linewidth, height=2.5cm]{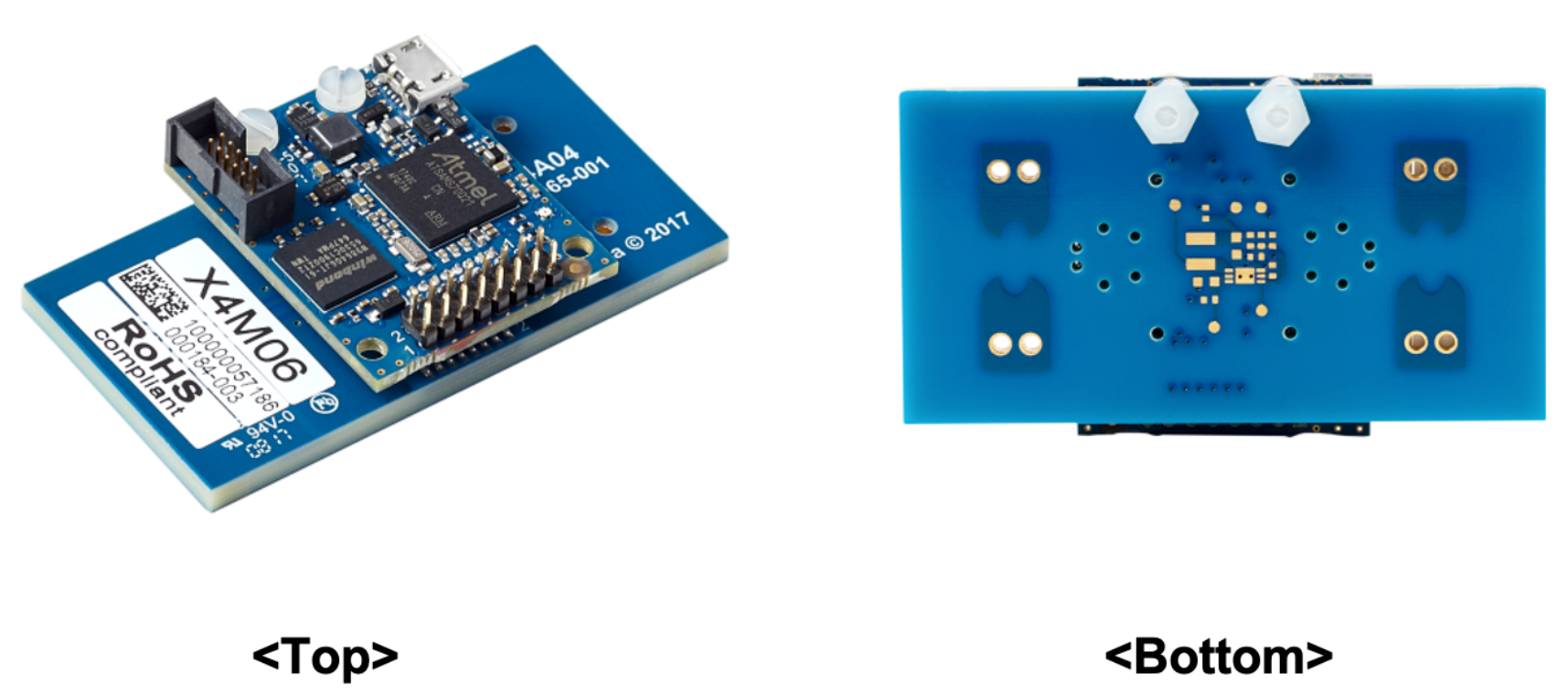}
        \caption{Xethru X4M06 UWB radar sensor by Novelda.}
        \label{fig:device}
        \vspace{-15pt}
    \end{minipage}
\end{table}

We capture the \ALERT dataset using the UWB radar sensor Xethru X4M06, developed by Novelda~\cite{x4m06}, as depicted in Fig.~\ref{fig:device}. The X4 series are the off-the-shelf devices in UWB radar technology and are widely utilized in several UWB HAR studies~\cite{RaDA, HARSANET, SleepPoseNet}. The radar features two directional patch antennas (Tx and Rx), optimized for frequencies between \mbox{7.25--10.2}~GHz with an opening angle of 65$^{\circ}$ in both azimuth and elevation. The detailed parameters of the radar are presented in Table~\ref{table1_1}.

\subsection{Dataset Manipulation}
\begin{figure}[t]
    \centering
    \includegraphics[width=0.45\textwidth]{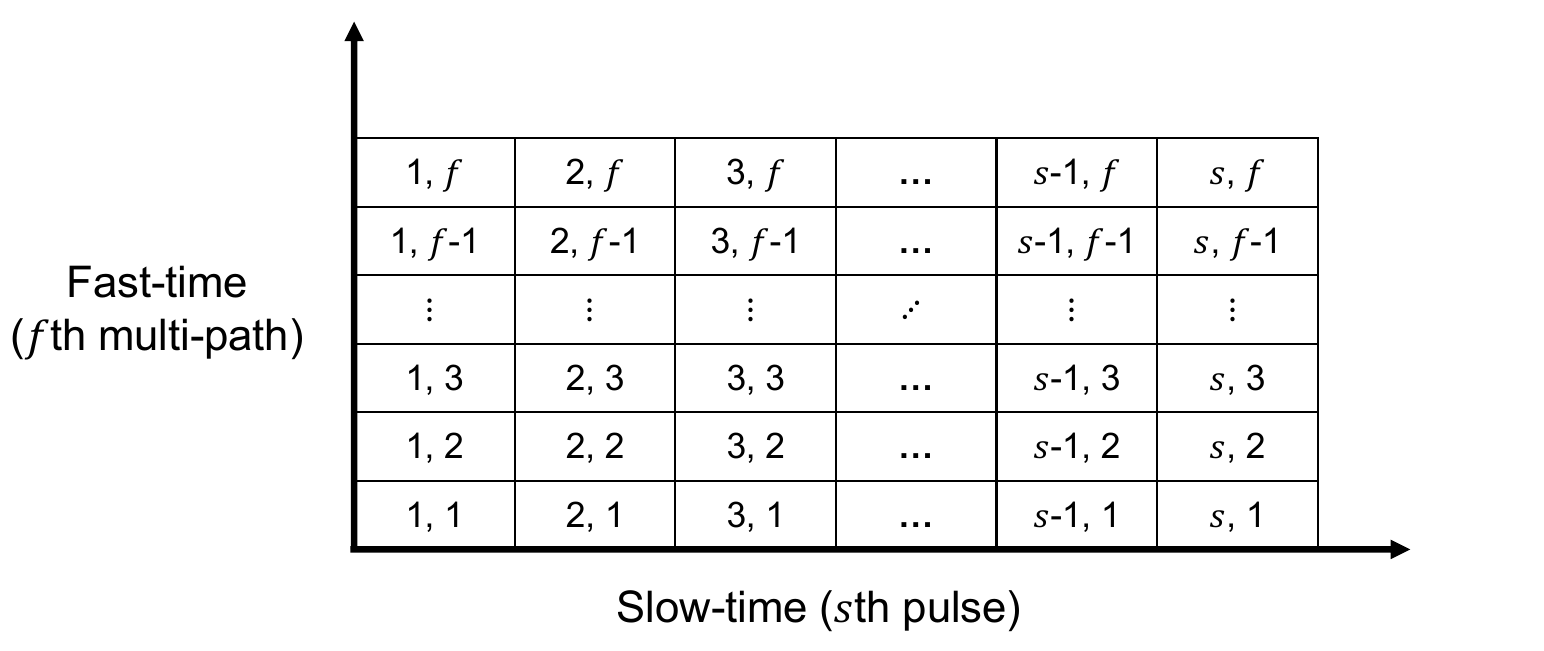}
    \vspace{-10pt}
    \caption{Manipulation of UWB signal with two-dimensional fast-time vs. slow-time.}
    \label{fig:Fastvsslow}
    \vspace{-10pt}
\end{figure}

We need to process the UWB data from radar sensors to extract and provide various types of information. 
The IR-UWB radar periodically transmits the impulse signal; $s$th pulse has the slow-time index $s$. When the radar transmits the $s$th pulse, it comes back to radar as several signals due to multi-path. We can generalize the multi-path signal as $f$th multi-path; $f$th multi-path has the fast-time index $f$. To sum up, the fast-time index indicates $f$th multi-path signal, and the slow-time index indicates $s$th transmitted impulse signal as shown in Fig.~\ref{fig:Fastvsslow}. Simply, we can express the received signal as follows
\begin{equation}
    r_{s}(t) = \sum^{F}_{f=1}\alpha_{f}\delta(t-\tau_{f}) * m_{s}(t) + n_{s}(t),
\end{equation}
where $r_{s}(t)$ is a received signal from $s$th impulse, $F$ is the number of multi-path for transmitting signal $m_{s}(t)$, $n_{s}(t)$ is noise, and $\alpha_{f}$ is the attenuation factor of $f$th multi-path. For each pulse, the multi-path indicates time delay so that we call the fast-time vs. slow-time 2D data as range data. Moreover, we can also obtain the frequency data by performing fast Fourier transform~(FFT) to range data. As frequency data can express the Doppler frequency shift, we can get the movement characteristic of object. 

\begin{figure}[t]
\centering
    \subfigure[Range data]{
    \hspace{0.5cm}
    \includegraphics[width=7cm, height = 3cm]{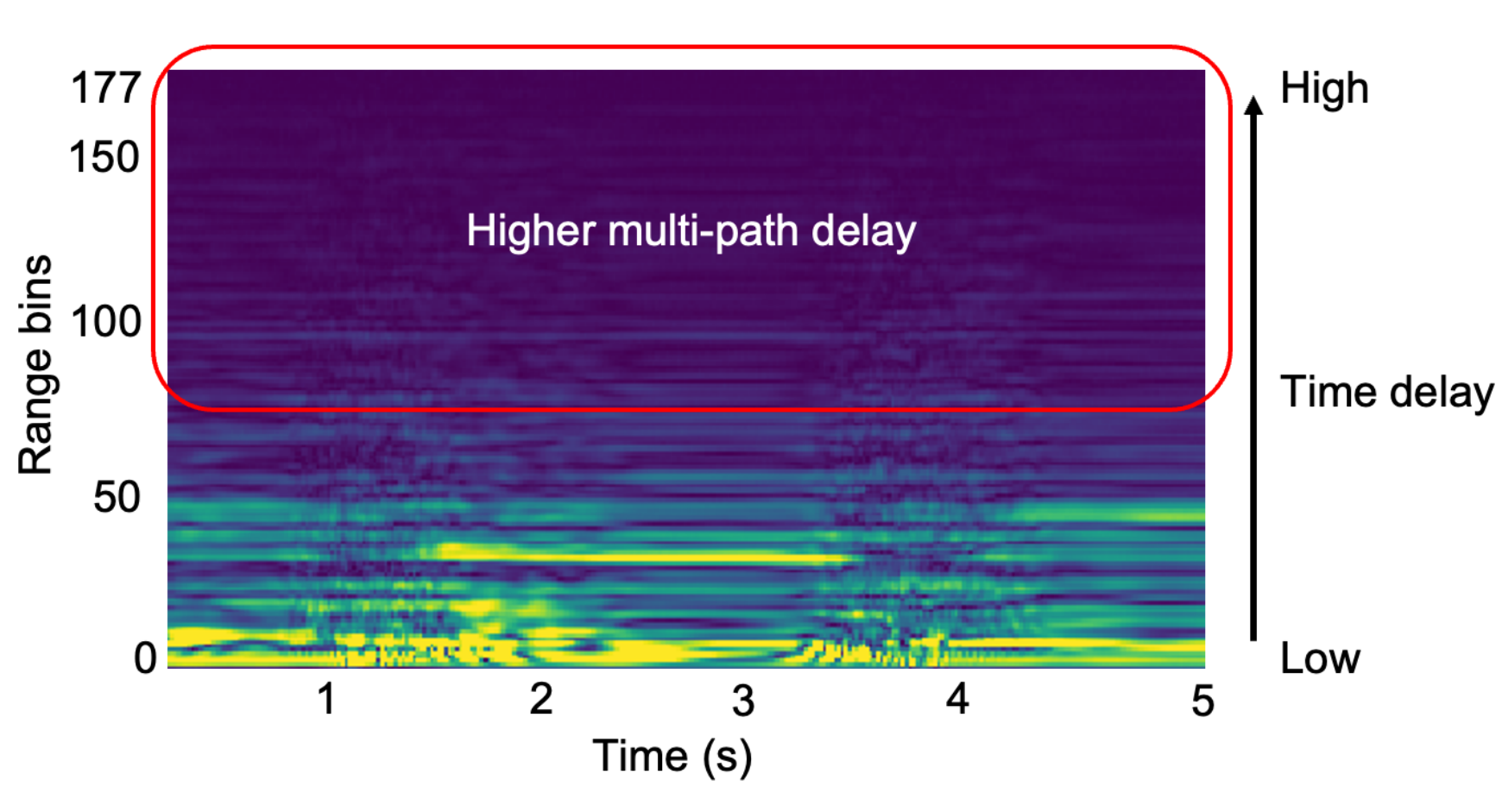}
    \label{fig:temporal}
    }
    \hfill
    \centering
    \subfigure[Frequency data]{
    \hspace{1cm}
    \includegraphics[width=7cm, height = 3cm]{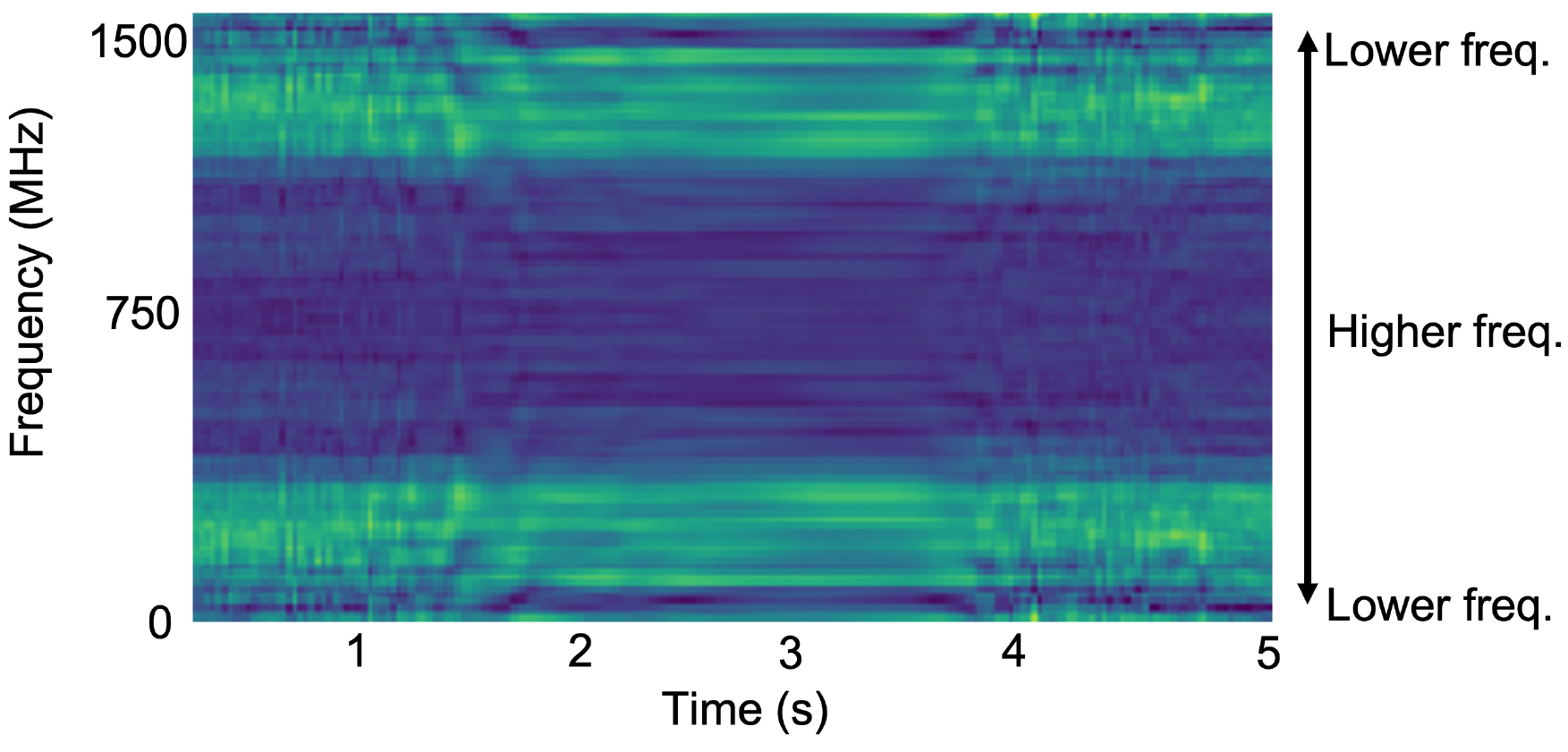}
    \label{fig:frequency}
    }
    \hfill    
    \vspace{-10pt}
    \caption{Range and frequency data from UWB signal.}
    \label{fig:signal manipulation}
    \vspace{-10pt}
\end{figure}

We obtain the range data as shown in Fig.~\ref{fig:temporal}. The Y-axis of Fig.~\ref{fig:temporal} represents the fast-time index, which indicates range bin from the radar device. Although the distance between the radar and the driver is at most less than 2 meters, the upper part of Fig.~\ref{fig:temporal} may provide critical information for detecting activities according to the learning algorithm (e.g., the size of convolutional layer filters). To this end, we provide our dataset, which allows users to customize the information about the range according to their specific needs. We provide the whole range bins of 178 by default~(about 9 m). 

For frequency data, discrete FFT transforms the signal into periodic expressions, as illustrated in Fig.~\ref{fig:frequency}. Movements can be expressed across a range of frequencies, so significant information can be obtained from specific frequency bands relevant to the activity. To this end, we provide a dataset that enables users to freely crop the frequency information to facilitate versatile data utilization; we provide 178 frequency bins by default. Additionally, we offer an adjustable observation time (slow-time window), allowing users to select data with an observation time that best suits their needs.

\begin{figure}[t]
        \subfigure[Detailed in-vehicle environment and radar deployment]{
        \hspace{0.5cm}
            \includegraphics[width=6cm, height=4cm]{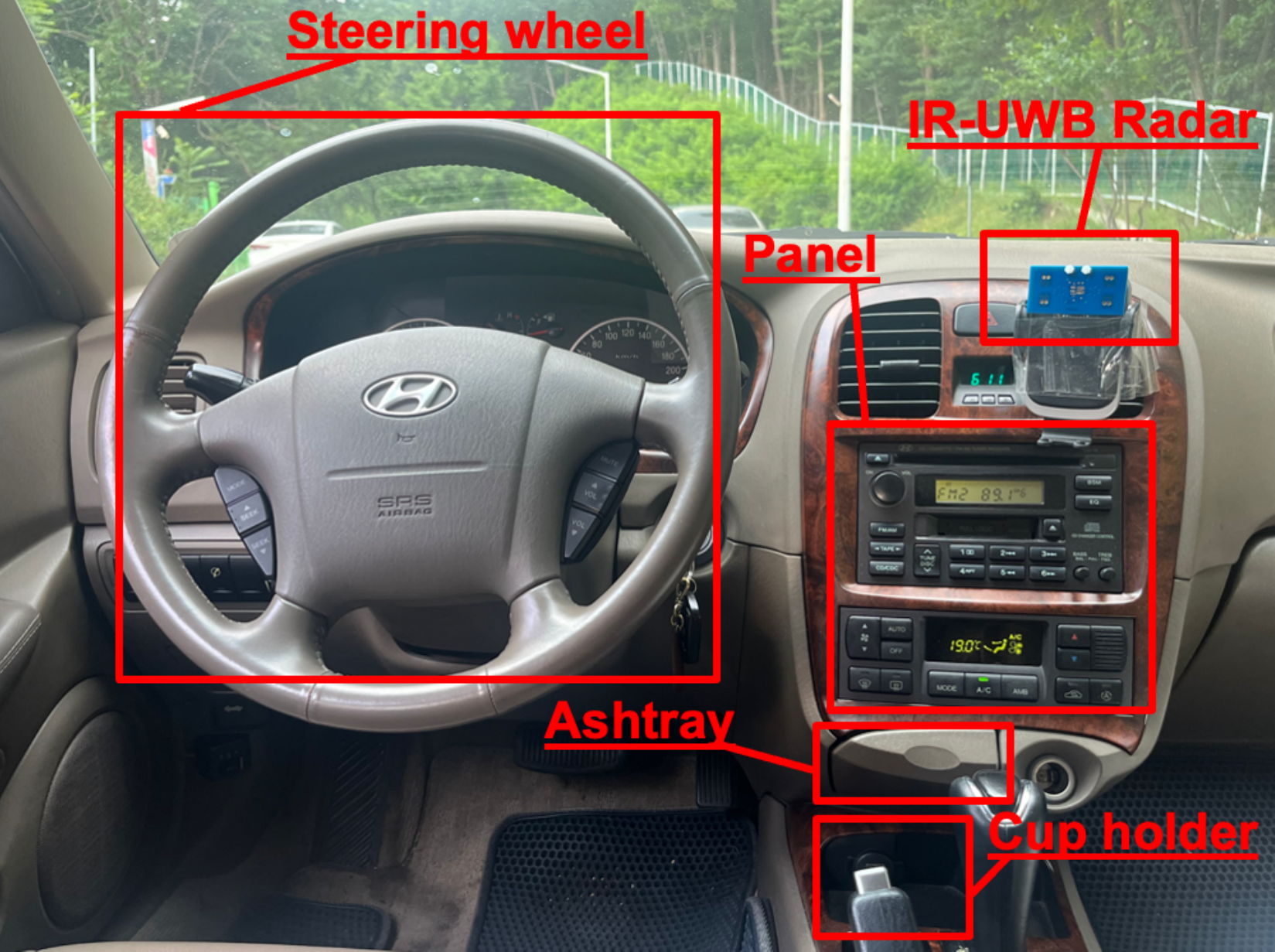}
            \label{fig:incar}
        }
        \hspace{1pt}
        \subfigure[Urban route]{
            \includegraphics[width=3.4cm, height=4.5cm]{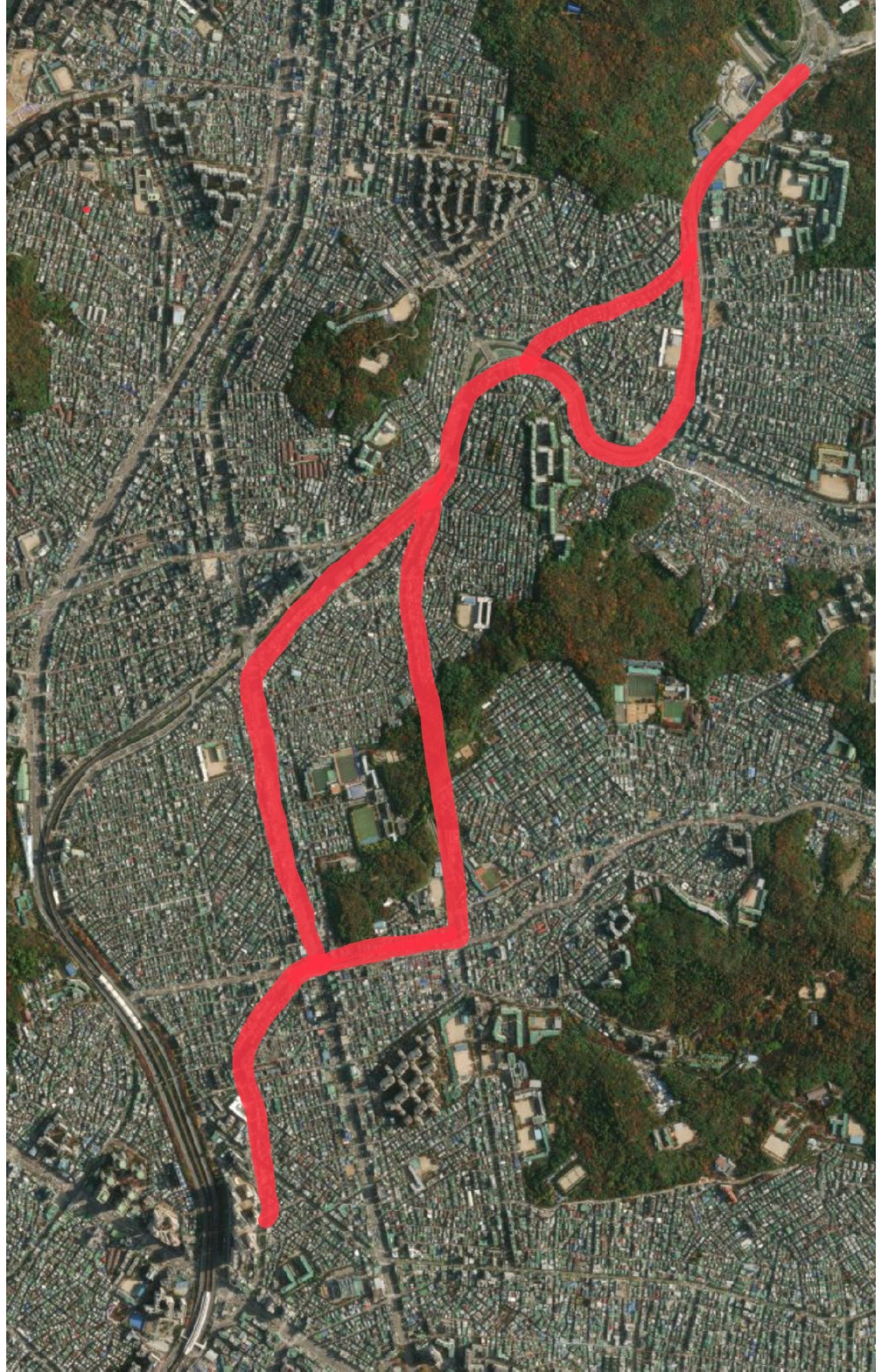}
            \label{fig:DT1}
        }
            \hspace{3pt}
        \subfigure[Campus]{
            \includegraphics[width=3.4cm, height=4.5cm]{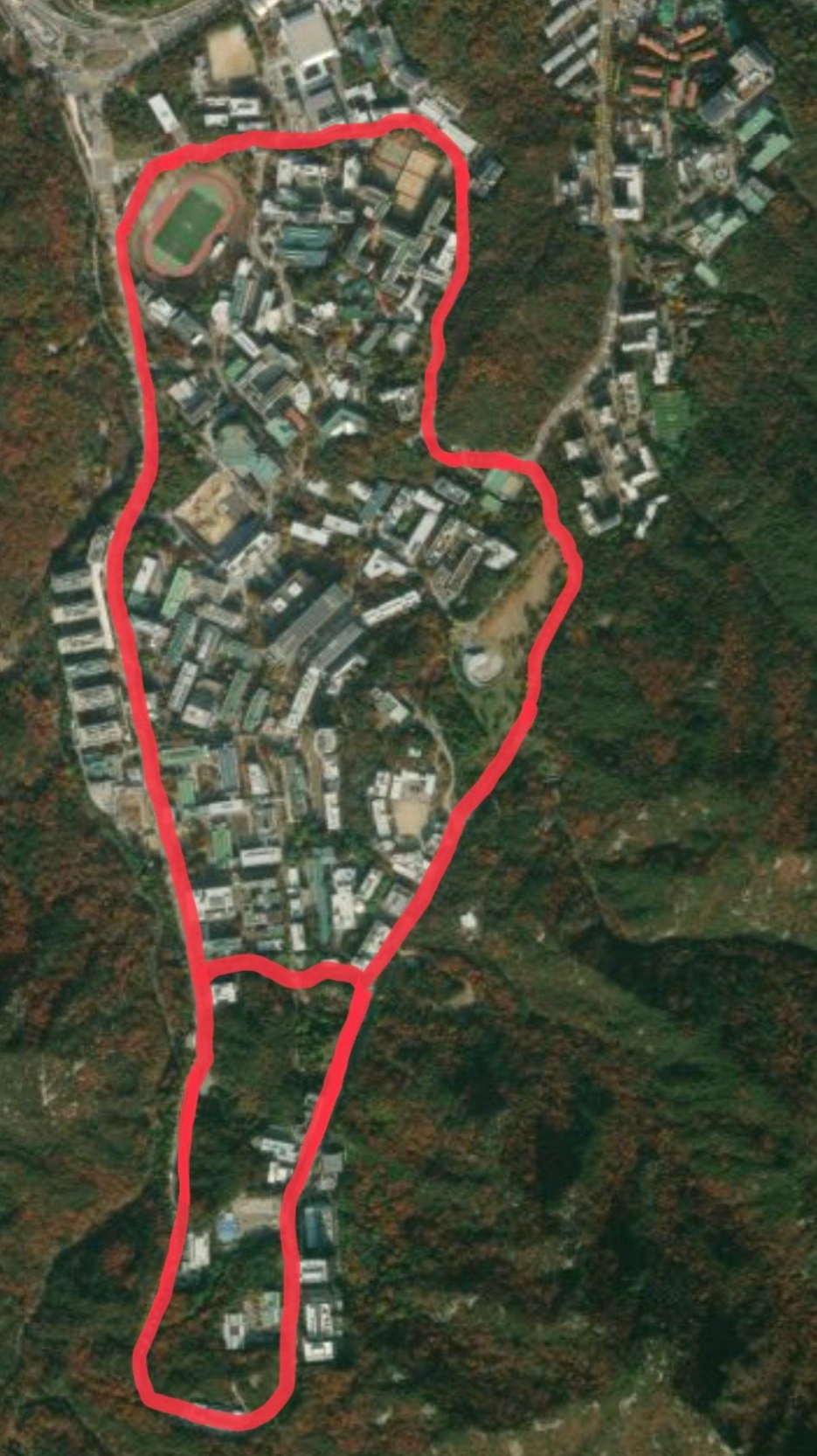}    
            \label{fig:DT2}
        }
            \hfill    
            \caption{Driving environment and driving course for data collection.}
            \label{fig:DT}
            \vspace{-15pt}
\end{figure}

\subsection{Data Collection}
Our dataset exhibits a significant strength: We collect the dataset in a real-driving environment. Accordingly, we mount the radar within an actual vehicle, as depicted in Fig.~\ref{fig:incar}. We strategically attach the radar to the vehicle's air vent.
\re{This position does not obstruct the driver’s field of view and is located at a height similar to the driver’s upper body (eye and chest level), enabling effective capture of driver movements with minimal occlusion from passengers or interior structures.
By fixing the radar at this location, a stable and repeatable sensing geometry is maintained during driving, which is essential for consistent signal acquisition.}

Moreover, considering that modern smartphones are already equipped with built-in UWB chipsets, we have explored the potential integration of UWB radar into mobile phones. Since most drivers already use smartphone holders clipped to the air vent, mounting the UWB radar in the same location is not only practical but also aligns well with real-world usage patterns, supporting easy installation and potential commercialization.

\begin{figure*}[t]
    \centering
    \subfigure[Asphalt of urban route]{
        \centering
        \includegraphics[height = 2cm, width=0.2\textwidth]{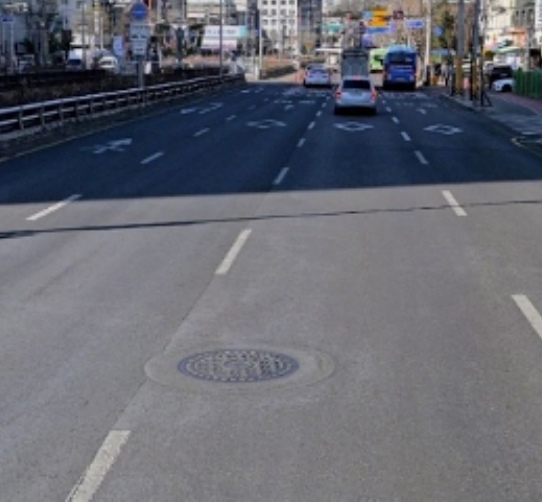}
        \label{fig:urban1}
    }
    \subfigure[Tiled-concrete of urban route]{
        \centering
        \includegraphics[height = 2cm, width=0.2\textwidth]{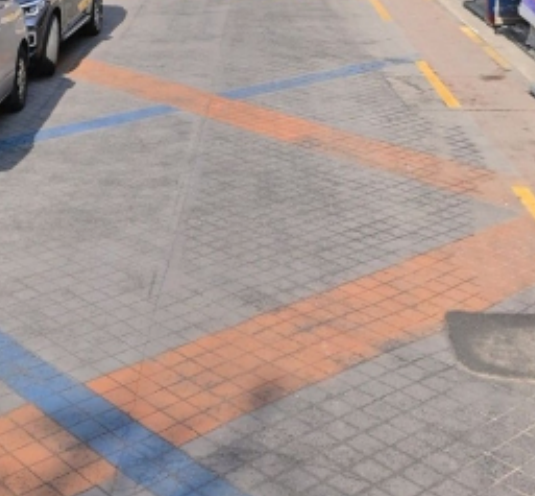}
        \label{fig:urban2}
    }

    \vspace{-5pt} 

    \subfigure[Cobblestone of campus route]{
        \centering
        \includegraphics[height = 2cm, width=0.2\textwidth]{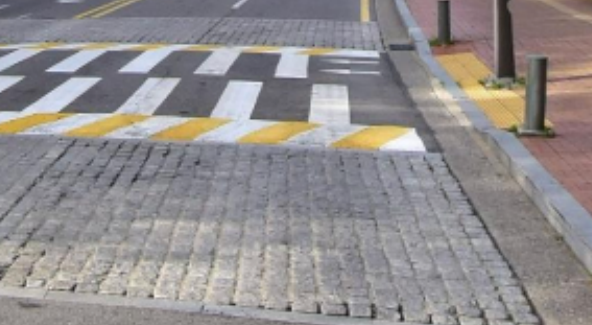}
        \label{fig:campus1}
    }
    \subfigure[Dirt path of campus route]{
        \centering
        \includegraphics[height = 2cm, width=0.2\textwidth]{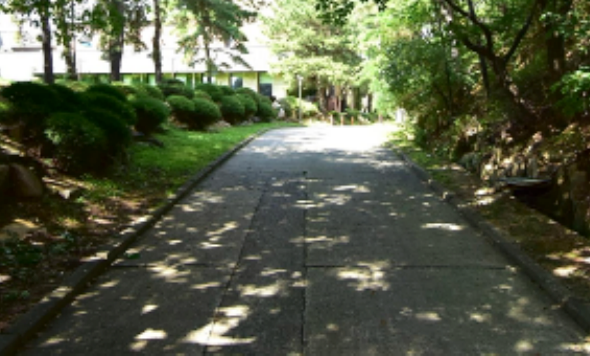}
    }
    \subfigure[Concrete of campus route]{
        \centering
        \includegraphics[height = 2cm, width=0.2\textwidth]{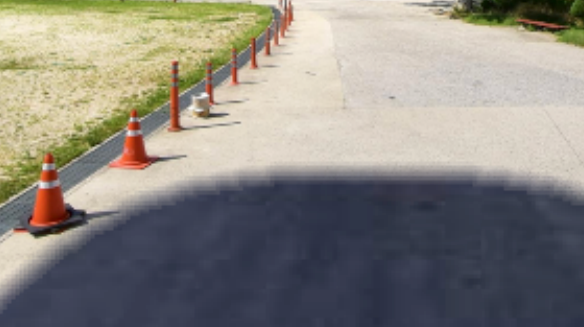}
    }
    \subfigure[Speed bump of campus route]{
        \centering
        \includegraphics[height = 2cm, width=0.2\textwidth]{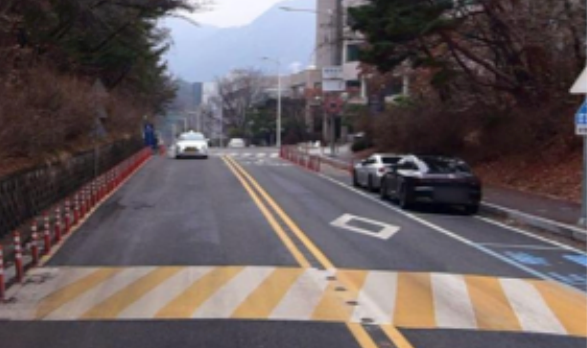}
        \label{fig:campus4}
    }
    \caption{\rev{Various road conditions of urban and campus routes.}}
    \label{fig:2plus4layout}
    \vspace{-10pt}
\end{figure*}

\re{We drove along the designated red-line routes and collected the dataset as shown in Figs.~\ref{fig:DT1} and \ref{fig:DT2}. The urban route (12~km) provides relatively smooth, highway-like driving with light traffic and few signals; it is mostly flat with one uphill and one downhill segment (each $\sim$10$\degree$). The surface is primarily asphalt with partial tiled-concrete sections (Figs.~\ref{fig:urban1} and \ref{fig:urban2}), yielding generally stable vibrations mainly due to road irregularities.}

\re{The campus route is a 6~km loop with a 30~km/h speed limit, resulting in low-speed driving and frequent stop-and-go events at pedestrian crossings. It consists of roughly 50\% sloped and 50\% flat sections, with average gradients of $\sim$10.2$\degree$ uphill and 9.2$\degree$ downhill. While asphalt is dominant, it also includes cobblestone, dirt-road segments, and numerous speed bumps (Fig.~\ref{fig:campus1}--\ref{fig:campus4}), producing more diverse and irregular vibrations.}

\re{Together, the two routes cover contrasting real-driving conditions---smooth and steady motion versus low-speed stop-and-go with frequent slope changes and road-surface transitions---thus providing diverse acquisition scenarios while maintaining safe and repeatable data collection.}

To ensure a safer experiment, we adhere to the speed limit and have attached a sign to the vehicle indicating that an experiment is underway while driving. 

To detect the driver's activity while driving, we select seven labels: Relaxation~(\emph{Relax}), steering wheel control~(\emph{Drive}), nodding~(\emph{Nod}), smoking~(\emph{Smoke}), drinking~(\emph{Drink}), panel control~(\emph{{Panel}}), and using smartphone~(\emph{Phone}). Except for \emph{Drive}, the drivers must not perform all other activities while driving. \rev{According to the National Highway Traffic Safety Administration (NHTSA)~\cite{NHTSA}, the Governors Highway Safety Association (GHSA)~\cite{GHSA}, the Centers for Disease Control and Prevention (CDC)~\cite{CDC}, and the Insurance Institute for Highway Safety (IIHS)~\cite{IIHS}, National Institutes of Health (NIH)~\cite{NIH} in the United States and World Health Organization (WHO)~\cite{WHO_phone}, five of these activities---\emph{Nod, Smoke, Drink, Panel, and Phone}---are considered distracted driving and cause severe traffic accidents commonly observed in daily life.}

Additionally, we include the \emph{Relax} label to address the current issue of \emph{blind trust} in autonomous driving technology. Many drivers using autonomous driving technology tend to disengage from driving and take their hands off the steering wheel. This behavior poses significant risks, highlighting the need for monitoring even seemingly benign activities like relaxation while driving. \re{We explain seven activities as follows: \textbf{Relax}: the driver keeps a relaxed posture with hands off the wheel; \textbf{Drive}: normal driving with both hands on the wheel; \textbf{Nod}: drowsiness-like head dropping and recovery; \textbf{Smoke}: bringing a cigarette to the mouth and lowering the hand; \textbf{Drink}: lifting a drink to the mouth and returning it; \textbf{Panel}: reaching toward and interacting with the center console (e.g., buttons/controls); \textbf{Phone}: raising a phone to the ear (or near the head) and lowering it.}

\begin{table}[t]
    \centering
    \caption{Sample distribution for volunteers and activities.}
    \vspace{-5pt}
    \begin{tabular}{c}
        \includegraphics[width=0.48\textwidth]{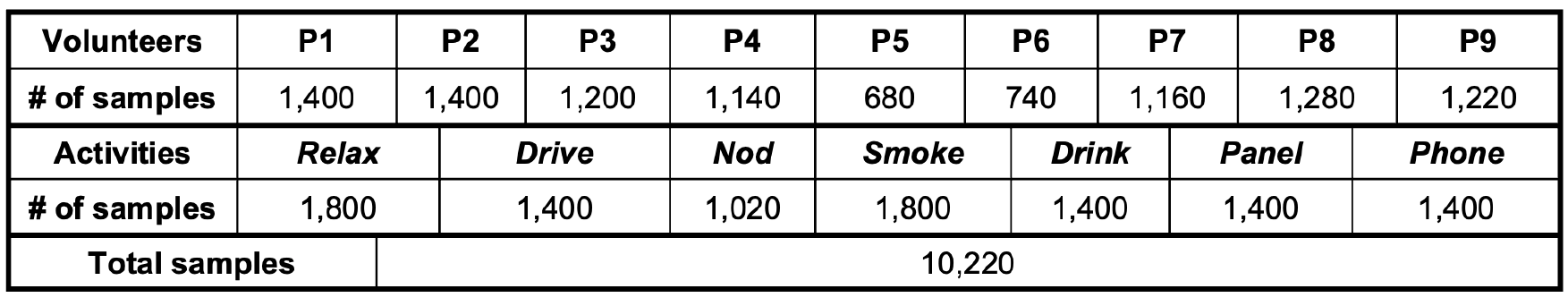}
    \end{tabular}
    \label{fig:data_dist}
    \vspace{-10pt}
\end{table}

\rev{The activities of \emph{Drive}, \emph{Smoke}, \emph{Phone}, and \emph{Panel} are continuous and can be sustained over a long period. For these activities, we have collected data continuously during actual driving.  
In contrast, \emph{Drink} and \emph{Nod} are inherently short in duration. For these activities, we asked participants to perform the actions 2–3 times within a 10-second window in a natural and realistic manner. For \emph{Nod}, the data captured moments when the driver’s steering control weakened, their head dropped, and then returned to the original position. For \emph{Drink}, the sequence included retrieving a drink from the cup holder, taking a sip, and placing it back.}

\rev{Additionally, \emph{Relax}, and \emph{Nod} are impractical to perform safely during real driving. Therefore, data for \emph{Relax} and \emph{Nod} are collected while the engine is running but the vehicle is stationary.  
In summary, for short-duration activities like \emph{Drink} and \emph{Nod}, participants are only given a time window and acted based on their natural behavior without detailed instructions.
}

Importantly, we provide minimal guidance on using the right hand for \emph{Smoke}, \emph{Drink}, and \emph{Panel} activities, as the ashtray, cup holder, and panel are typically located on the driver’s right-hand side as shown in Fig.~\ref{fig:incar}. For this reason, drivers do not intentionally cross their arms with their left hand to perform these actions. Other than these instances, the drivers can perform the actions freely, reflecting habitual behaviors in the aforementioned activities. 

We have obtained the dataset from nine volunteers~\footnote{\re{All data were collected with informed consent and in accordance with applicable ethical standards. The UWB radar signals were anonymized and do not contain visual, audio, or personally identifiable information, thereby preserving participants’ privacy even for potentially sensitive behaviors.}}, yielding a total of 10,220 samples (per 5 seconds) as shown in Table~\ref{fig:data_dist}. \rev{The participants consist of individuals with heights ranging from 165 to 188 cm and weights between 53 and 100 kg. Their ages range from 26 to 35 years old, with an average age of 32. Their driving experience spans from 1 to 10 years, with an average of 6 years. All participants are familiar with typical passenger vehicles, primarily sedans and SUVs. In addition, all participants reported prior experience with the types of distracted driving activities included in the ALERT dataset.}

\subsection{Dataset Representation}
\begin{figure*}[t!]
    \centering
    \begin{minipage}[b]{0.13\textwidth}
        \centering
        \includegraphics[width=\linewidth]{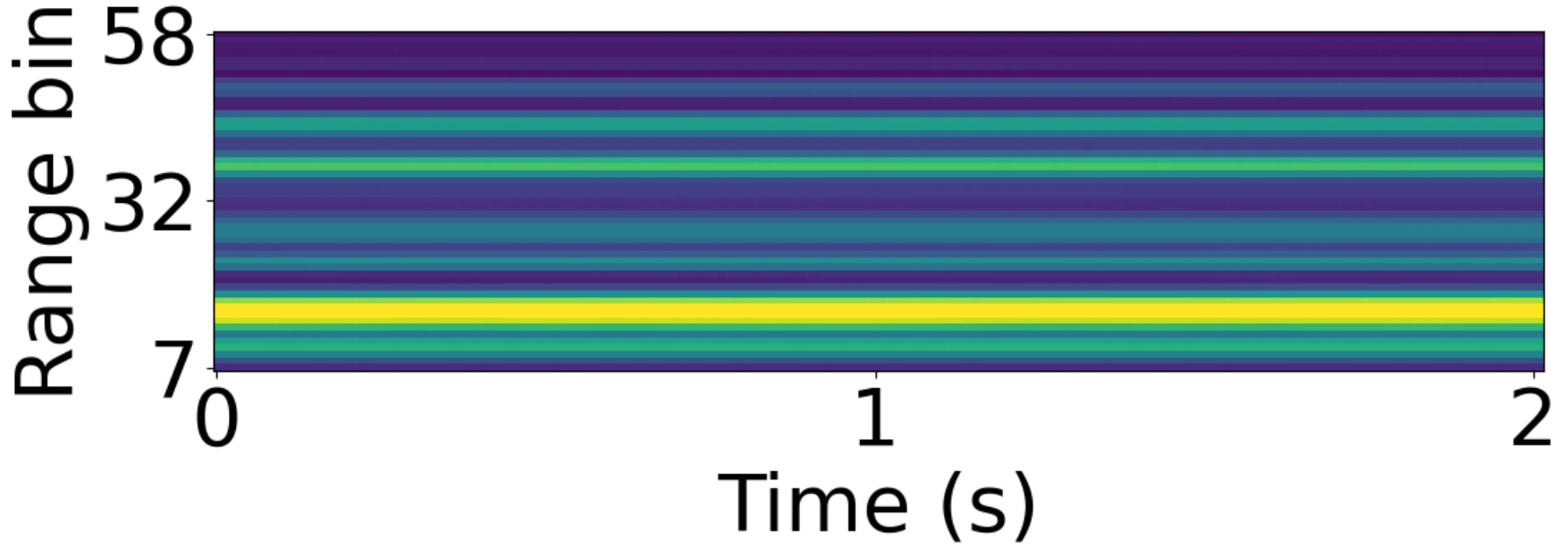}\\[-4pt]
        \includegraphics[width=\linewidth]{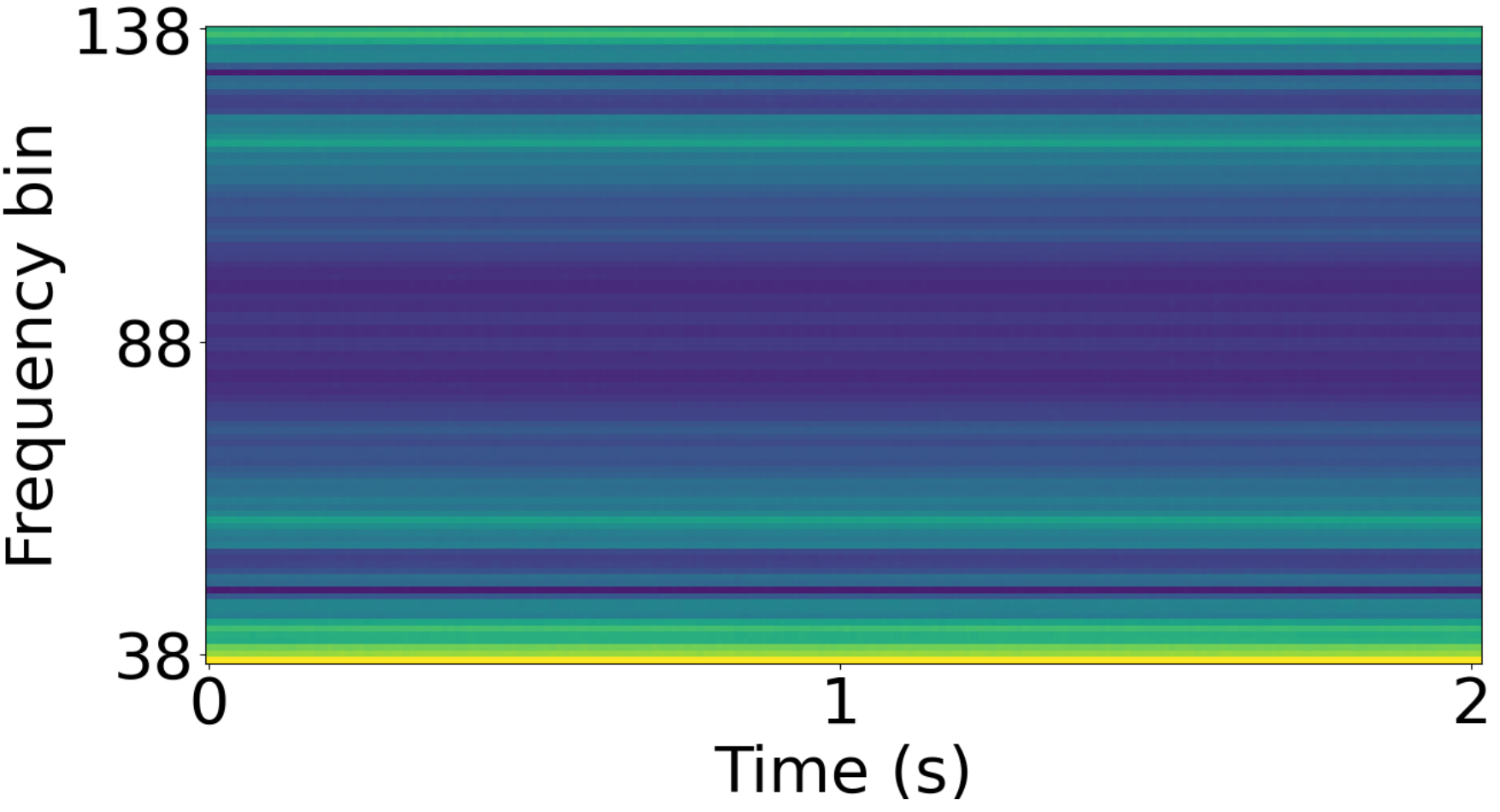}
        \\[1pt]
        \footnotesize\emph{Relax}
    \end{minipage}
    \hspace{-2pt}
    \begin{minipage}[b]{0.13\textwidth}
        \centering
        \includegraphics[width=\linewidth]{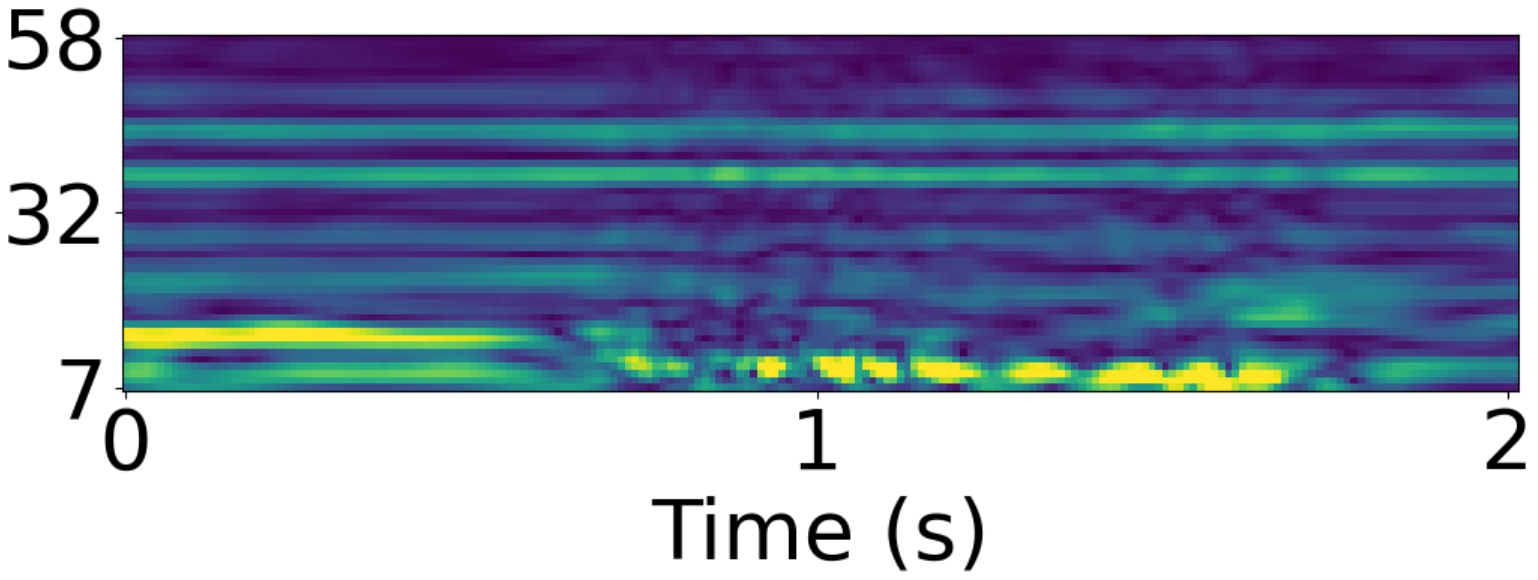}\\[-4pt]
        \includegraphics[width=\linewidth]{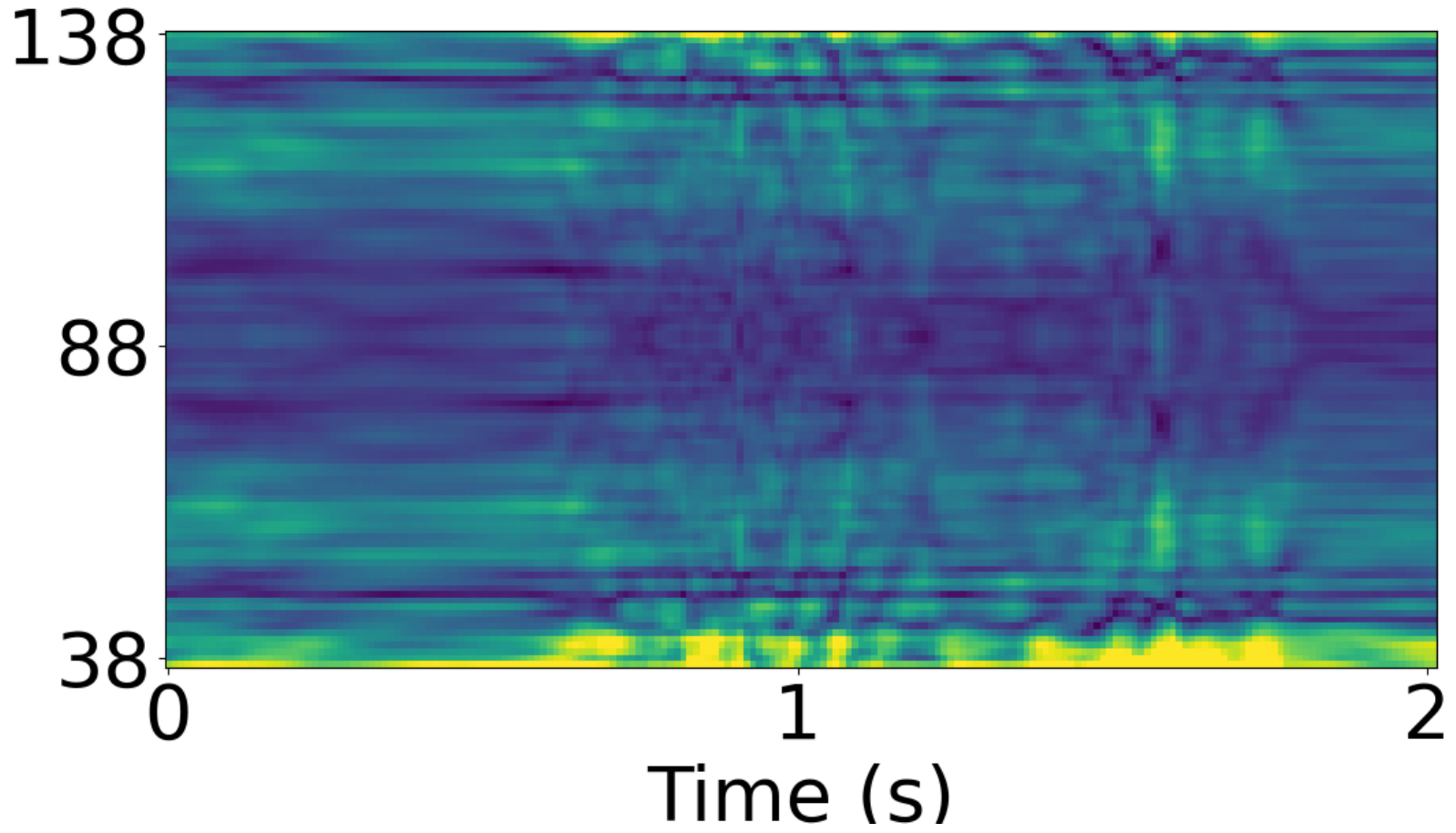}
        \\[1pt]
        \footnotesize\emph{Drive}
    \end{minipage}
    \hspace{-2pt}
    \begin{minipage}[b]{0.13\textwidth}
        \centering
        \includegraphics[width=\linewidth]{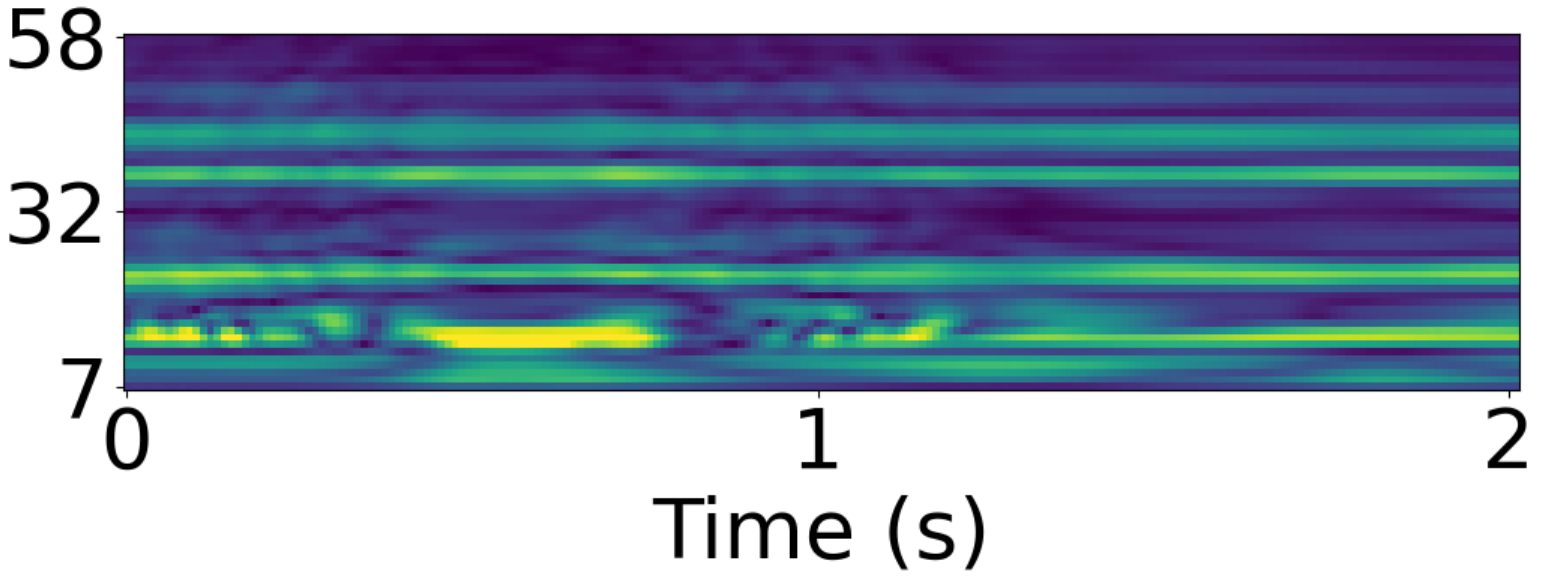}\\[-4pt]
        \includegraphics[width=\linewidth]{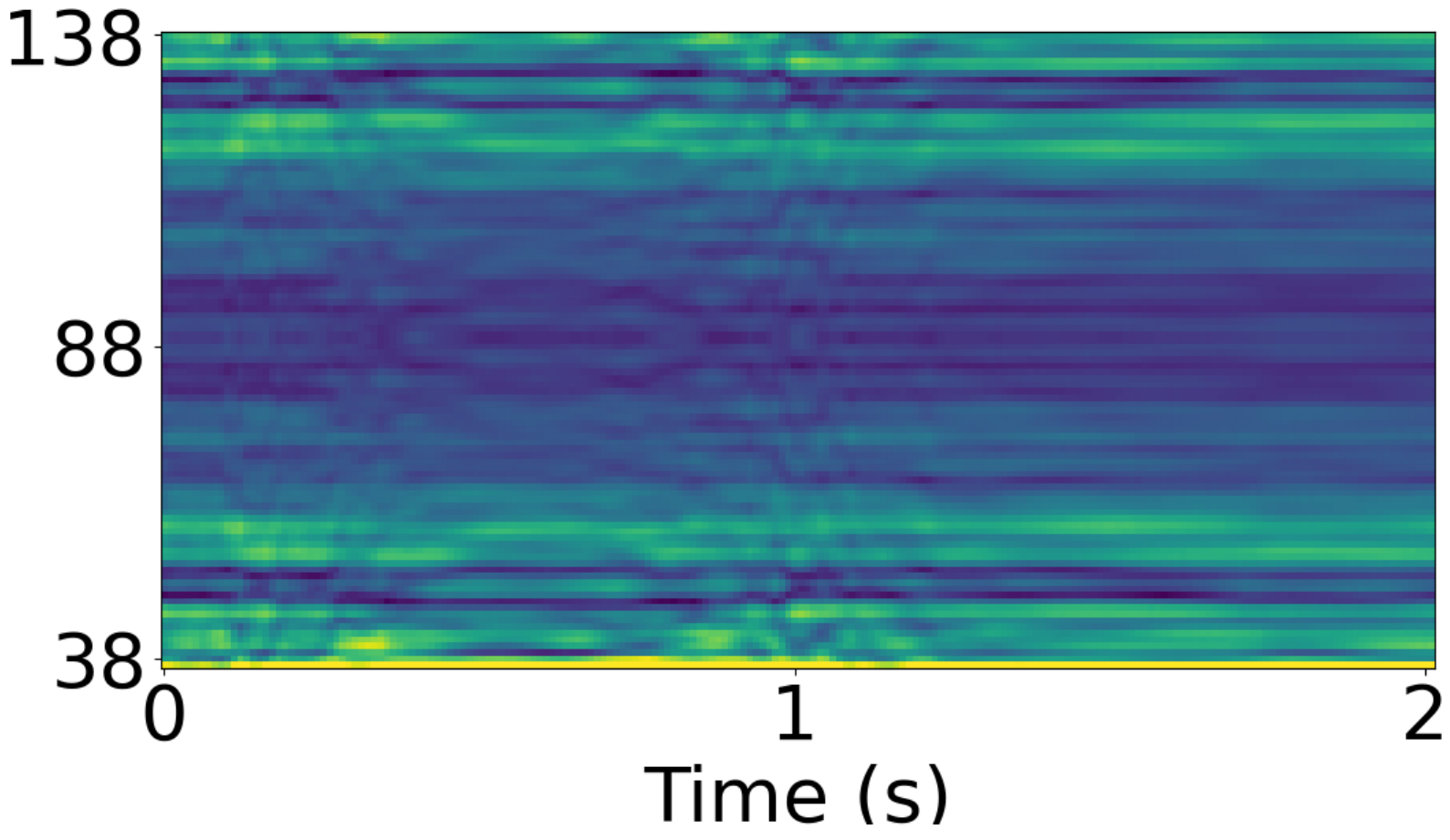}
        \\[1pt]
        \footnotesize\emph{Nod}
    \end{minipage}
    \hspace{-2pt}
    \begin{minipage}[b]{0.13\textwidth}
        \centering
        \includegraphics[width=\linewidth]{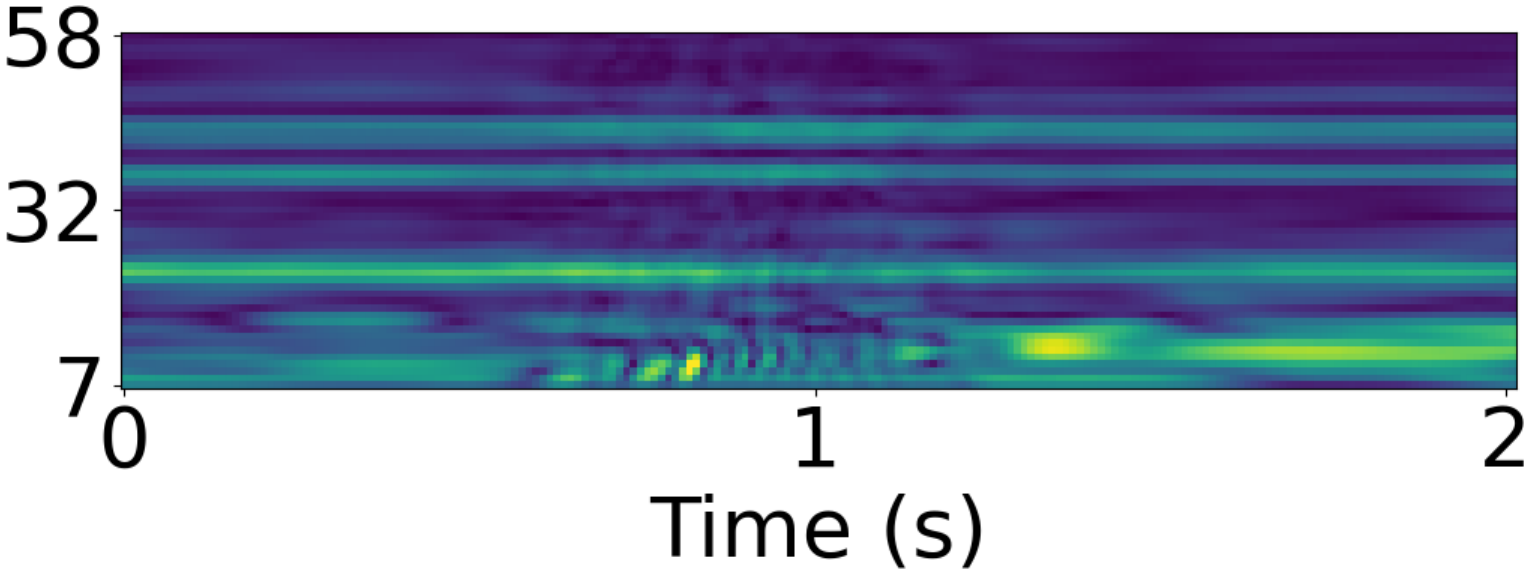}\\[-4pt]
        \includegraphics[width=\linewidth]{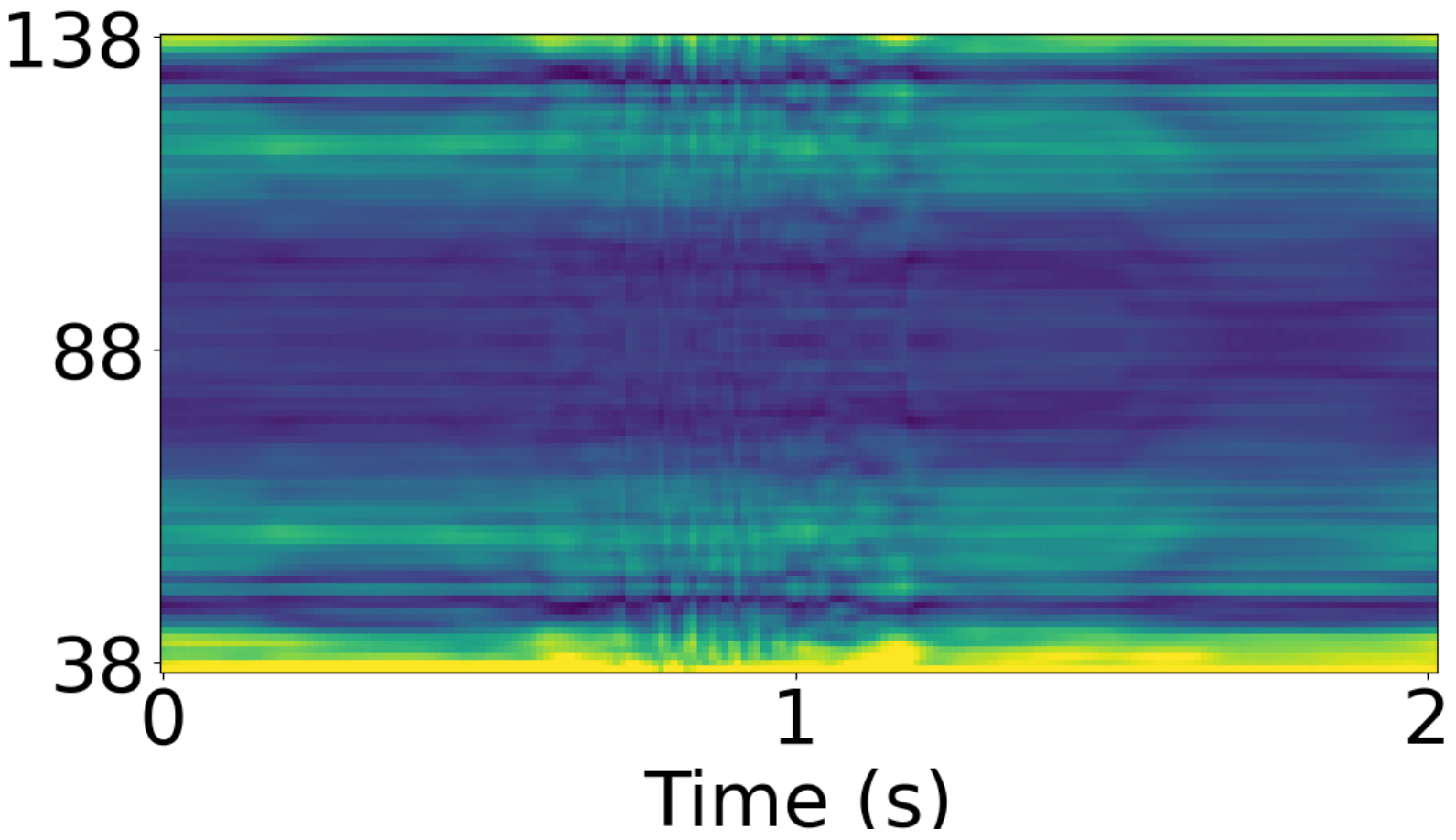}
        \\[1pt]
        \footnotesize\emph{Smoke}
    \end{minipage}
    \hspace{-2pt}
    \begin{minipage}[b]{0.13\textwidth}
        \centering
        \includegraphics[width=\linewidth]{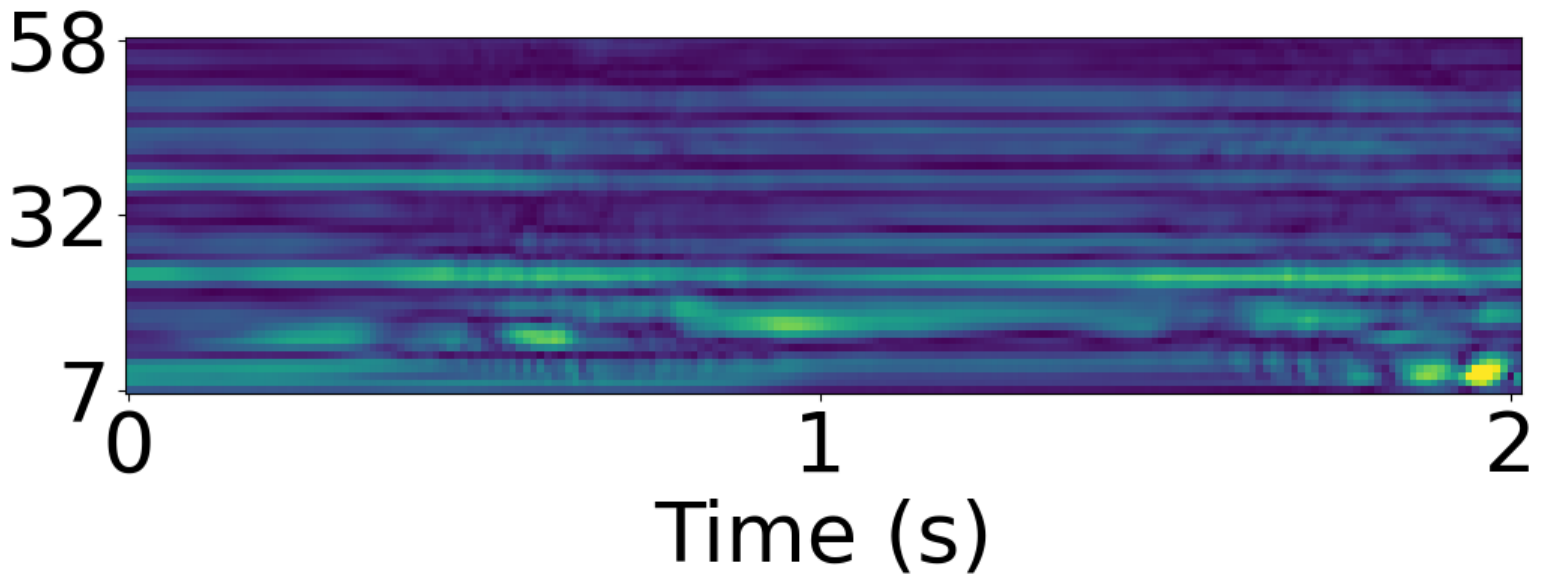}\\[-4pt]
        \includegraphics[width=\linewidth]{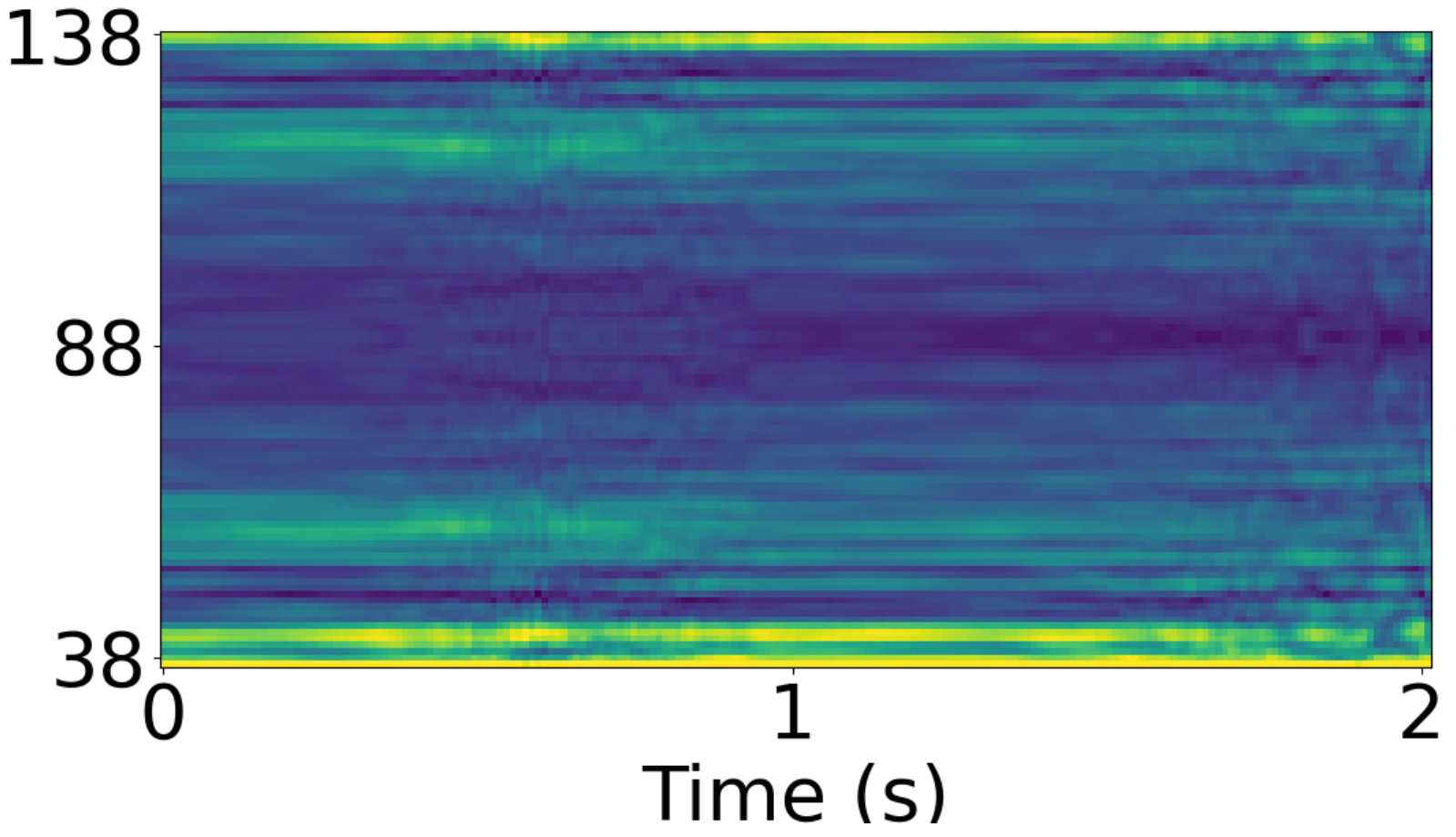}
        \\[1pt]
        \footnotesize\emph{Drink}
    \end{minipage}
    \hspace{-2pt}
    \begin{minipage}[b]{0.13\textwidth}
        \centering
        \includegraphics[width=\linewidth]{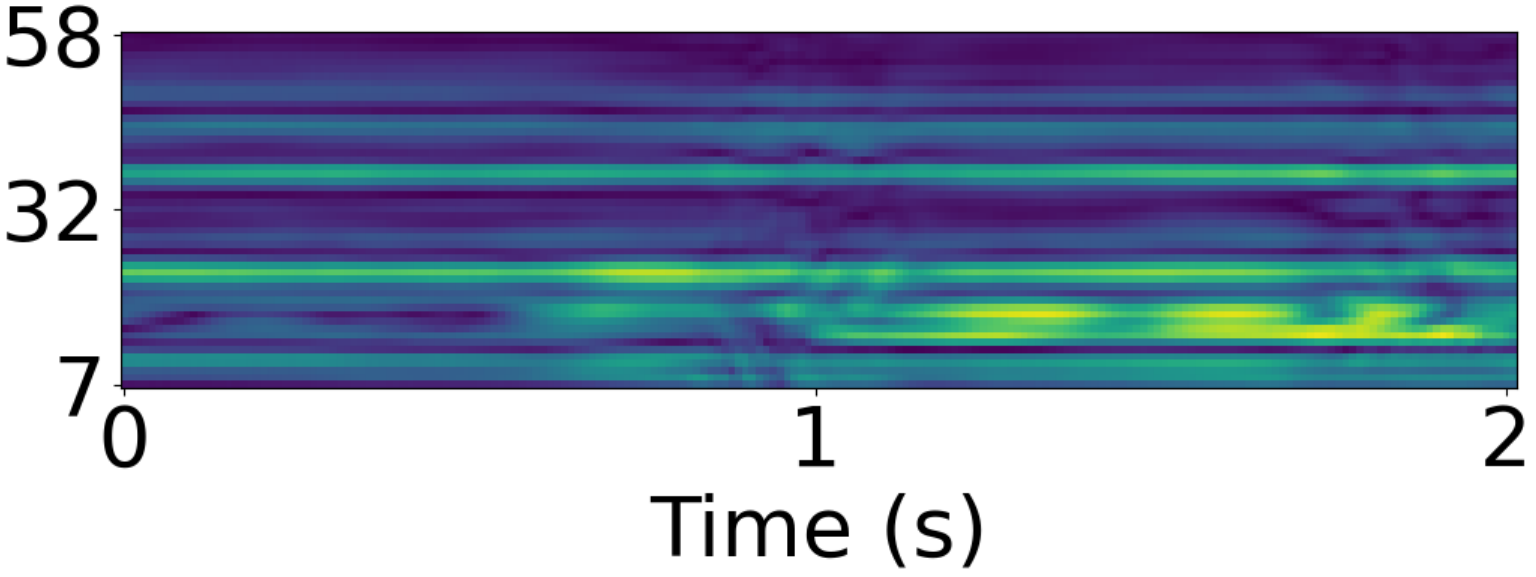}\\[-4pt]
        \includegraphics[width=\linewidth]{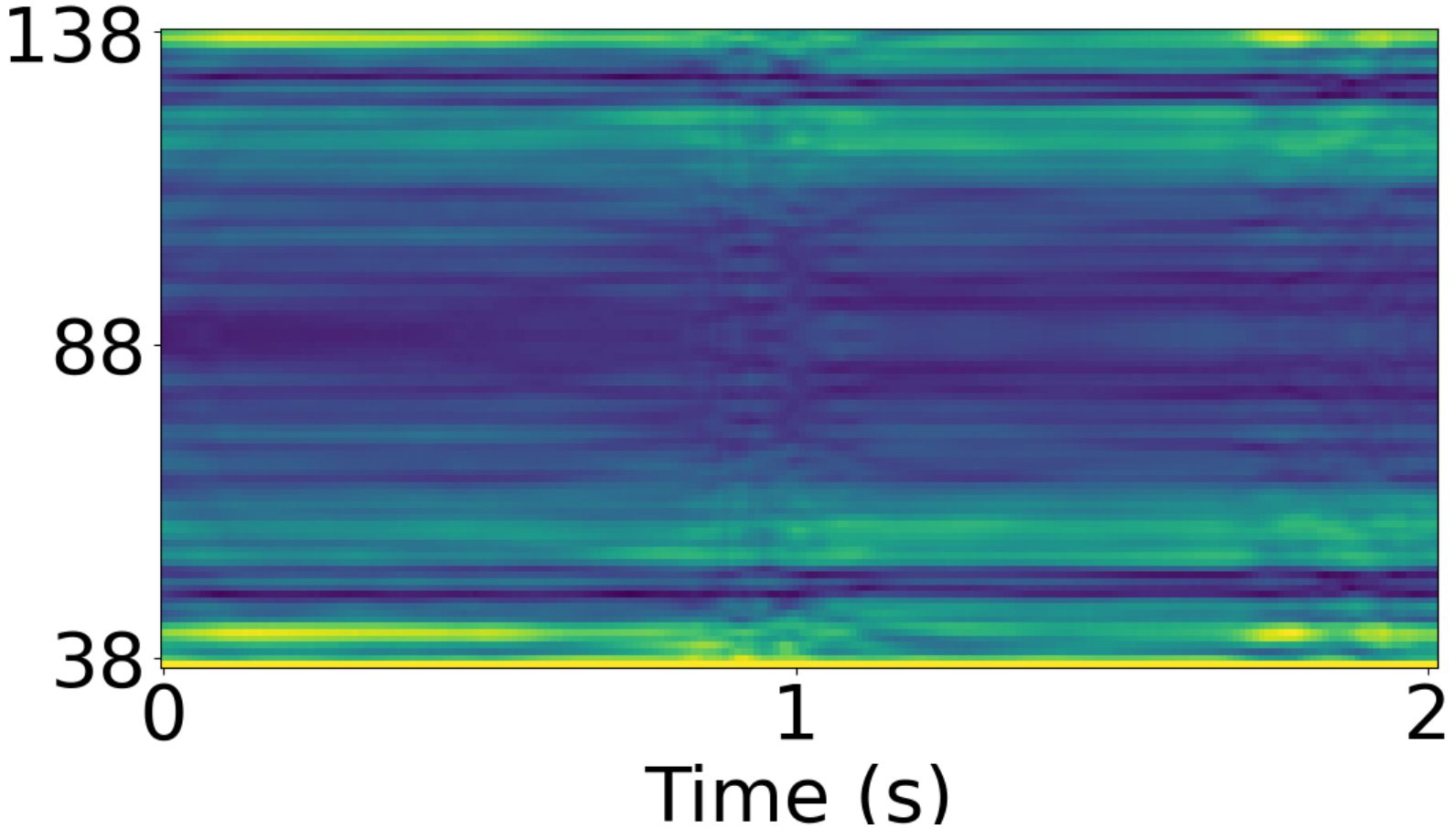}
        \\[1pt]
        \footnotesize\emph{Panel}
    \end{minipage}
    \hspace{-2pt}
    \begin{minipage}[b]{0.13\textwidth}
        \centering
        \includegraphics[width=\linewidth]{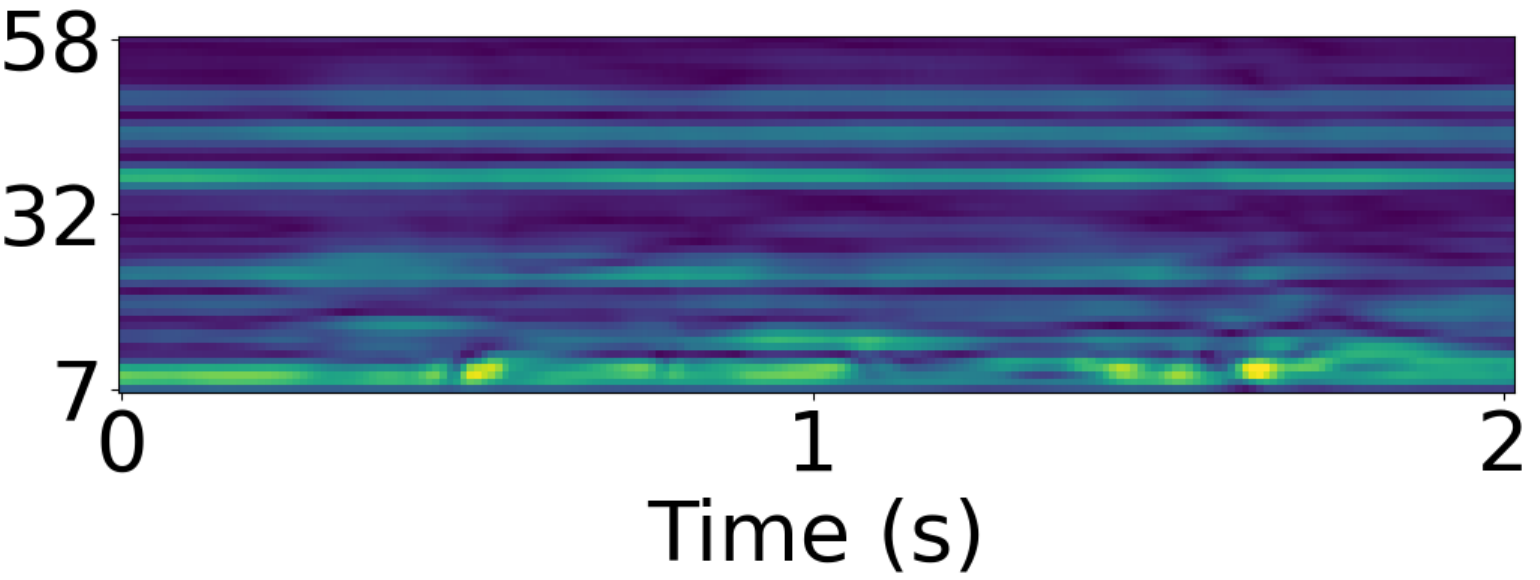}\\[-4pt]
        \includegraphics[width=\linewidth]{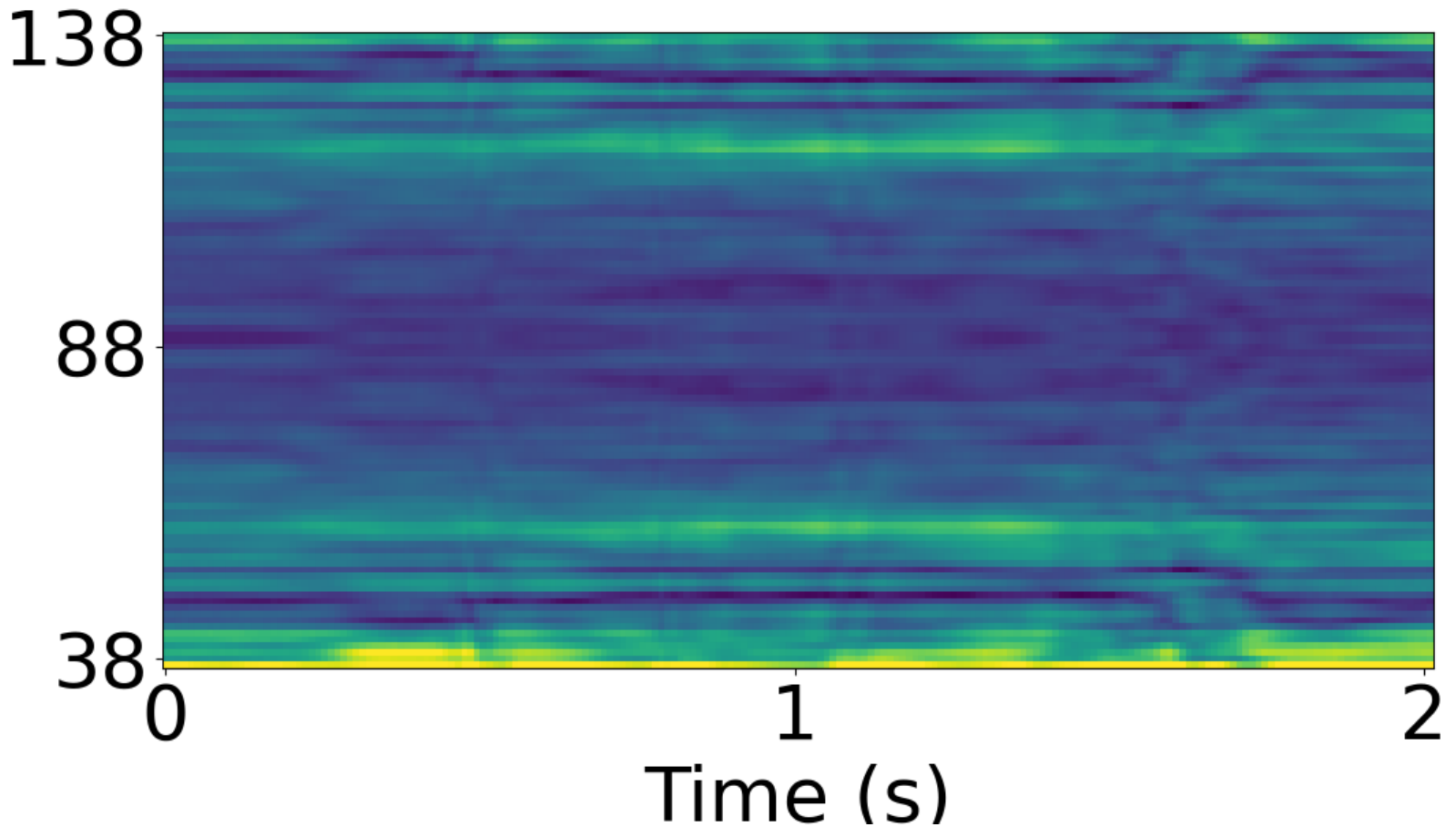}
        \\[1pt]
        \footnotesize\emph{Phone}
    \end{minipage}

    \caption{Examples of range and frequency data representations for 7 activities.}
    \label{fig:datarepre}
    \vspace{-10pt}
\end{figure*}

Through dataset manipulation, we obtain range and frequency data for each activity. We exemplify the representations with prominent features of 7 activities in Fig.~\ref{fig:datarepre}. Each representation exhibits distinct patterns or shapes unique to each activity, which are difficult to differentiate with the human eye alone, thus necessitating the use of learning methods for distinction. Someone may consider a vision-task approach appropriate due to the 2D nature of the manipulated data. Conversely, others might view the dataset as suitable for sequential prediction due to its time-series information. Thus, we describe the various approaches in Section~\ref{sec:benchmark} and evaluate the most appropriate method for UWB data in Section~\ref{sec:evaluation}.

\subsection{Dataset Comparison}
In UWB DAR, the existence of standard open dataset is significant to accelerate the studies. In~\cite{RaDA}, they provide the first open dataset, RaDA, for UWB DAR including comprehensive 6 activities such as driving, autopilot, sleeping, driving \& smartphone utilization, smartphone utilization, and talking to a passenger. However, it is not enough to serve as \emph{de facto} standard open dataset compared to our dataset, \ALERT. The detailed differences between our dataset and RaDA are as follows and summarized in Table~\ref{table:diff}. \re{Compared to RaDA, \ALERT dataset \textbf{is collected in real driving} with an \textbf{air-vent mounting} that maintains a stable sensing geometry without interfering with the driver's view. In addition, \ALERT provides \textbf{longer samples (5 s)} and \textbf{more fine-grained activity labels} designed to reflect natural in-vehicle behaviors with fewer restrictive execution rules. Moreover, \ALERT offers \textbf{both range-time and frequency-time representations}, enabling flexible input manipulations (e.g., cropping and window/size variations) for systematic evaluations. Finally, \ALERT includes a broader set of driving-induced motion conditions (e.g., stop-and-go and road-surface variations), which helps assess robustness under realistic vibration and traffic dynamics.}

\begin{table}[t!]
	\caption{\rev{Dataset comparison of \ALERT and RaDA}}
        \centering
	\label{table:diff}
	\setlength{\tabcolsep}{3pt}
	\begin{tabular}{|p{2cm}|p{2.9cm}|p{2.9cm}|}
		\hline
		\textbf{Features}& 
		\textbf{ALERT}&
            \textbf{RaDA~\cite{RaDA}}\\ 
		\hline
		\# of activities&
		7&
            6\\
		\hline
		\# of subjects&
		9&
            10\\
		\hline
            Activity restrictions&
		Lenient&
            Moderate\\
		\hline
            \# of samples&
		10,220 (5 s per sample)&
            10,406 (1 s per sample)\\
		\hline
            Dataset&
		Range-time \& freq.-time&
            Range-Doppler\\
		\hline
            Driving \newline environment&
		\emph{Real driving}&
            Simulated driving\\
		\hline
            Sensor \newline position&
		Air vent&
            Front of driver \newline (visual obstruction)\\
		\hline
	\end{tabular}
 \vspace{-10pt}
\end{table}

\subsection{Dataset and Code Availability}
We publicly release the \ALERT dataset to support further research and experimentation. We also provide the dataset processing code on GitHub, along with guidelines for using it with the \ALERT dataset. 
The URLs for the dataset and codes are as follows, \ALERT dataset: \url{https://github.com/ALERTdataset/ALERT.git}\\

\section{Benchmkaring Design}\label{sec:benchmark}
\begin{figure*}[t]
    \centering
    \includegraphics[width=15cm, height = 6.5cm]{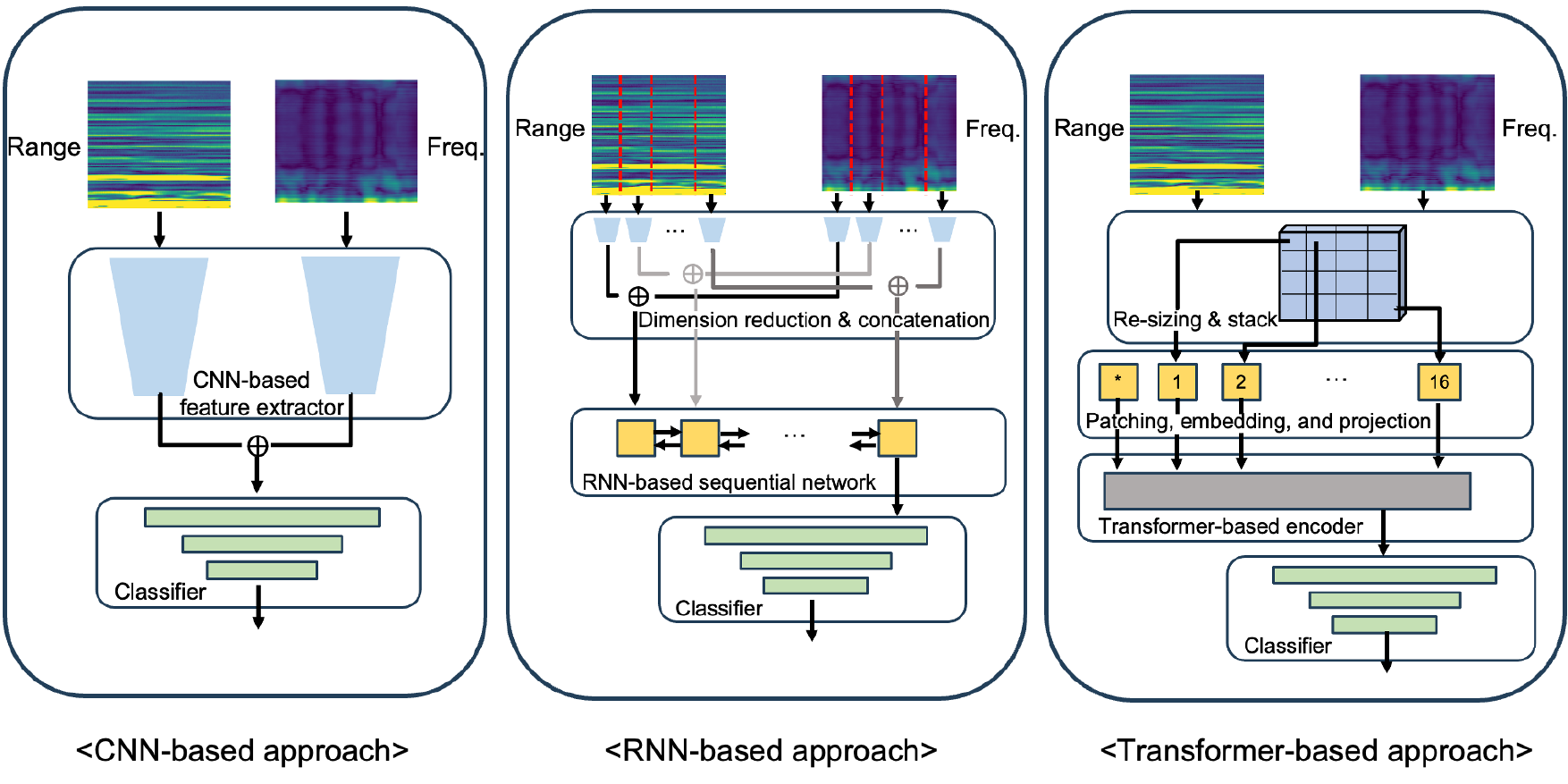}
    \vspace{-10pt}
    \caption{\rev{Design of learning algorithms for benchmarking.}}
    \label{fig:design}
    \vspace{-10pt}
\end{figure*}
To provide benchmarking performance, we design various learning models including CNN-based~\cite{googlenet, resnet, densenet, mobilenet}, RNN-based~\cite{maitre2020fall, maitre2021recognizing, noori2021ultra}, and transformer-based approaches~\cite{deit, vit}. Additionally, we incorporate a multi-view learning~(MVL)-based domain fusion method into all designs to demonstrate the scalability of using the \ALERT dataset and to achieve better performance in DAR.

To further explore the most effective utilization of the \ALERT dataset, we conduct various experiments through detailed observation. For instance, we investigate the observation time required to accurately identify DAR activities and evaluate the effects of multipath environments and frequency ranges across 8 different algorithms.

The various experiments allow us to identify optimal parameters, trends, and trade-offs specific to each algorithm. We present the results of these experiments in Section~\ref{sec:evaluation}, providing comprehensive insights. Since the \ALERT dataset can be customized to meet users' specific needs, sharing these experimental findings is crucial for guiding its optimal application.

\subsection{Domain Fusion}
\re{In IR-UWB sensing, range-time captures fine-grained spatial motion but often provides insufficient cues to separate visually similar driver actions (e.g., Smoke vs.\ Drink). Frequency-time, obtained via Fourier transform, emphasizes Doppler-driven velocity patterns and is typically less sensitive to multipath reflections, yielding cleaner signatures for motion dynamics. We therefore fuse both domains to leverage complementary cues and improve discrimination of confusing labels.}

In domain fusion, we leverage the various features obtained from multiple domains and concatenate each feature to classify the object. Similarly, some works~\cite{HARSANET, SleepPoseNet} apply the MVL-based domain fusion method to the UWB HAR for better performance. They transform the same UWB data into both range domain and frequency domain components, separately process each branch's range and frequency inputs, and then concatenate to classify the activities. The details of the domain fusion design are provided in the following, and the evaluation results using both single-domain datasets and domain fusion datasets are presented in Section~\ref{sec:evaluation}.

\subsection{Design for CNN-based Approach}
We benchmark the performance of various CNN architectures including GoogLeNet~\cite{googlenet}, ResNet~\cite{resnet}, DenseNet~\cite{densenet}, and MobileNet~\cite{mobilenet}. The aforementioned algorithms are CNN-based feature extractors as depicted in Fig.~\ref{fig:design}. The range and frequency data are fed into the CNN-based feature extractor. After processing through the feature extractor, features for range and frequency are concatenated and then inputted into a classifier to recognize the activity. The classifier comprises three fully connected layers, with various dimensions tailored to the specific algorithm employed.

\subsection{Design for RNN-based Approach}
\rev{To design the RNN-based approach, we follow the conceptual framework introduced in \cite{maitre2020fall, maitre2021recognizing, noori2021ultra}. Although the ALERT dataset includes sufficiently long sequences for temporal modeling, we do not adopt xLSTM~\cite{xlstm} for benchmarking. This is because xLSTM processes individual UWB pulses sequentially, and each pulse contains only limited information. In UWB radar, meaningful activity patterns emerge only when multiple pulses are accumulated over time. To address this, we adopt the CNN+LSTM architecture that groups several pulses into short snippets, extracts spatial features from each snippet using a CNN, and then models the temporal dependencies between these snippets using an LSTM.}

As depicted in Fig.~\ref{fig:design}, we split both range and frequency data into 10 snippets and use ResNet to extract features for refined input representation and dimension reduction. Subsequently, each feature extracted from the snippets is fed into an RNN-based sequential network, specifically implemented as a bidirectional LSTM model. The model outputs results from the LSTM at the last time step, which are then passed to the classifier. We set the hidden state size to 1024, considering the feature size, and utilize only one LSTM layer.

\subsection{Design for Transformer-based Approach}\label{sec:transformer}
For a Transformer-based approach, we utilize Vision Transformer~(ViT) model~\cite{vit} and Data-efficient Image Transformer~(DeiT) model~\cite{deit}, which is a variant of ViT.
In the transformer-based approach, we adopt the early fusion method, as illustrated in Fig.~\ref{fig:design}. Since transformer-based models are computationally intensive, applying the late fusion method in a real-time DAR system is challenging. Late fusion requires two independent ViT models for each input, followed by concatenating the features extracted from their outputs, which significantly increases computational overhead.

Although late fusion achieves slightly higher accuracy compared to early fusion, we choose the early fusion method to balance both accuracy and DAR system responsiveness. The comparison between early fusion and late fusion is detailed in Section~\ref{sec:evaluation}.

Consequently, we resize each dataset to an image size of $224 \times 224$ and assemble them into a two-channel image. The resized image is then divided into $14 \times 14$ patches. Each patch is projected into an embedding vector and combined with a positional embedding vector. These combined vectors are subsequently fed into the transformer encoder, enabling the classification of tokens derived from the transformer's output.

\section{Input-Size-Agnostic Vision Transformer (ISA-ViT)}\label{sec:ISA-ViT}

Transferring the remarkable capabilities of ViTs to the UWB domain is critical for effectively classifying 2D image-like UWB datasets. However, applying a ViT pre-trained on ImageNet~\cite{imagenet} to UWB data presents a significant challenge. 
\re{Beyond the input-size mismatch, transferring image-pretrained ViTs to UWB radar also involves a domain gap (i.e., different signal statistics and physical meaning), which can limit “plug-and-play” transfer and motivates fine-tuning on target UWB data.}
In image processing, ViTs typically operate on an input size of $224 \times 224$, divided into $16 \times 16$ patches, resulting in 196 patches (14 along both the height and width of the image). Each patch is transformed into a patch embedding vector and combined with \rev{the positional embedding vector~(PEV)} containing relative spatial information, forming the input to the transformer encoder. \re{This 14×14 token grid is an empirically established standard in pre-trained ViT model~\cite{vit} and is tightly coupled with the learned positional embeddings, making its preservation important for stable transfer learning.}

However, in practice, the input size of UWB data cannot always be guaranteed to match the fixed size of $224 \times 224$. \re{This
discrepancy makes it infeasible to directly apply pre-trained
PEVs to UWB data without altering the original token grid, which may degrade the effectiveness of transfer learning.} 
\re{Moreover, naïve up/down-sampling to fit 224×224 can discard signal details and weaken spatiotemporal structures, especially when the original input is much larger.}
To tackle this critical issue, we propose the \mine, a novel approach specifically designed to adapt ViTs for UWB data and introduce the details of \mine as follows.

\subsection{Preliminary Studies}\label{sec:preliminary}
\begin{figure}[t]
\centering
\subfigure[Image data]{
    \centering
    \includegraphics[width=4cm, height = 4cm]{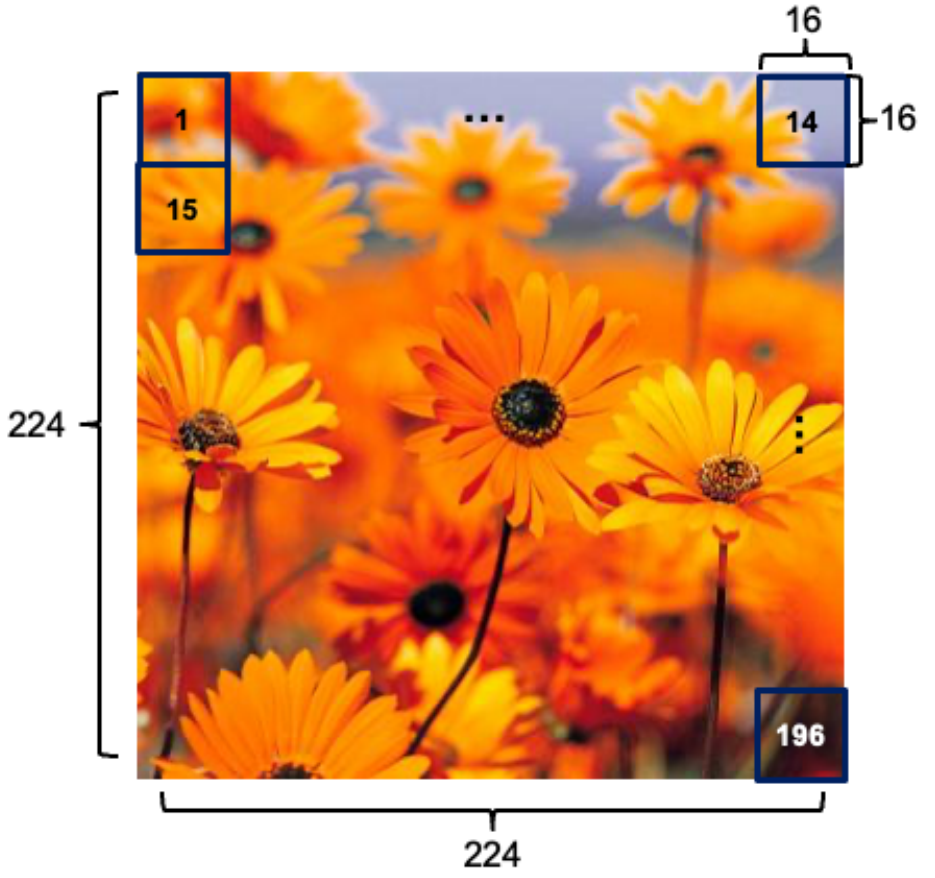}
    \label{fig:image}
}
    \centering
    \vspace{-10pt}
\subfigure[UWB data]{
    \centering
    \includegraphics[width=8cm, height = 3.5cm]{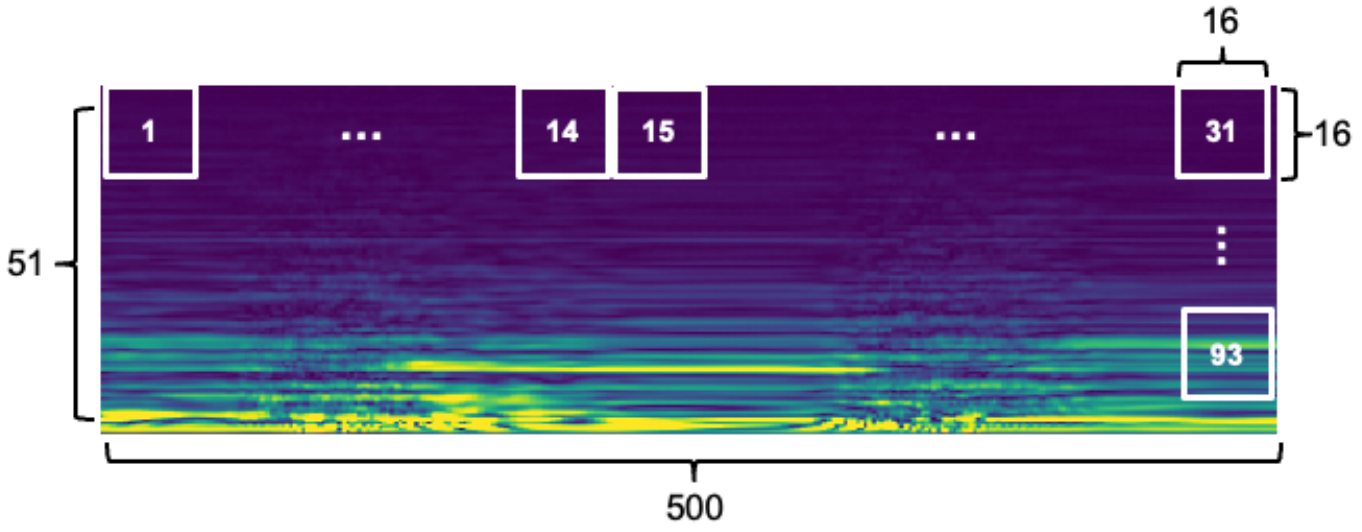}
    \label{fig:uwb}
}
    \caption{\rev{Application of PEVs according to input of different domain data.}}
    \label{fig:different_domain}
    \vspace{-10pt}
\end{figure}

The PEV is a significant parameter for capturing spatial information from the pre-trained ViT. Since the UWB data has a different size compared to image data (i.e., 224$\times$224), applying the PEV to the UWB data requires a different way compared to image processing. As shown in Fig.~\ref{fig:image}, the PEVs of the 14th and 15th patches are trained to be dissimilar in ViT image tasks because of their distant spatial locations. However, if we apply these pre-trained PEVs to UWB data with different dimensions, as shown in Fig.~\ref{fig:uwb}, the 14th and 15th patches are treated as if they are far apart, despite being adjacent in reality. This mismatch can significantly degrade performance, highlighting the importance of appropriately adapting the pre-trained PEVs.

To handle non-image inputs sizes, we can consider two approaches: 1) \textbf{maintaining the pre-trained 14$\times$14 sequence of PEVs}, or 2) \textbf{manipulating a different number and sequence of PEVs} from the pre-trained ViT for image tasks, as illustrated in Fig.~\ref{fig:uwb}. 

The first approach for \textbf{maintaining pre-trained 14$\times$14 PEVs} involves two methods: 1) \emph{resizing the input} or 2) \emph{adjusting the shape of the patch}. The \emph{resizing the input} method exploits up \& down-sampling techniques to make it 244$\times$244 in size. Alternatively, we can \emph{adjust the shape of the patch} by keeping the input data unchanged and applying patches of a different shape, rather than using the 16$\times$16 size typically used for images. We divide the height and width of the input data by 14 to determine the patch size for both dimensions, and we add zero-padding if necessary.
The common goal of both methods is utilizing the pre-trained 14$\times$14 sequence of PEVs.

We can refer to \cite{ast} to \textbf{manipulate the PEVs}, as mentioned above the second approach. They calculate the number of patches (using 16$\times$16 patches) along the height and width of the input data. If the number of patches in the side exceeds 14, they apply interpolation to the PEVs for height; otherwise, they use truncation. We implement \emph{manipulating PEVs} by following the conceptual approach used in \cite{ast}.

We compare each method to determine which one is the most suitable for using pre-trained PEVs. For this evaluation, we use two UWB open datasets for DAR: the \ALERT dataset and the \emph{RaDA} dataset~\cite{RaDA}. Both datasets include 6--7 activities for recognizing distracted driving behaviors, such as driving, nodding, using a smartphone, etc. The \ALERT dataset has input data samples of size 51$\times$500, while the \emph{RaDA} dataset has samples of size 1024$\times$24.

\begin{table}[t]
\centering
\caption{Accuracy according to handling method for input of UWB data}
\vspace{-10pt}
\begin{tabular}{|c|c|c|}
\hline
\diagbox{Method}{Dataset} & \ALERT & \emph{RaDA}~\cite{RaDA}\\
\hline
Resizing input & \textbf{52.20}\% & \textbf{56.10}\% \\
\hline
Adjusting patch shape & 43.55\% & 53.92\% \\
\hline
Manipulating PEVs~\cite{ast} & 36.74\% & 49.25\% \\
\hline
\end{tabular}
\label{table1}
\vspace{-10pt}
\end{table}

Table~\ref{table1} shows the average accuracy of each method for the two datasets. The methods for maintaining the pre-trained 14$\times$14 sequence of PEVs, \emph{resizing input} and \emph{adjusting patch shape}, demonstrate better performance than \emph{manipulating PEVs}.

The patch sizes in the \emph{adjusting patch shape} method are constrained to highly limited and unconventional shapes, such as wide-long or bottom-long configurations. The patches are set to 4$\times$36 in the \ALERT dataset and 74$\times$2 in the \emph{RaDA} dataset. These extreme rectangular patch shapes do not contribute positively to model performance, as they fail to capture sufficient information effectively. 

The \emph{manipulating PEVs} method underperforms due to the unique characteristics of UWB data, where the \ALERT and \emph{RaDA} datasets contain samples with wide or bottom-long shapes. As a result, both cutting and interpolation are applied to the pre-trained PEVs. For both datasets, we interpolate the PEVs for the width by more than twice and 4.5 times, respectively. Excessive interpolation may result in the loss of fine-grained spatial information or lead to positional embedding saturation, reducing the meaningfulness of the PEVs. Consequently, the \emph{manipulating PEVs} struggles to adapt effectively to the shape of UWB data.

Through the preliminary studies, we observe the importance of \textbf{preserving the pre-trained $14\times14$ PEV sequence} for non-image input sizes. 
\re{These results highlight that better performance is achieved when the pre-trained PEVs are reused in their original structure while the input is adapted in an information-preserving manner. This insight directly motivates our design goal in ISA-ViT: to enable effective utilization of pre-trained PEVs for highly asymmetric UWB inputs by minimizing resizing-induced information loss and maintaining the pre-trained PEV sequence.}
In the following section, we introduce \mine, a detailed method for maintaining the pre-trained sequence of PEVs, designed to effectively apply PEVs to UWB data.

\subsection{Proposed Scheme}
Our goal is to enhance the performance of DAR using IR-UWB. To achieve this, we need to design a model that effectively utilizes UWB data and develop a method that can further improve DAR performance. Therefore, we introduce \mine and the domain fusion method, as shown in Fig.~\ref{fig:proposed}. The \mine enables the application of UWB data to ViT regardless of data size, while domain fusion improves DAR performance by combining information from multiple domains, resulting in better classification.

\subsection{ISA-ViT}
\begin{figure}[t!]
\centering
\subfigure[The design of \emph{ISA-ViT}]{
    \centering
    \includegraphics[width=8.5cm, height = 6cm]{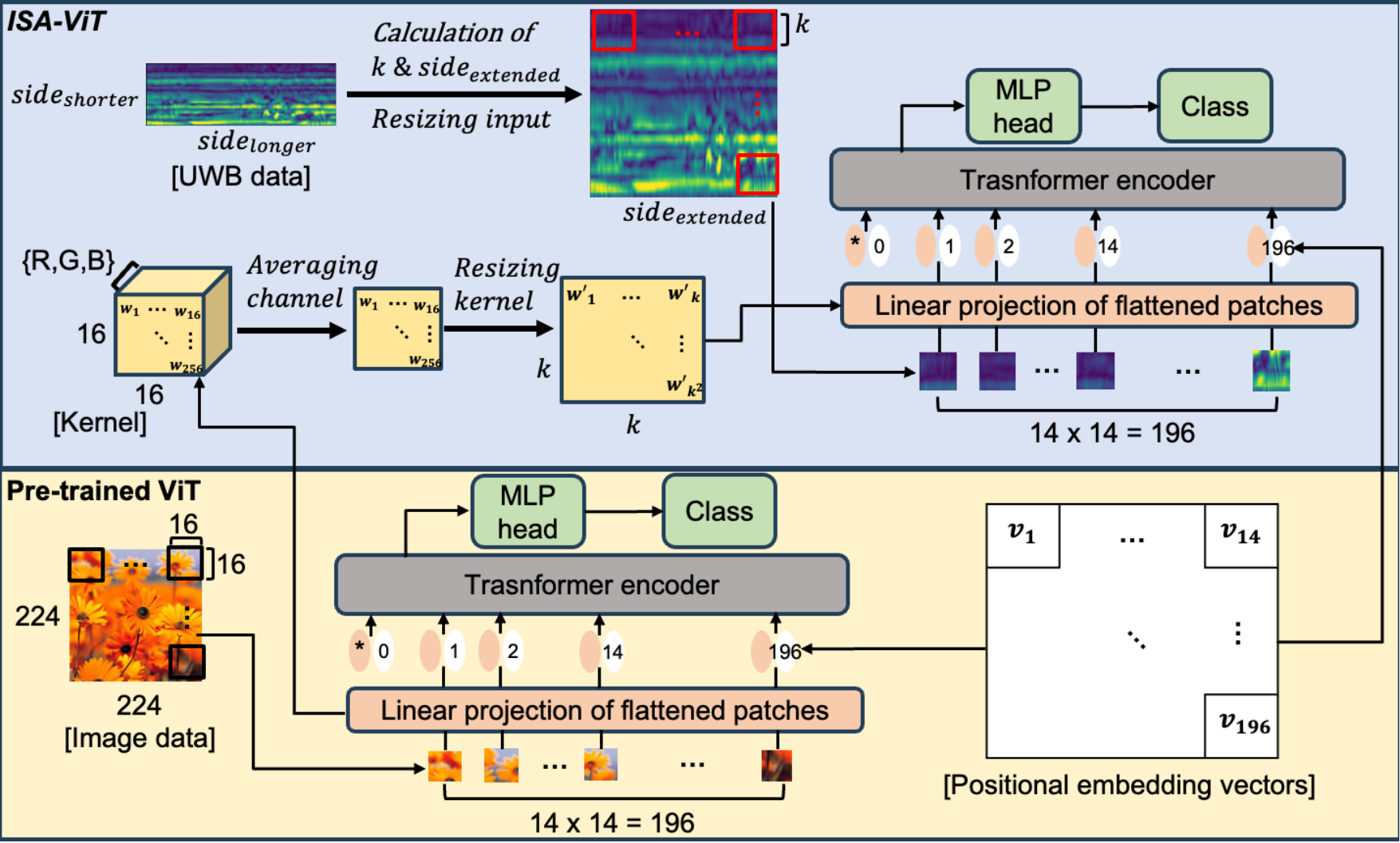}
    \label{fig:ISA-ViT}
}
\hfill
\subfigure[The design of domain fusion]{
    \centering
    \includegraphics[width=6.5cm, height = 5cm]{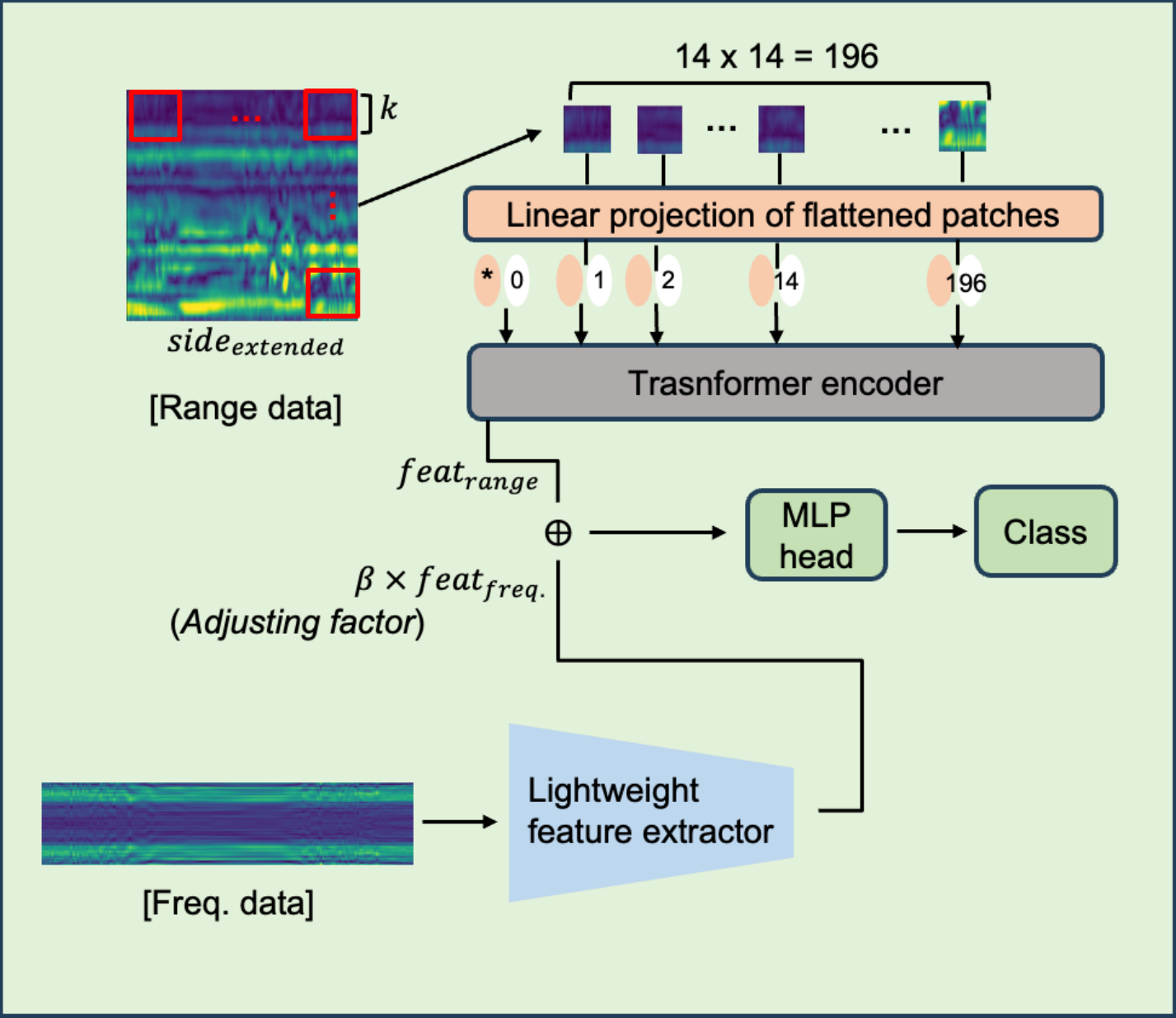}
    \label{fig:domain_fusion_method}
}
\vspace{-10pt}
    \caption{Overview of the proposed scheme.}
    \label{fig:proposed}
    \vspace{-10pt}
\end{figure}
In the previous section, we verified that maintaining the pre-trained $14\times14$ sequence of PEVs yields better performance than manipulating PEVs without considering the input size. However, using simple resizing (via up \& down-sampling) may result in severe information loss when the input size is much larger than the image size (i.e., 224$\times$224). Therefore, we designed \mine with no information loss from the input data, as described in Fig.~\ref{fig:ISA-ViT}.

Our conceptual approach is as follows: UWB data commonly has a rectangular shape rather than a square (though it can also be square-shaped). First, we extend the shorter side of the input to match the size of the longer side, without any loss of information. Afterward, we calculate the patch size to divide the extended input into 14$\times$14 patches. One important consideration is that the sides of the extended input may not be divisible by 14. Therefore, we use the following formulas to determine the patch size and apply resizing to adjust the input accordingly.

\begin{equation}
k = \min \{ k \in \mathbb{Z} \mid 14 \times k \geq side_{longer} \},
\end{equation}
\begin{equation}
side_{extended} = 14 \times k,
\end{equation}
where $k$ is the size of the patch, $side_{longer}$ is the size of the longer side of the input, and $side_{extended}$ is the size of the side to be extended. This process ensures that there is no loss of information.

Moreover, we must consider one additional factor, which is patch embedding. After dividing the extended input into patches of size $k$, each patch is transformed into a patch embedding vector through linear projection using a simple CNN layer. Typically, in the image domain, a 16$\times$16 pre-trained kernel is used, matching the patch size. In the UWB domain, we need to align not only the CNN kernel size but also the pre-trained weights of the kernel. The pre-trained CNN kernel has a 16$\times$16 size with RGB channels, but since UWB data has only one channel, we average the weights of the RGB channels into one channel.

After that, we adjust the pre-trained CNN kernel size and weights to match the size of $k \times k$. We simply use interpolation (when $k>16$) or average pooling (when $k<16$) to adjust the kernel weights. Through this process, we finally combine the patch embedding vectors with the pre-trained PEVs and pass them to the transformer encoder. Fig.~\ref{fig:ISA-ViT} and Algorithm~\ref{alg.1} present the summarized method of \mine.
\begin{algorithm}[t]
\scriptsize
\caption{Adjustment of the size of UWB data, patch, and kernel of \mine}\label{alg.1}
\begin{algorithmic}[1]
\Statex \textbf{Input:} $data_{UWB}$
\Statex \textbf{Output:} Patch embedding vectors
\Statex \textbf{Initialize:} $num_{patches}\!\leftarrow\!14$, $k_{original}\!\leftarrow\!16$

\State \textbf{Input size and patch-size calculation:}
\Statex $(side_{shorter},\,side_{longer}) \leftarrow size(data_{UWB})$
\Statex $k \leftarrow \left\lceil \frac{side_{longer}}{num_{patches}} \right\rceil$;\quad
$side_{extended} \leftarrow num_{patches}\times k$

\State \textbf{Resizing input:}
\Statex $data_{UWB} \leftarrow resize\!\left(data_{UWB},\,(side_{shorter},side_{longer})\rightarrow\right.$
\Statex $\left.\qquad\qquad\qquad\qquad(side_{extended},side_{extended})\right)$

\State \textbf{Averaging channel and resizing kernel:}
\Statex $weight_{kernel}(P_{original}) \leftarrow channel\_avg\!\left(weight_{kernel}(P_{original})\right)$
\Statex $weight_{kernel}(P_k) \leftarrow resize\!\left(weight_{kernel}(P_{original}),\,k_{original}\rightarrow k\right)$

\State \textbf{Linear projection:}
\Statex \textbf{return} $CNN\!\left(data_{UWB},\,weight_{kernel}(P_k)\right)$
\end{algorithmic}
\end{algorithm}

\subsection{Domain Fusion of ISA-ViT}
\re{A straightforward approach for domain fusion is to use two parallel ViTs to extract features from the range and frequency domains. However, directly fusing two ViT backbones results in a heavy model, as discussed in Sec.~\ref{sec:transformer}. To balance accuracy and responsiveness, we propose a lightweight late-fusion scheme (Fig.~\ref{fig:domain_fusion_method}): we apply \mine to the range data, while using a lightweight feature extractor for the frequency data. The range input is partitioned into $k\times k$ patches (i.e., $14\times14$), projected into patch embeddings, combined with pre-trained PEVs, and encoded by the transformer to obtain a range-domain representation. In contrast, the frequency input is fed into the lightweight extractor without resizing, which further reduces computation. This design improves efficiency while prioritizing the more informative range-domain cues.}

\re{We emphasize the range domain because it provides essential spatial context for behavior understanding. Range information helps distinguish actions with similar motion patterns but different spatial configurations (e.g., holding a phone in front of the body versus bringing a cigarette to the mouth). In contrast, frequency information mainly reflects motion dynamics (e.g., speed changes and movement direction) and is often insufficient alone; for example, with similar motion speeds, frequency data cannot reliably differentiate dozing off from attentive driving. Therefore, we treat frequency features as complementary cues and assign greater weight to range-domain features in fusion.}

We primarily utilize features extracted from the range domain, with frequency domain data providing supplementary information. Instead of concatenating frequency features at the same ratio as range features, we apply an adjusting factor $\beta$ to balance the frequency domain features before concatenation. 
\re{The contribution of frequency-domain features is not known \emph{a priori}; therefore, we scale the frequency-domain feature by $\beta$ prior to concatenation:}
\re{\begin{equation}
\mathbf{z} = [\mathbf{feat}_{range}; \beta \times \mathbf{feat}_{freq.}],
\end{equation}}
\re{where $\mathbf{feat}_{range}$ and $\mathbf{feat}_{freq.}$ denote the range- and frequency-domain feature vectors, respectively.}
The rationale for this approach is that range and frequency data do not always complement each other positively in domain fusion. This method ensures that the contribution from the frequency domain is balanced and does not overpower the more informative range domain features. 
\re{We treat $\beta$ as a trainable scalar parameter and optimize it jointly with the network parameters via backpropagation during training.}

Our domain fusion approach enhances DAR performance by effectively combining the unique characteristics of the range and frequency domains, addressing the limitations of classification performance when using the single domain. Detailed results from our proposed scheme are presented in the Section~\ref{sec:evaluation}.

\section{Evaluations}\label{sec:evaluation}
\subsection{Experimental Settings}

\noindent
\textbf{Benchmarking.}
\rev{We leverage a system running Ubuntu 22.04.5 LTS, equipped with an Intel® Core™ i9-12900K CPU and a single NVIDIA GeForce RTX A6000 GPU with 49 GB of memory in our experiments. The software environment included Python 3.10.12 and CUDA 11.5.119 with CUDA support.}

We conduct extensive experiments using the \ALERT dataset, which comprises 7 activities with 10,220 samples collected from 9 participants. The observation window is set to 5 seconds, with 8–58 range bins and 89–178 frequency bins in the range and frequency domain data.

Using the \ALERT dataset, we benchmark several learning algorithms, including GoogLeNet~\cite{googlenet}, ResNet~\cite{resnet}, DenseNet~\cite{densenet}, MobileNet~\cite{mobilenet}, an RNN-based model~\cite{maitre2020fall, maitre2021recognizing, noori2021ultra} (denoted as RNN), and transformer-based models such as DeiT~\cite{deit} and ViT~\cite{vit}. Each model is implemented as described in Section~\ref{sec:benchmark}.

\re{To ensure adequate fine-tuning, we train the models for 30 epochs, using a learning rate between 0.00001 and 0.0001 to achieve convergence of the training loss for each algorithm. Cross-entropy loss is employed as the loss function, and the Adam optimizer is used for all models. To ensure a fair comparison, we fix the train and test batch sizes to 25 for all models and set the adaptation batch size to 5. In addition, features extracted from each backbone are fed into the same classifier head for classification, and no weight decay, dropout, or additional regularization is applied.}

\noindent
\textbf{Evaluation of \mine.}
We thoroughly evaluate the performance of DAR using \mine and the proposed domain fusion method.
To conduct extensive experiments, we employ the \ALERT dataset and \emph{RaDA} dataset~\cite{RaDA} which are open dataset leveraging IR-UWB to detect driver distraction activities. 

The \ALERT dataset not only offers comprehensive data recorded in real driving environments but also provides manipulable data tailored to users’ specific needs. We utilize both range-time and frequency-time domain data from the \emph{ALERT} dataset to capture detailed features across different modalities.

The \emph{RaDA} dataset comprises 10,406 samples across 6 activities (i.e., driving, autopilot, sleeping, driving while using a smartphone, smartphone use, and talking to a passenger) in a simulated environment. Since the \emph{RaDA} dataset provides only single domain data, we cannot apply it to the domain fusion method and instead primarily use it to evaluate the impact of \mine.

For model training, we strictly separate the training and testing datasets; the testing dataset consists of data from drivers who were not seen during the training phase. \re{In addition, we employ a subject-wise leave-one-subject-out protocol: in each fold, all samples/segments from one driver are held out as the test set, and the remaining drivers are used for training. Thus, no driver appears in both training and testing, ensuring strict subject independence and preventing leakage across splits. We also note that we do not apply any data-dependent preprocessing such as normalization or standardization.}
We employ a learning rate as 0.00001 to ensure convergence the training loss. The base model is a ViT pre-trained on ImageNet~\cite{imagenet}, which uses embedding vectors of size 1,024. We fine-tune the model over 30 epochs, optimizing it with cross-entropy loss and the Adam optimizer to ensure robust performance.

\begin{figure*}[t!]
\centering
    \subfigure[\re{Average accuracy}]{
    \includegraphics[width=0.45\textwidth, height = 4cm]{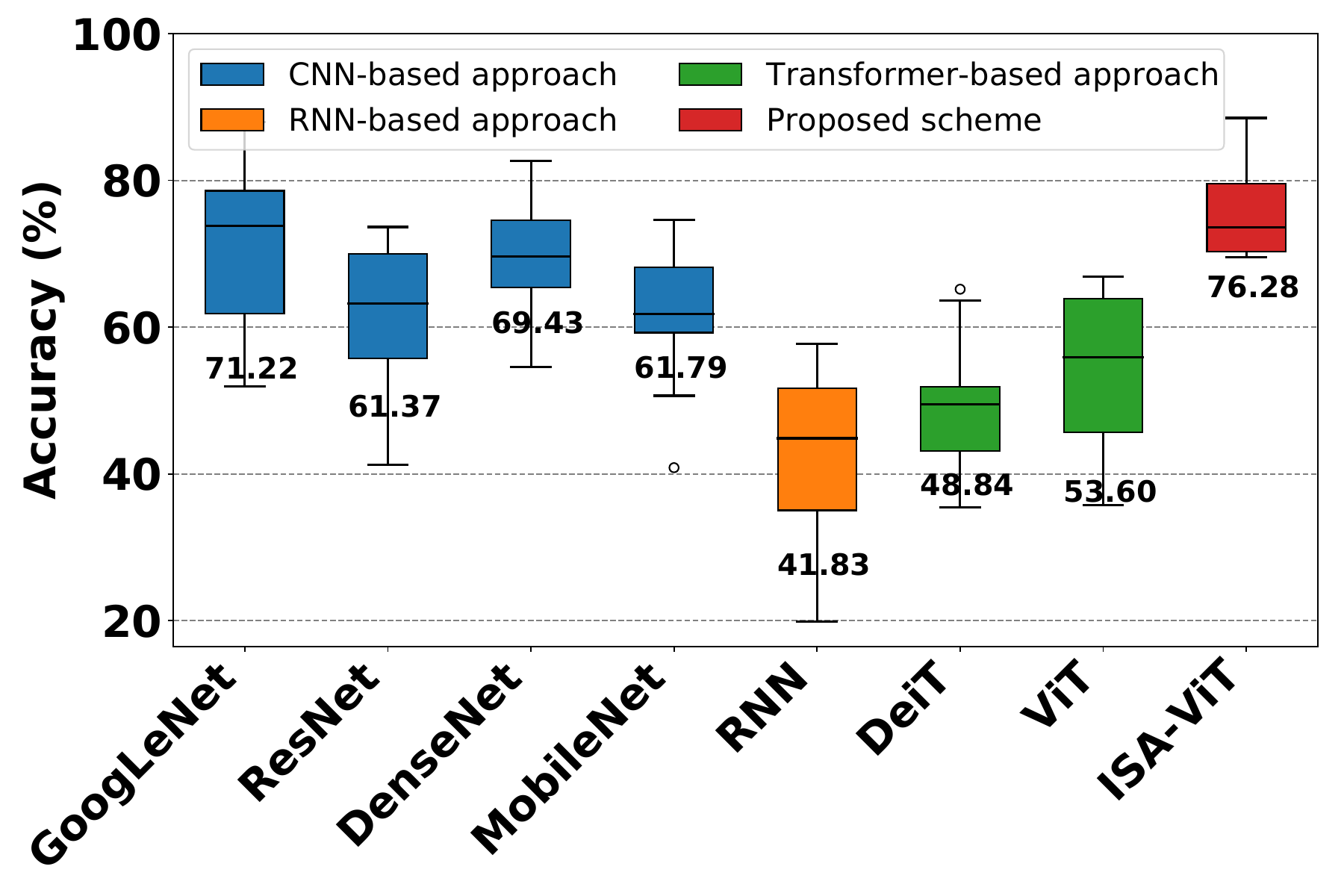}
    \label{fig:overallALERT}
    }
    \subfigure[Variation of precision and recall]{
    \includegraphics[width=0.45\textwidth, height = 4cm]{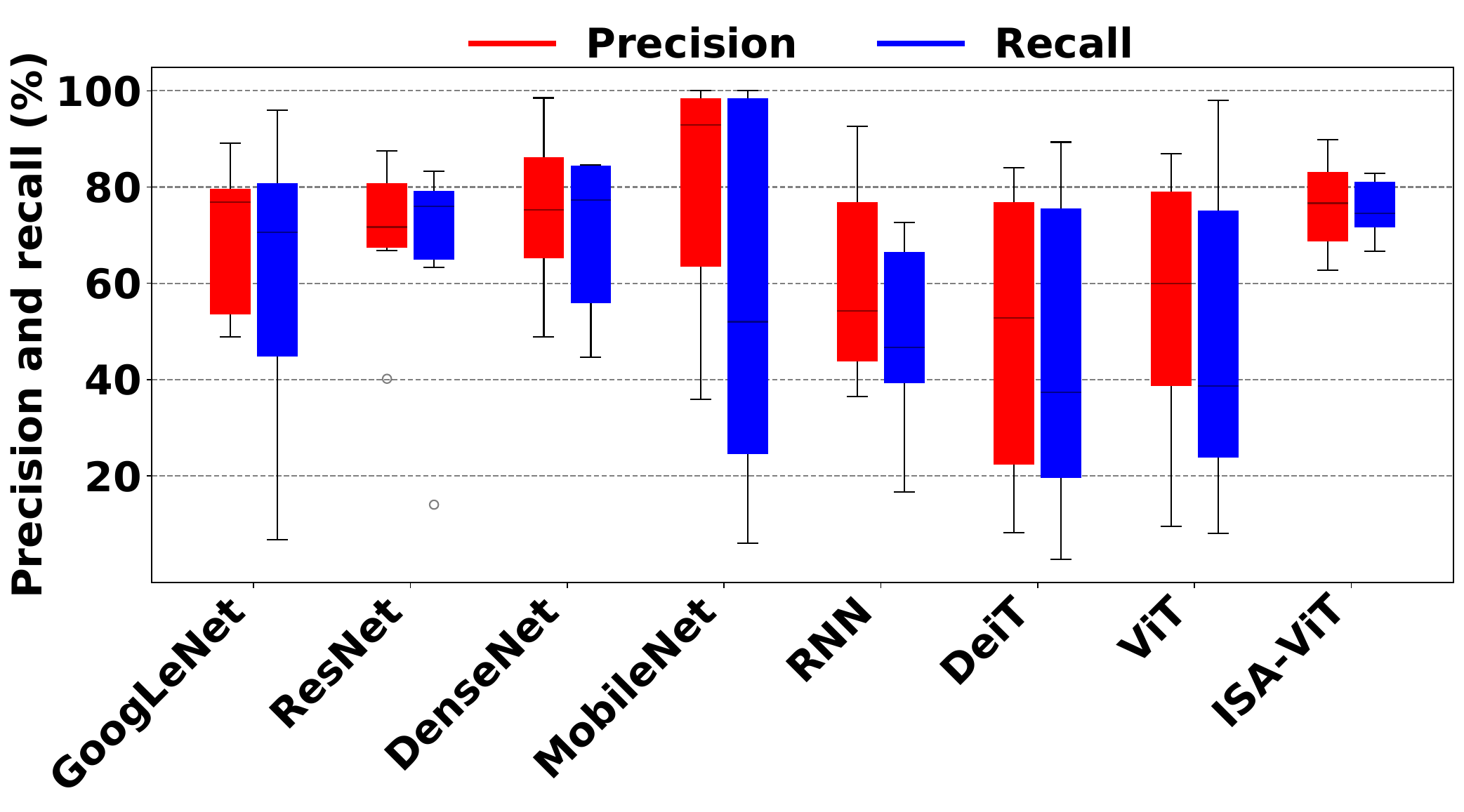}
    \label{fig:boxplot}
    }
    \caption{\rev{Benchmarking results for overall performance: accuracy, precision, and recall.}}
    \vspace{-10pt}
\label{fig:benchmarking}
\end{figure*}
\begin{table*}[t!]
\centering
\caption{\re{Tester-wise standard deviation ($\sigma$) of learning algorithms.}}
\label{tab:tester_std_fig9a_wide}
\setlength{\tabcolsep}{10pt}
\renewcommand{\arraystretch}{1.05}
\begin{tabular}{lcccccccc}
\hline
\textbf{Method} & GoogLeNet & ResNet & DenseNet & MobileNet & RNN & DeiT & ViT & ISA-ViT \\
\hline
$\sigma$ & 9.97 & 7.95 & 7.33 & 7.22 & 14.13 & 8.81 & 12.29 & 7.54 \\
\hline
\end{tabular}
\label{std}
\end{table*}

\begin{table*}[t!]
\caption{Benchmarking [precision / recall] of learning algorithms}
\vspace{-10pt}
\centering
\resizebox{\textwidth}{!}{ 
    \begin{tabular}{|c|c|c|c|c|c|c|c|c|}  
    \hline  
    \diagbox{Act.}{Alg.} & GoogLeNet & ResNet & DenseNet & MobileNet & RNN & DeiT & ViT & \mine\\ \hline
    \emph{Drive} & 89.07 / 70.66 &84.72 / 81.33 & 90.71 / 84.66 & 65.17 / 97.33 &68.62 / \underline{46.66} &	78.01 / 73.33 & 86.88 / \underline{35.33} & 68.51 / 79.76\\ \hline
    \emph{Relax} & 80.00 / \underline{34.66} & 76.68 / 83.33 & 72.35 / 59.33 & 61.98 / 100.0 &  84.72 / \underline{40.66} & 75.32 / 77.33 & 72.41 / 98.00 & 89.87 / 69.50 \\ \hline
    \emph{Nod} & 50.20 / 82.00 & 66.86 / 76.66 &  58.29 / 77.33 & 96.66 / \underline{38.66} &  \underline{45.41} / 72.66 & 52.83 / \underline{37.33} & 85.29 / \underline{38.66}  & 69.12 / 82.83\\ \hline
    \emph{Drink} & \underline{48.81} / 96.00 & 68.34 / 63.33 & 75.23 / 52.66 & 100.0 / \underline{6.00} &  92.59 / \underline{16.66} & 84.00 / \underline{28.00} & 60.00 / 54.00 & 76.64 / 82.00\\ \hline
    \emph{Phone} & 79.04 / 55.33& 71.69 / 76.00 & 81.29 / 84.00 & 92.85 / 52.00 &  \underline{42.46} / 71.33 & \underline{34.35} / 89.33 & \underline{41.26} / 96.00 & 81.50 / 74.17\\ \hline
    \emph{Smoke} & 76.92 / \underline{6.66} &  \underline{40.16} / 66.66 & \underline{48.84} / 84.66 & \underline{35.90} / 99.33 &  \underline{36.50} / 61.33 & \underline{10.62} / \underline{11.33} & \underline{9.45} / \underline{12.66} & 62.75 / 66.67\\ \hline
    \emph{Panel} & 57.21 / 79.33 & 87.5 / \underline{14.00} & 98.52 / \underline{44.66} & 100.0 / \underline{10.66} &  54.28 / 38.00 & \underline{8.16} / \underline{2.66} & \underline{36.36} / \underline{8.00} & 84.34 / 74.50\\ \hline
    \end{tabular}
}
\label{f1score}
\end{table*}

\subsection{Benchmarking Results for ALERT Dataset}

\subsubsection{\textbf{Benchmarking of learning algorithms}}
For rigorous evaluation, we conduct benchmark tests using an unseen dataset during the training phase. In these experiments, we fine-tune pre-trained models, leveraging the pre-trained models' great capabilities for feature extraction effectively to the UWB domain.
\re{This fine-tuning step is also essential to reduce the residual image-to-radar domain gap even after resolving the structural input-size and PEV mismatch.}

To identify a suitable model for UWB DAR, we first present the average accuracy of each algorithm. We select four unseen testers and perform leave-one-out cross-validation to calculate their average accuracy.
\re{In each fold, the held-out tester is used only for testing, while all remaining testers are used for training, so that no subject overlaps between training and testing.} 
\re{Fig.~\ref{fig:overallALERT} and Table.~\ref{std} show a box plot and tester-wise standard deviation of various learning algorithms over three repeated runs (N=3) with different random seeds.}

\re{CNN-based approaches achieve 61.37--71.22\% accuracy, outperforming both RNN- and transformer-based methods. The weaker performance of the RNN-based approach is mainly due to the absence of a pre-trained LSTM: although it uses pre-trained backbones for snippet-level feature extraction, the LSTM itself is trained from scratch, limiting overall effectiveness.}

\re{Transformer-based approaches also underperform CNNs in this UWB setting. While transformers are strong for image tasks, naïvely applying them to UWB requires resizing inputs and aligning pre-trained PEVs; without properly handling these issues (Sec.~\ref{sec:ISA-ViT}), they cannot fully realize their potential.}

\re{In contrast, \mine addresses this mismatch by resizing UWB data without information loss and adapting pre-trained PEVs accordingly, achieving the best accuracy of 76.28\%.}

\re{Beyond overall accuracy, we also report per-activity precision and recall, which provide complementary insight into classification performance. Table~\ref{f1score} summarizes the precision and recall for each activity across the evaluated algorithms; the underlined entries highlight activities with notable fluctuations. Such variance is important because it indicates how consistently a model performs across different activities. Fig.~\ref{fig:boxplot} visualizes the mean and deviation of these metrics for each algorithm. As shown, several baselines exhibit large deviations, whereas \mine maintains relatively stable variance, resulting in more consistent per-activity performance.}

\begin{figure*}[t]
\centering
\subfigure[Accuracy according to observation window size]{
    \includegraphics[width=7cm, height=4cm]{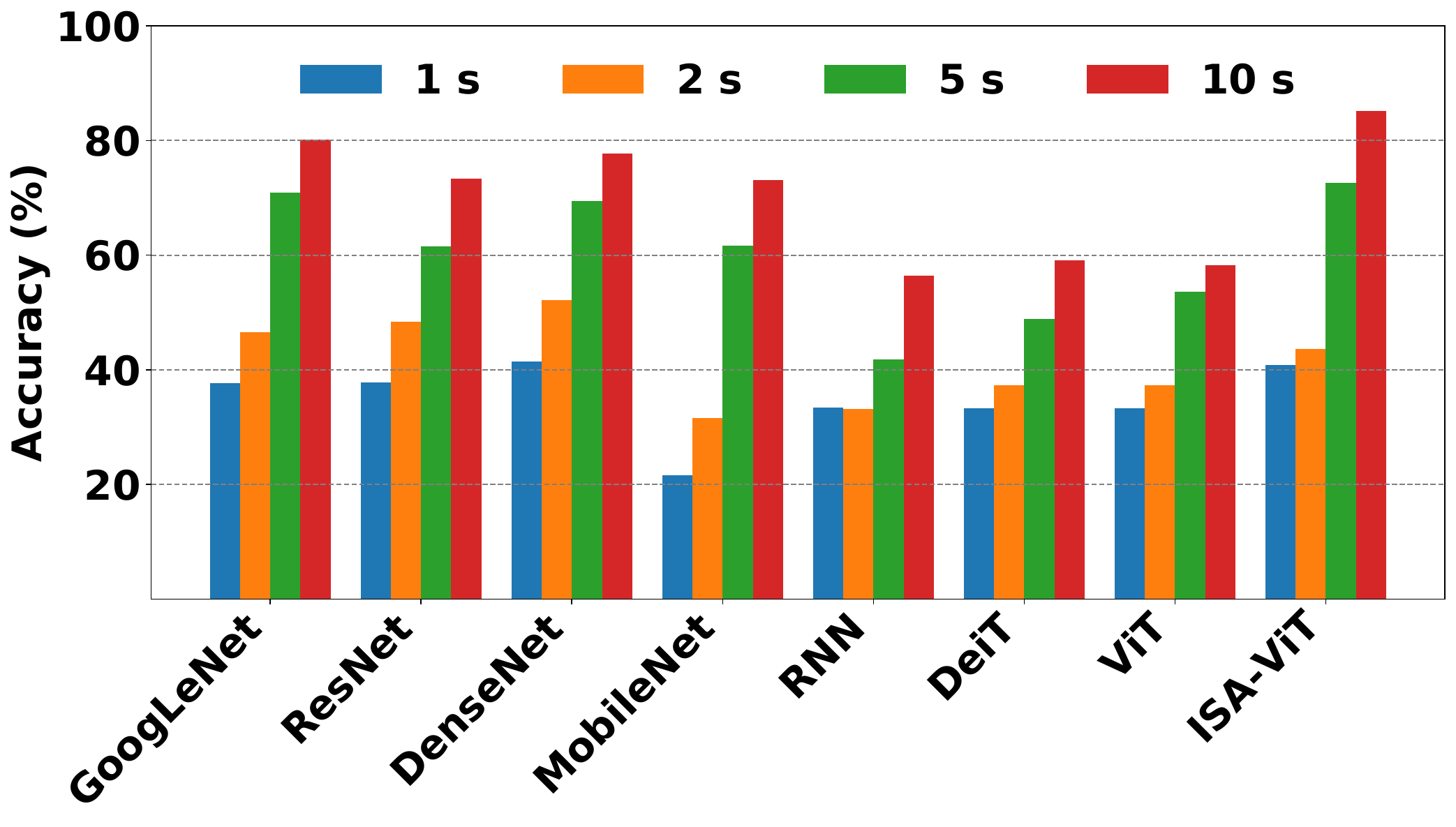}
    \label{fig:obserwindow}
}
\hspace{-5pt}
\subfigure[Accuracy according to information cropping]{
    \includegraphics[width=6.8cm, height=4cm]{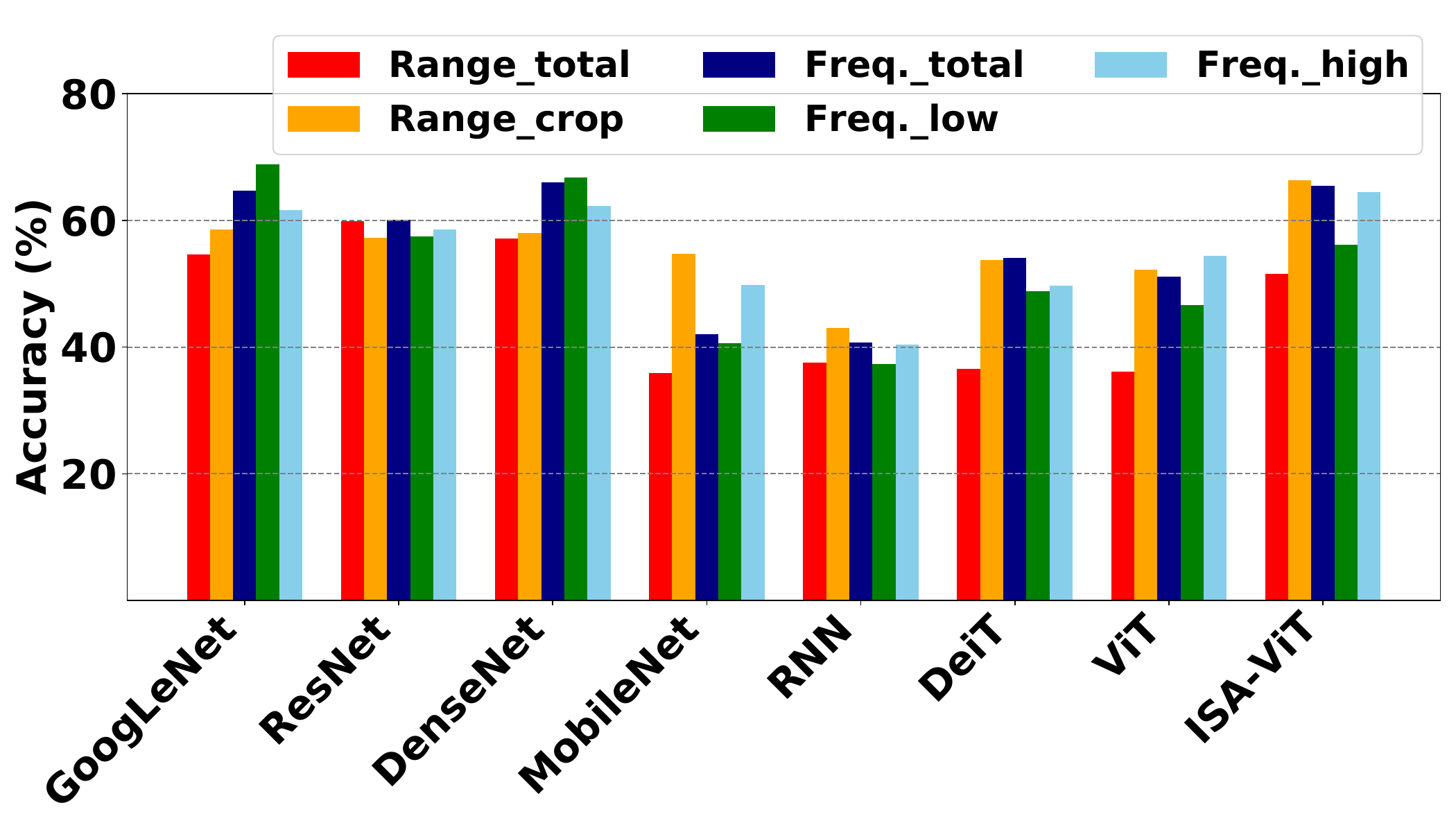}
    \label{fig:crop}
}
\subfigure[Accuracy according to few shot adaptation]{
    \includegraphics[width=7cm, height=4cm]{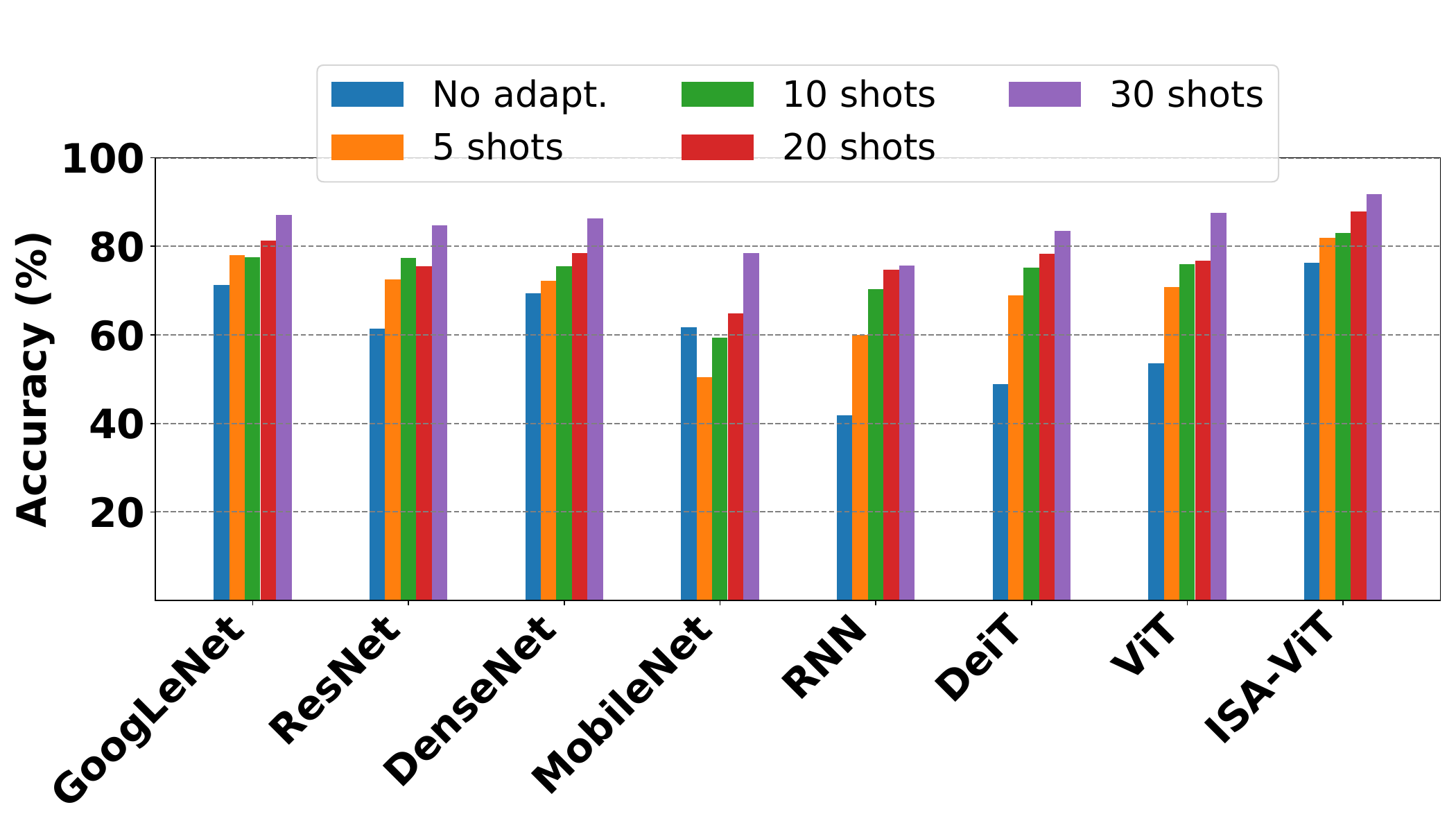}
    \label{fig:fewshot}
}
\hspace{-5pt}
\subfigure[Accuracy according to domain usage]{
    \includegraphics[width=7cm, height=4cm]{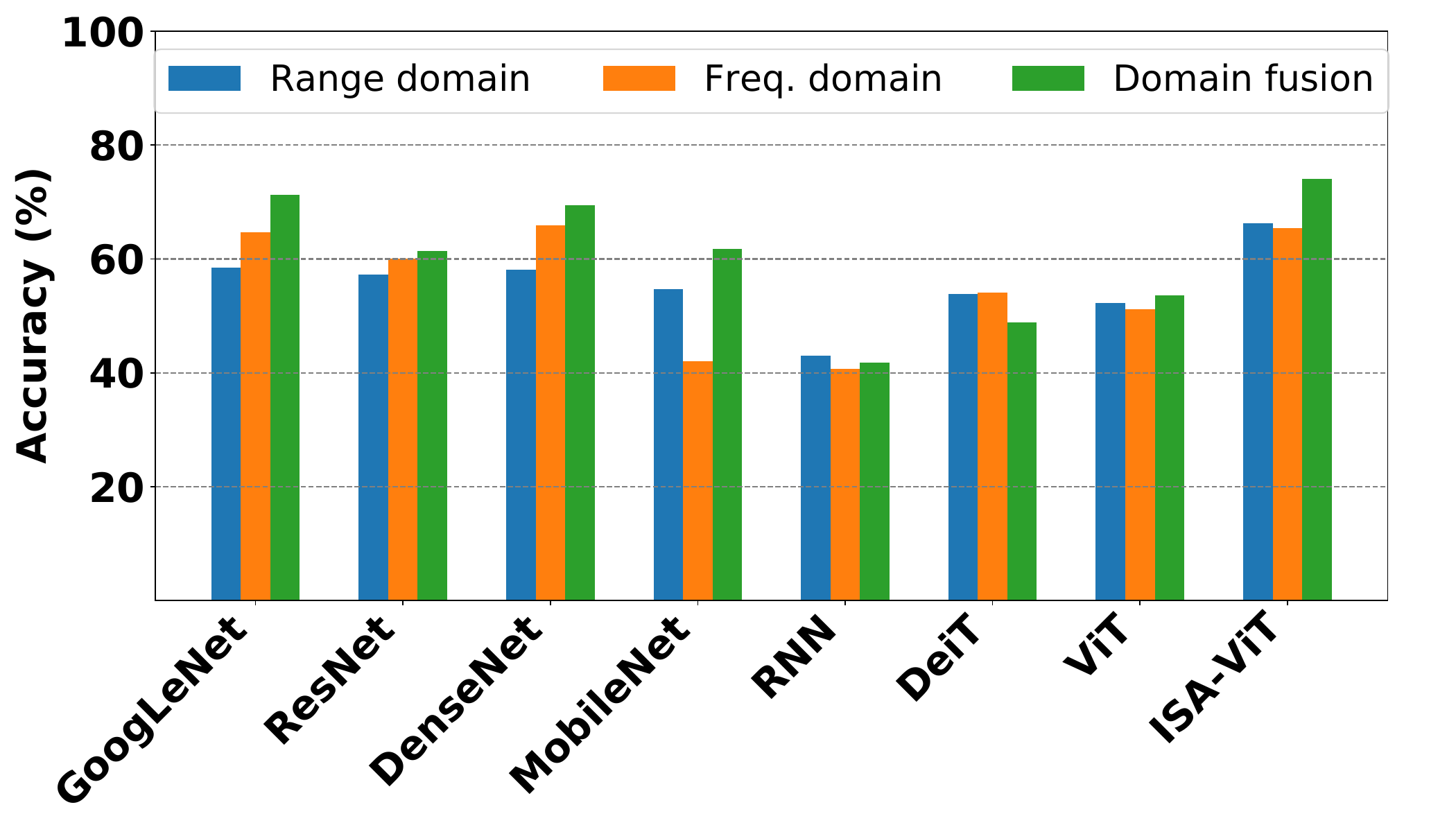}
    \label{fig:mvl}
}
\caption{Ablatioin studies for \ALERT dataset.}
\label{fig:able}
\vspace{-10pt}
\end{figure*}

\subsection{Ablation Studies for ALERT dataset}
We conduct various experiments to identify the optimal parameters, trends, and trade-offs for each algorithm when utilizing the \ALERT dataset.
\re{We provide granular ablations beyond resizing/patch selection, including data-size manipulation, few-shot shots, domain usage (range/frequency/fusion), range-Doppler conversion, and fusion strategy (early vs late), as summarized in Figs.~\ref{fig:able}--\ref{fig:early_fusion}.}

\subsubsection{\textbf{Analysis for manipulation of data size}}
We conduct experiments with varying observation window sizes, range bins, and frequency bins of \ALERT dataset to determine the optimal data size for accurately capturing activities.

\re{As shown in Fig.~\ref{fig:obserwindow}, accuracy increases with a longer observation window. Short windows (1--2~s) provide limited temporal context and thus yield lower performance, whereas 5--10~s windows offer sufficient duration to better capture activity patterns. However, larger windows also increase inference latency and computational cost. Therefore, the observation window should be selected by balancing accuracy, system responsiveness, and practical efficiency.}

\re{In Fig.~\ref{fig:crop}, we compare using all range bins (about 9~m) versus a cropped driver-centric ROI ($<$2.7~m) to assess the effect of long-delayed multipath components. While the benefit depends on the model structure, cropping improves performance for all algorithms except ResNet. \rev{Although long-delayed multipath signals may occasionally include contextual cues (e.g., passengers or vehicle type), they typically behave as noise for most algorithms; therefore, cropping is generally beneficial.}}

\re{We also evaluate lower ($<$375~MHz), higher (375--750~MHz), and full frequency bands to identify which region best captures motion characteristics for each activity. The optimal choice varies across algorithms, reflecting differences in feature-extraction capability; thus, these results can guide selecting an appropriate model and frequency band.}

\subsubsection{\textbf{Analysis for the number of shots}}

We present the impact of varying the number of shots on a few-shot adaptation. 
\re{We adopt few-shot adaptation to reduce the domain gap when transferring the PEV capability learned from large-scale image pre-training to the UWB domain.}
To obtain a few shot samples, the driver has to capture the activities, which can be burdensome; however, this occurs only once when the system is initially operated.
For the few-shot adaptation phase, we utilize various number of shot samples and 10 epochs adaptation, which are enough to fit the model to a driver.
\re{During adaptation, we fine-tune the entire model (pre-trained PEVs, backbone, and classifier) for 10 epochs using Adam with the same learning rate as training (model-dependent, 0.00001–0.0001) and an adaptation batch size of 5.}
As shown in Fig.~\ref{fig:fewshot}, the accuracy increases when all models' number of shots increases. 
\re{In our experiments, the few-shot samples are taken only from the Relax behavior, and they are strictly excluded from the testing set to prevent any data leakage.}
We observe that even with the addition of only about 5 shots, the accuracy gain is significant. In particular, when applying 30 shots to ISA-ViT, it exhibits a maximum performance of 91.75\%.

However, we spend 50, 100, and 150 seconds per activity to obtain 10, 20, and 30 shots, respectively. It can be a burden to the drivers although we collect only once at the first time. We should increase the number of shots in perspective of accuracy but decrease the number of shots to reduce the driver's burden. Thus, we should select a moderate number of shots, considering the balance between accuracy and burdens.

\subsubsection{\textbf{Analysis for the input data domain}}
The \ALERT dataset provides both range and frequency domain data, enabling users to select the data domain for their specific needs. To evaluate the impact of each domain on performance, we assessed the use of range only, frequency only, and a combination of both domains, with the results illustrated in Fig.~\ref{fig:mvl}.

In CNN-based approaches, except for MobileNet, algorithms perform better when using the frequency domain compared to the range domain. Conversely, in RNN-based, transformer-based, and ISA-ViT models, using the range domain yield better results than using frequency domain.

Interestingly, using both domains does not consistently improve performance.
Except for RNN and DeiT, all algorithms demonstrate improved performance when using domain fusion compared to using a single domain. 
While combining two domains generally enhances performance through complementary synergy, this is not always true for all algorithms. These results indicate that the optimal domain should be chosen based on the characteristics of the specific algorithm.

\re{In particular, CNN-based models tend to benefit from frequency-domain inputs that provide richer local patterns, whereas transformer-based models (including \mine) more effectively exploit range-domain spatial cues; domain fusion is most beneficial when the model can balance complementary information from both domains.}

\subsubsection{\rev{Analysis for the range-Doppler data domain}}
\begin{figure}[t!]
  \centering
  \begin{minipage}[b]{0.45\textwidth}
    \centering
    \includegraphics[width=\textwidth, height=4cm]{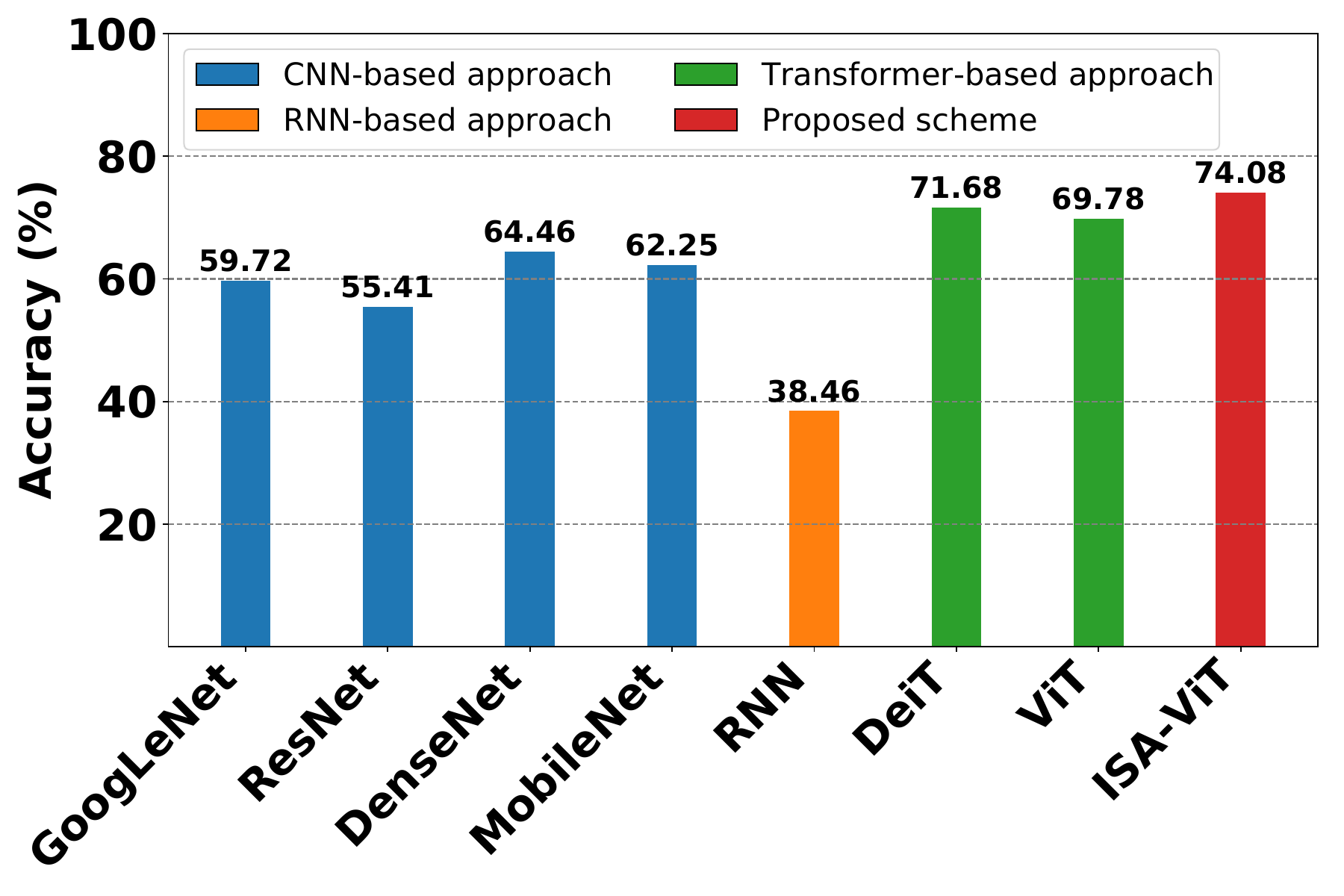}
    \caption{\rev{Average accuracy using range-Doppler domain of \ALERT.}}
    \label{fig:RD}
  \end{minipage}
  \hfill
  \begin{minipage}[b]{0.45\textwidth}
    \centering
    \includegraphics[width=7cm, height=4cm]{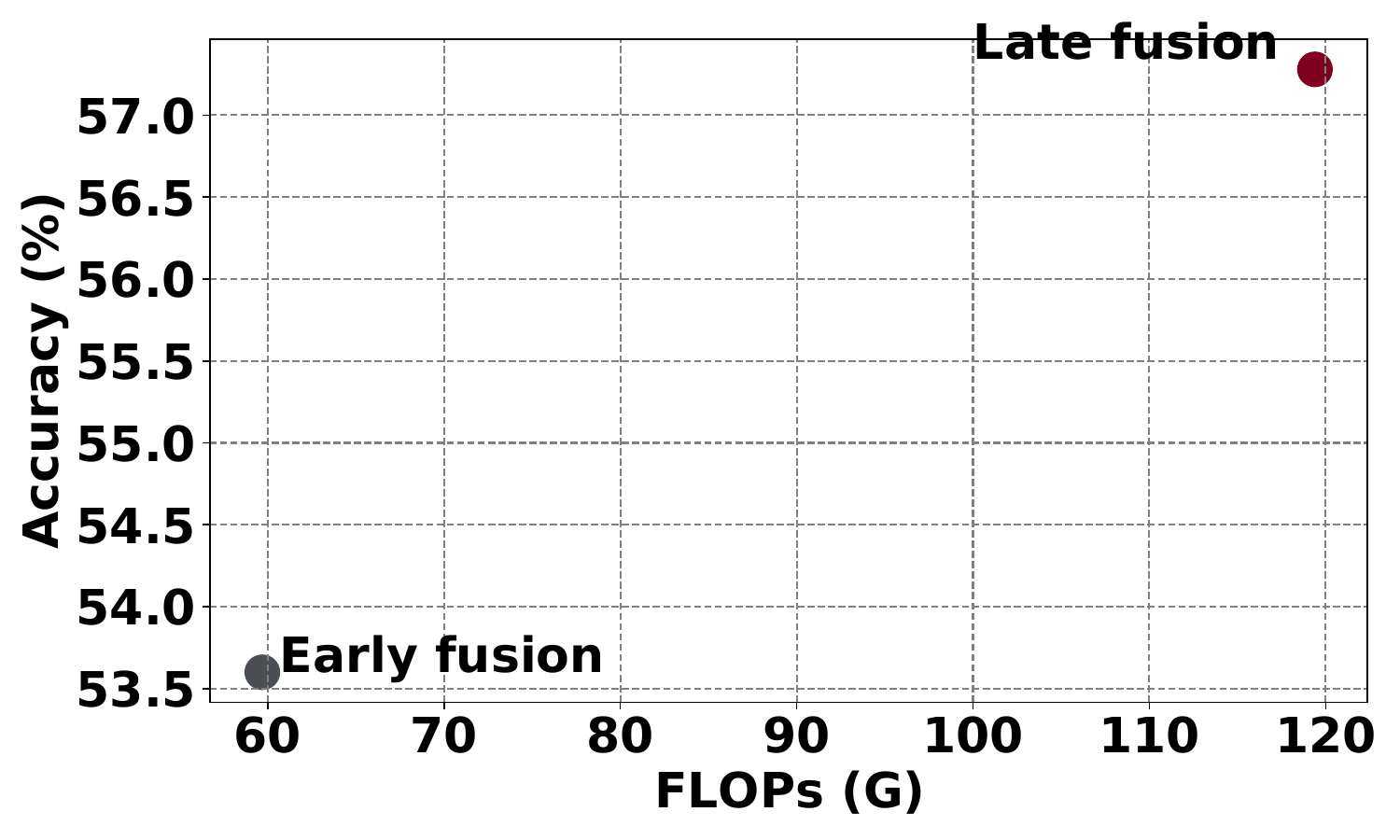}
    \caption{\rev{Accuracy and computational cost comparison of early fusion with late fusion in transformer-based approach.}}
    \label{fig:early_fusion}
  \end{minipage}
  \vspace{-10pt}
\end{figure}
\re{As shown in Fig.~\ref{fig:signal manipulation}, \ALERT provides UWB representations in the range--time and frequency--time domains and allows users to flexibly reformat the data for different needs. To demonstrate this flexibility, we additionally evaluate the range--Doppler representation, which is used in datasets such as RaDA and is also supported by \ALERT. This conversion can be performed using the provided script (ALERT\_makeDataset.py) in our public repository (\url{https://github.com/ALERTdataset/ALERT.git}).}

\re{As shown in Fig.~\ref{fig:RD}, CNN- and RNN-based methods with range--Doppler inputs do not outperform their domain-fusion counterparts in Fig.~\ref{fig:overallALERT}. In contrast, ViT-based models (DeiT and ViT) achieve higher performance with range--Doppler inputs than with domain fusion.}

\re{This trend is mainly related to input resizing and resolution loss. CNNs and RNNs can process UWB inputs directly without resizing, whereas DeiT and ViT require resizing to $224\times224$, which reduces resolution. In domain fusion, the time resolution is reduced by more than half for both range--time and frequency--time inputs; converting to range--Doppler also reduces the Doppler resolution by about half. Our results suggest that, for these ViT baselines, losing time resolution is more detrimental than losing Doppler resolution, making range--Doppler relatively preferable to domain fusion.}

\re{The proposed \mine builds on a ViT backbone but introduces a resizing strategy that preserves the original information, enabling improved performance even with range--Doppler inputs. Moreover, \mine employs complementary fusion of range and frequency features; as shown in Fig.~\ref{fig:overallALERT}, this yields higher accuracy than using range--Doppler alone.}

\subsubsection{\rev{Comparison of early fusion and late fusion}}

\rev{Unlike CNN-based and RNN-based approaches, transformer-based approaches have possible design choices for domain fusion, early fusion and late fusion. However, we applied the early fusion method in Section~\ref{sec:benchmark}. To verify the effectiveness of early fusion, we compare the accuracy and computational cost of early fusion and late fusion as shown in Fig.~\ref{fig:early_fusion}.}

\rev{Fig.~\ref{fig:early_fusion} shows the performance of ViT. While late fusion, which extracts features using separate ViTs, achieves approximately 3\% higher accuracy than early fusion, its computational cost is nearly double. Although early fusion may yield slightly lower accuracy, it provides significant benefits in terms of computational efficiency. In late fusion, each input is processed by a separate ViT, and the resulting features are combined and passed to a classifier. Since ViTs are computationally expensive, using two parallel ViTs imposes a heavy system burden. Therefore, we adopt early fusion for the transformer-based approach.}

\subsection{Data Verification}
\begin{figure}[t]
\centering

\subfigure[GoogLeNet]{%
  \includegraphics[width=0.12\textwidth]{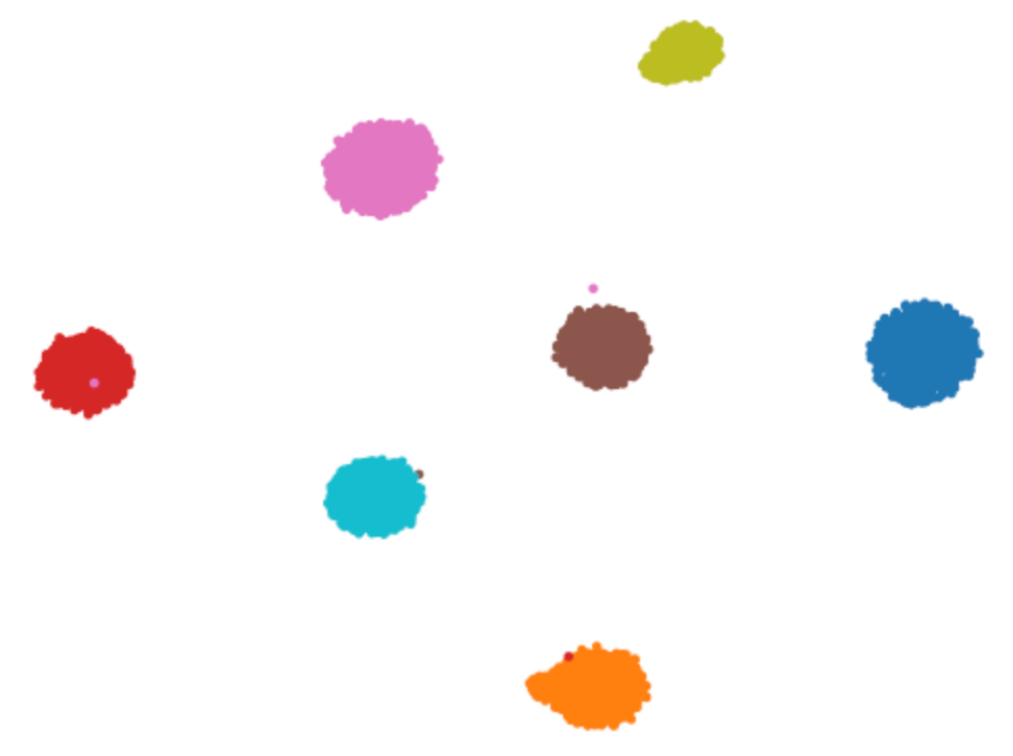}%
  \label{fig:sub1}%
}\hfill
\subfigure[ResNet]{%
  \includegraphics[width=0.12\textwidth]{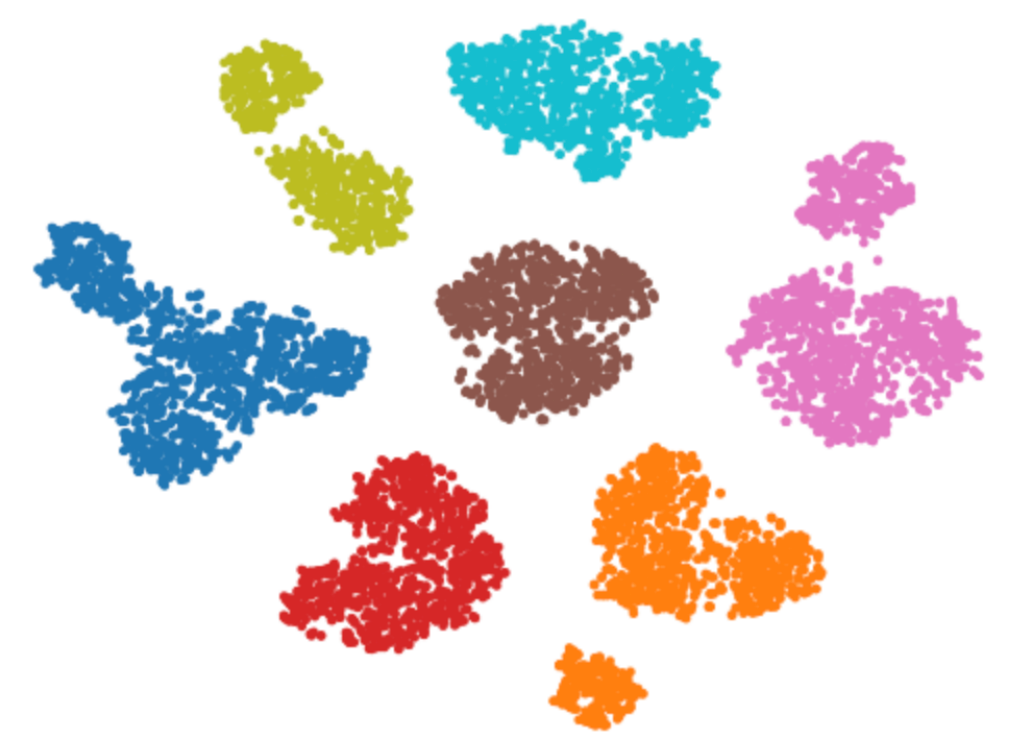}%
  \label{fig:sub2}%
}\hfill
\subfigure[DenseNet]{%
  \includegraphics[width=0.12\textwidth]{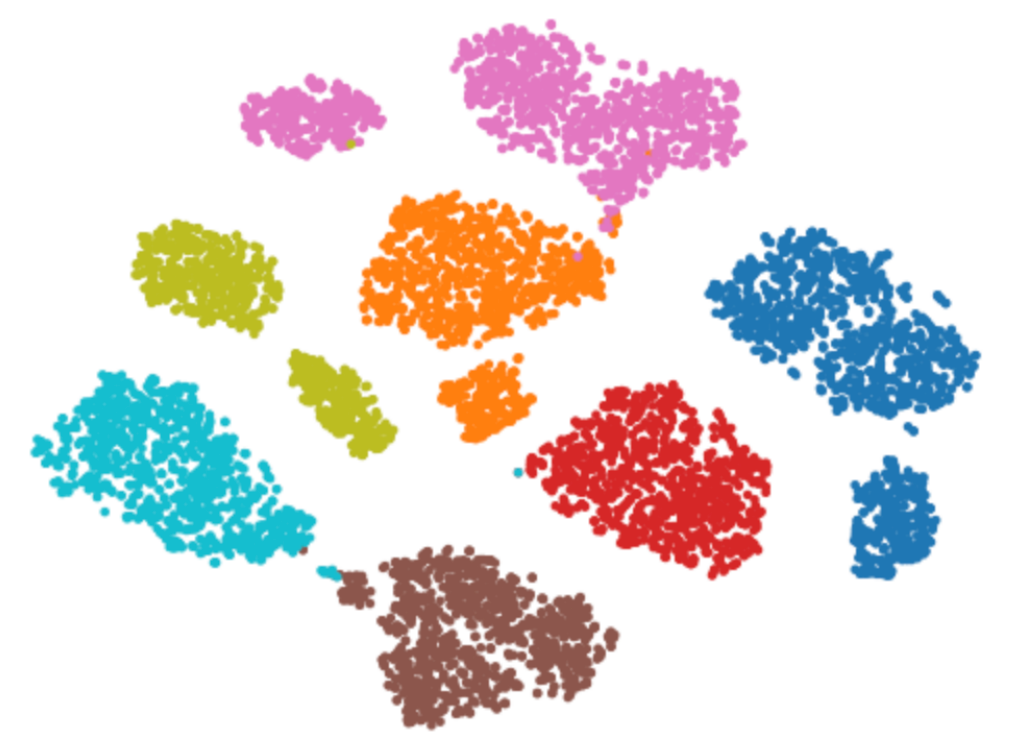}%
  \label{fig:sub3}%
}\hfill
\subfigure[MobileNet]{%
  \includegraphics[width=0.12\textwidth]{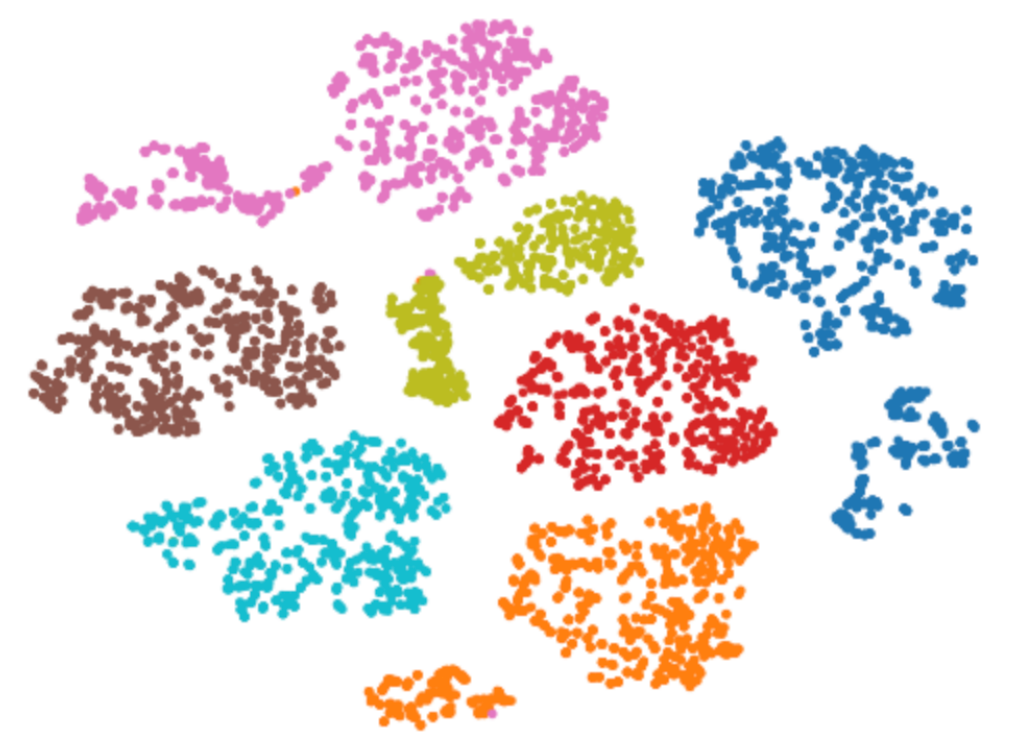}%
  \label{fig:sub4}%
}

\vspace{0.5em}

\subfigure[RNN]{%
  \includegraphics[width=0.12\textwidth]{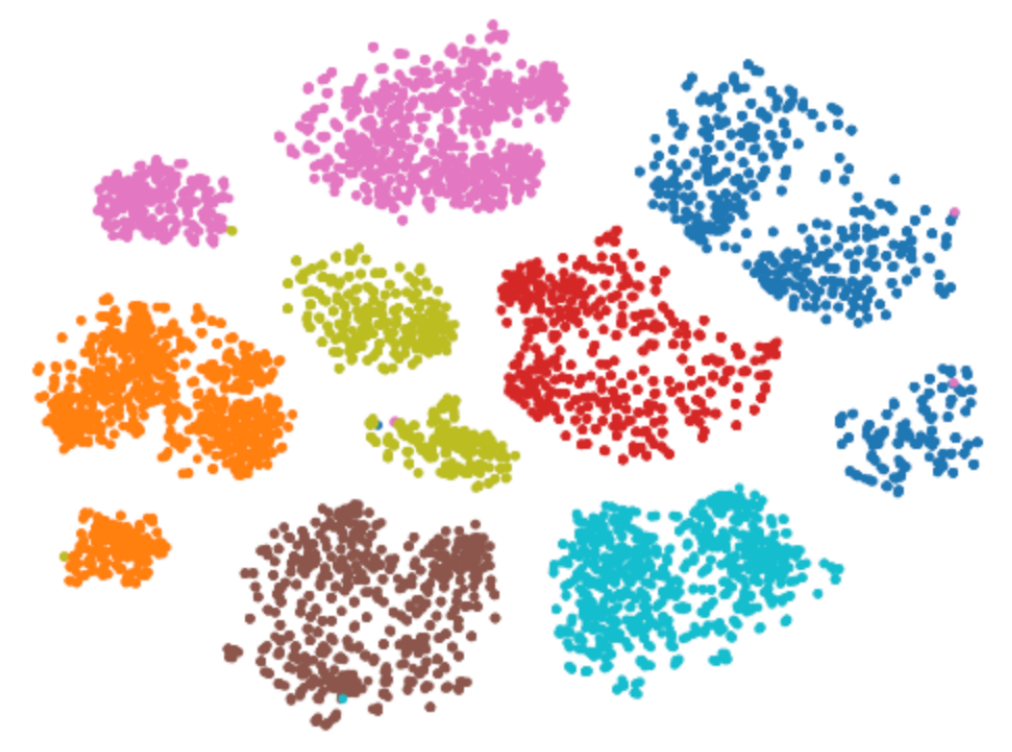}%
  \label{fig:sub5}%
}\hfill
\subfigure[DeiT]{%
  \includegraphics[width=0.12\textwidth]{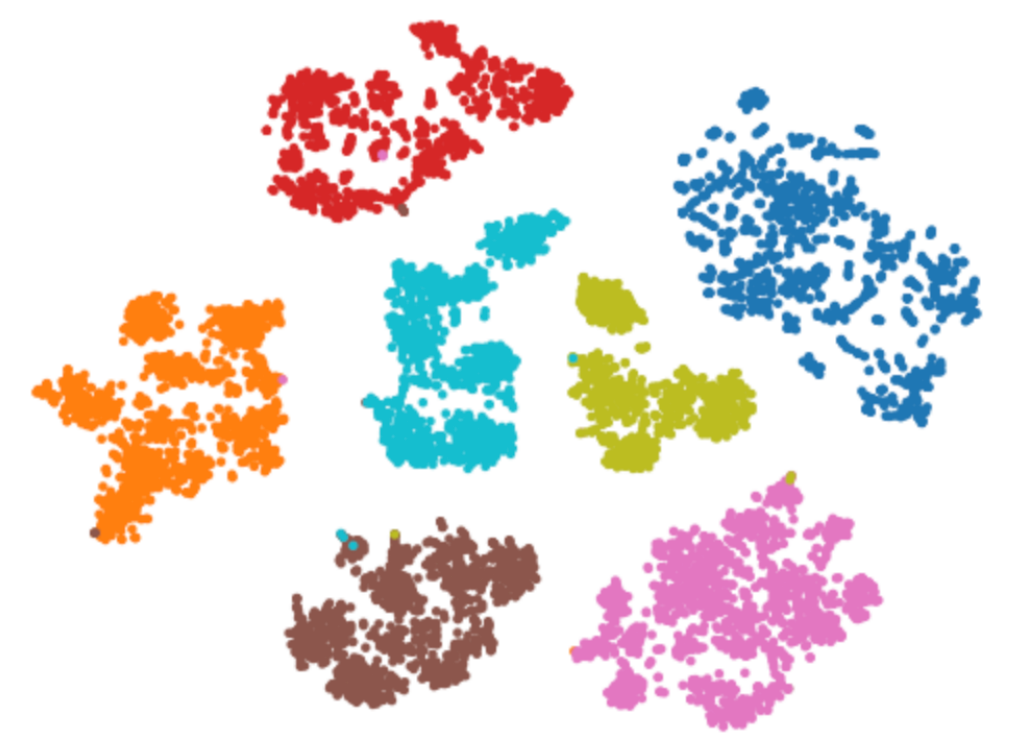}%
  \label{fig:sub6}%
}\hfill
\subfigure[ViT]{%
  \includegraphics[width=0.12\textwidth]{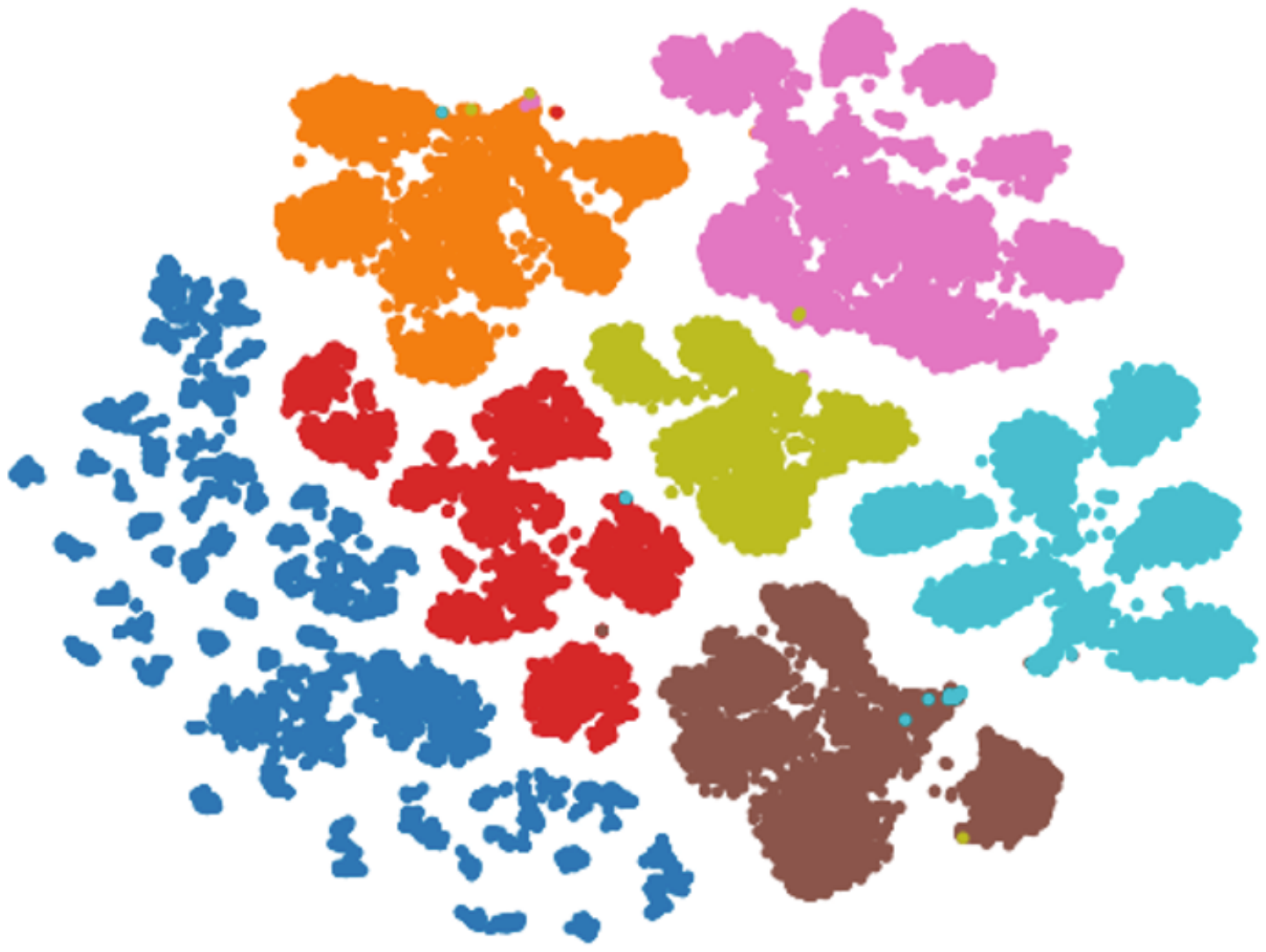}%
  \label{fig:sub7}%
}\hfill
\subfigure[\mine]{%
  \includegraphics[width=0.12\textwidth]{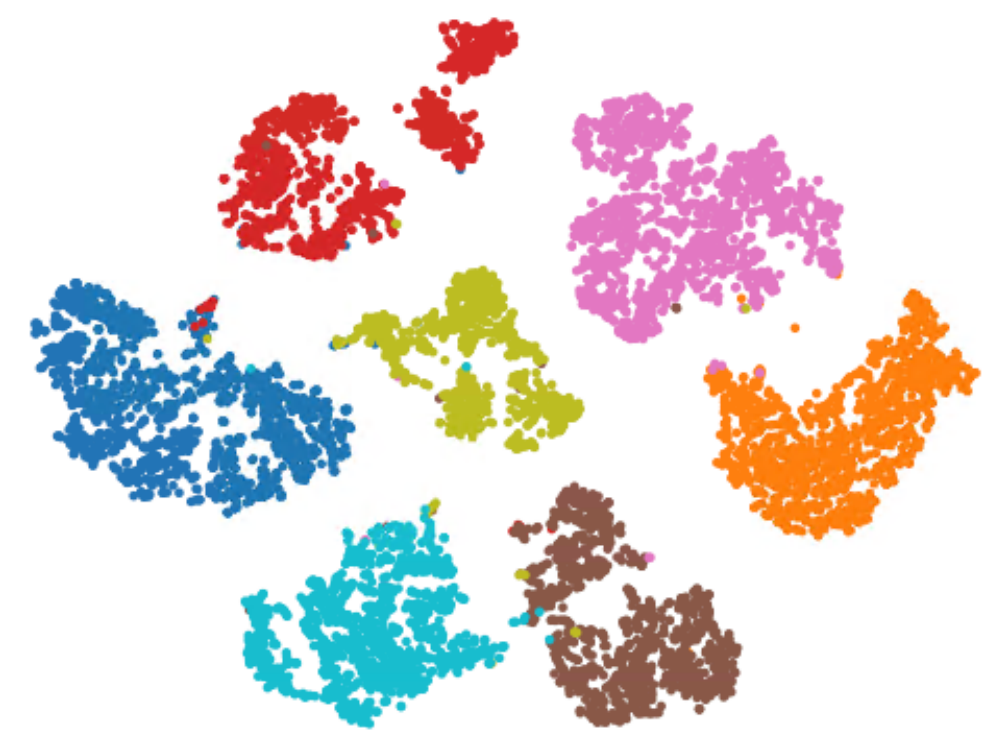}%
  \label{fig:sub8}%
}

\vspace{0.5em}

\begin{minipage}{0.45\textwidth}
\centering
\includegraphics[width=\textwidth, height=0.4cm]{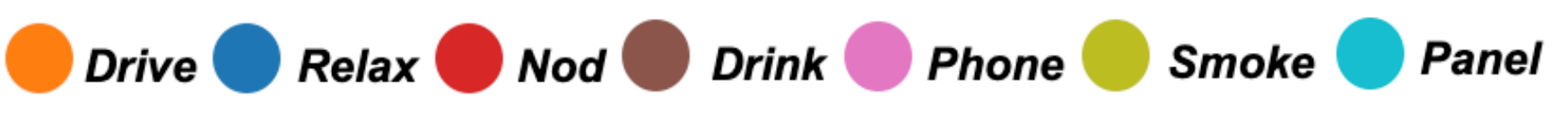}
\end{minipage}

\caption{\rev{T-SNE based feature visualization of learning algorithms.}}
\label{fig:tsne}
\vspace{-10pt}
\end{figure}

To confirm that the \ALERT dataset is suitable for training models, we indirectly verify it by analyzing its feature distribution. Typically, we use visualization of feature distributions to verify models; but, if we can confirm well-distributed features when we pass through the \ALERT dataset to the verified model, we can verify our dataset. Fig.~\ref{fig:tsne} illustrates the feature distribution of each algorithm using t-distributed stochastic neighbor embedding (T-SNE). We reduce the dimension of features from the penultimate layer of each model. 

All models exhibit differentiation of features. Overall, this analysis confirms that the \ALERT dataset is meaningfully assembled. Note that the results only verify whether our data can be used well enough for the models, so there is no relationship between overall accuracy. Accuracy may vary depending on the distribution of testers. Thus, the results guarantee training accuracy, not testing accuracy.

\begin{figure*}[t]
\centering

\subfigure[Drive]{%
  \includegraphics[width=0.14\textwidth]{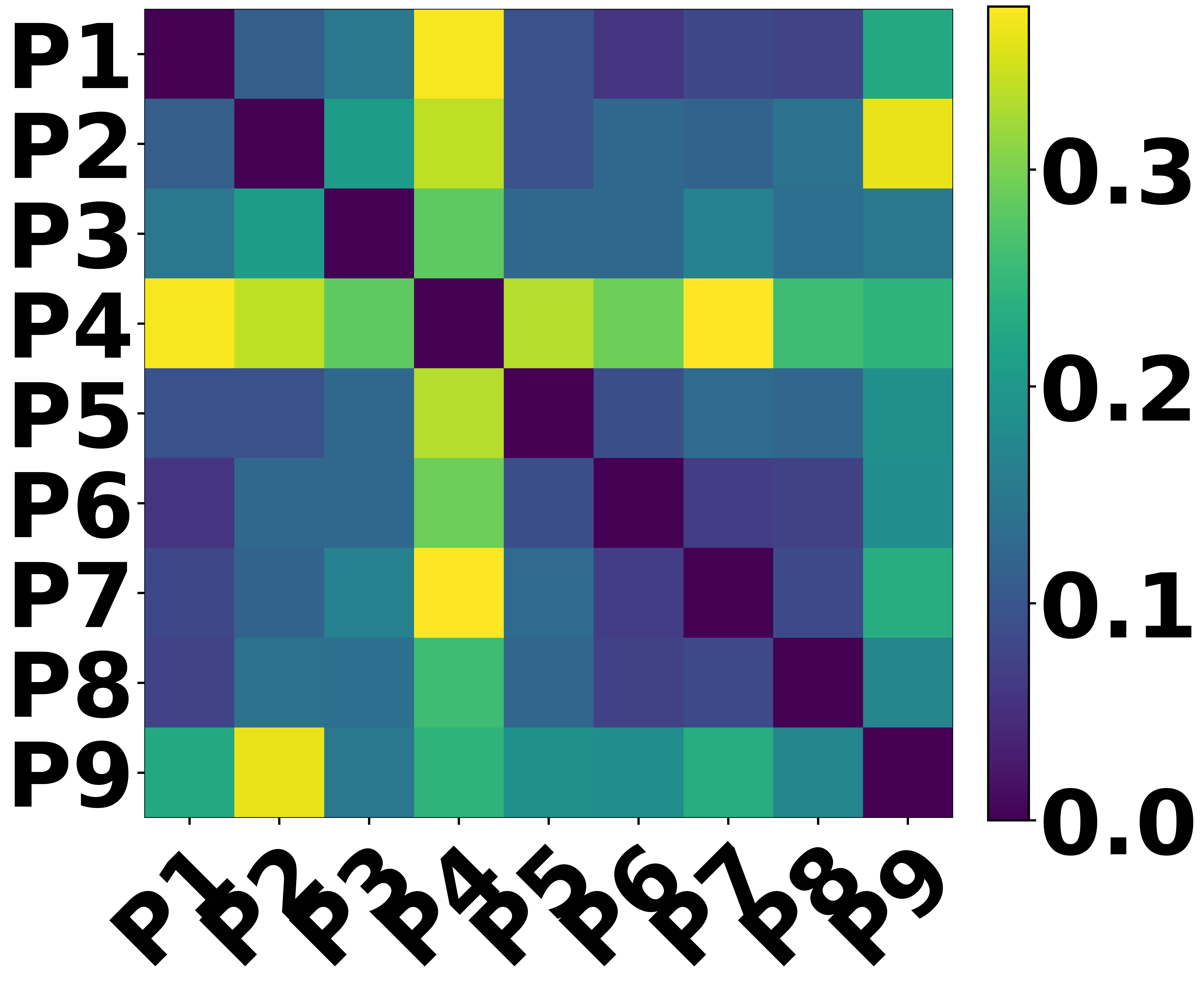}%
  \label{fig:hm_tsw}%
}\hfill
\subfigure[Relax]{%
  \includegraphics[width=0.14\textwidth]{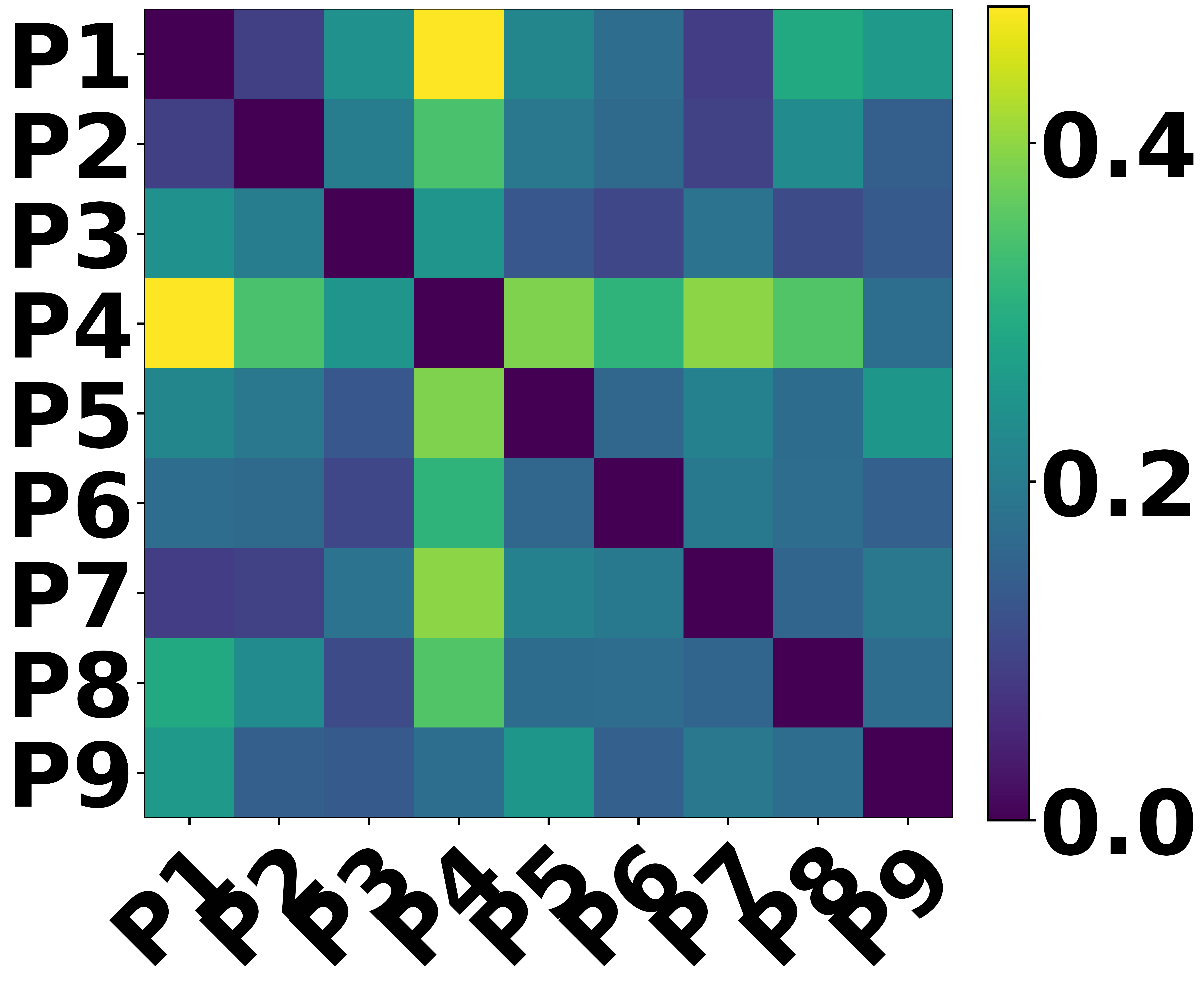}%
  \label{fig:hm_relax}%
}\hfill
\subfigure[Nod]{%
  \includegraphics[width=0.14\textwidth]{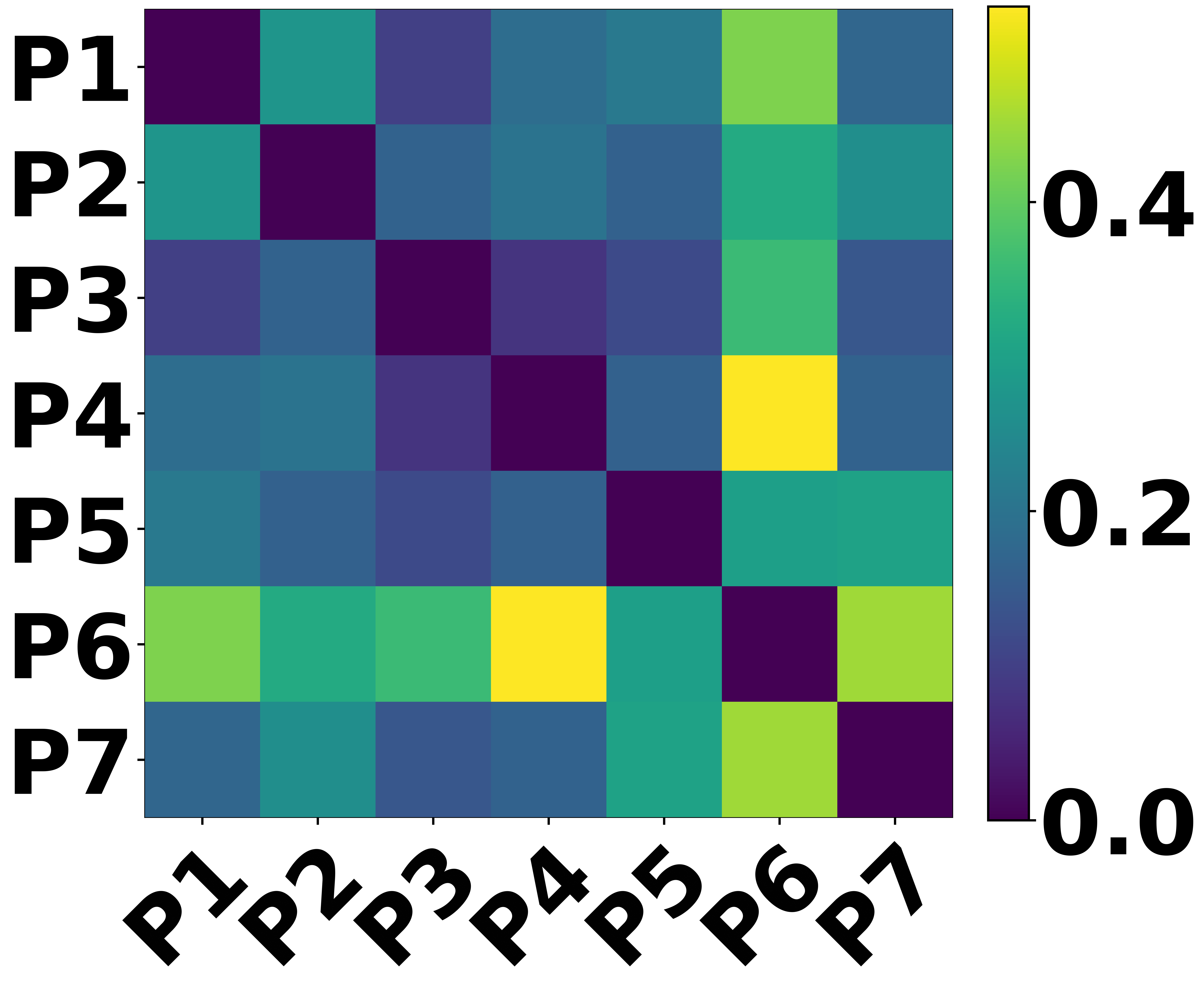}%
  \label{fig:hm_nod}%
}\hfill
\subfigure[Drink]{%
  \includegraphics[width=0.14\textwidth]{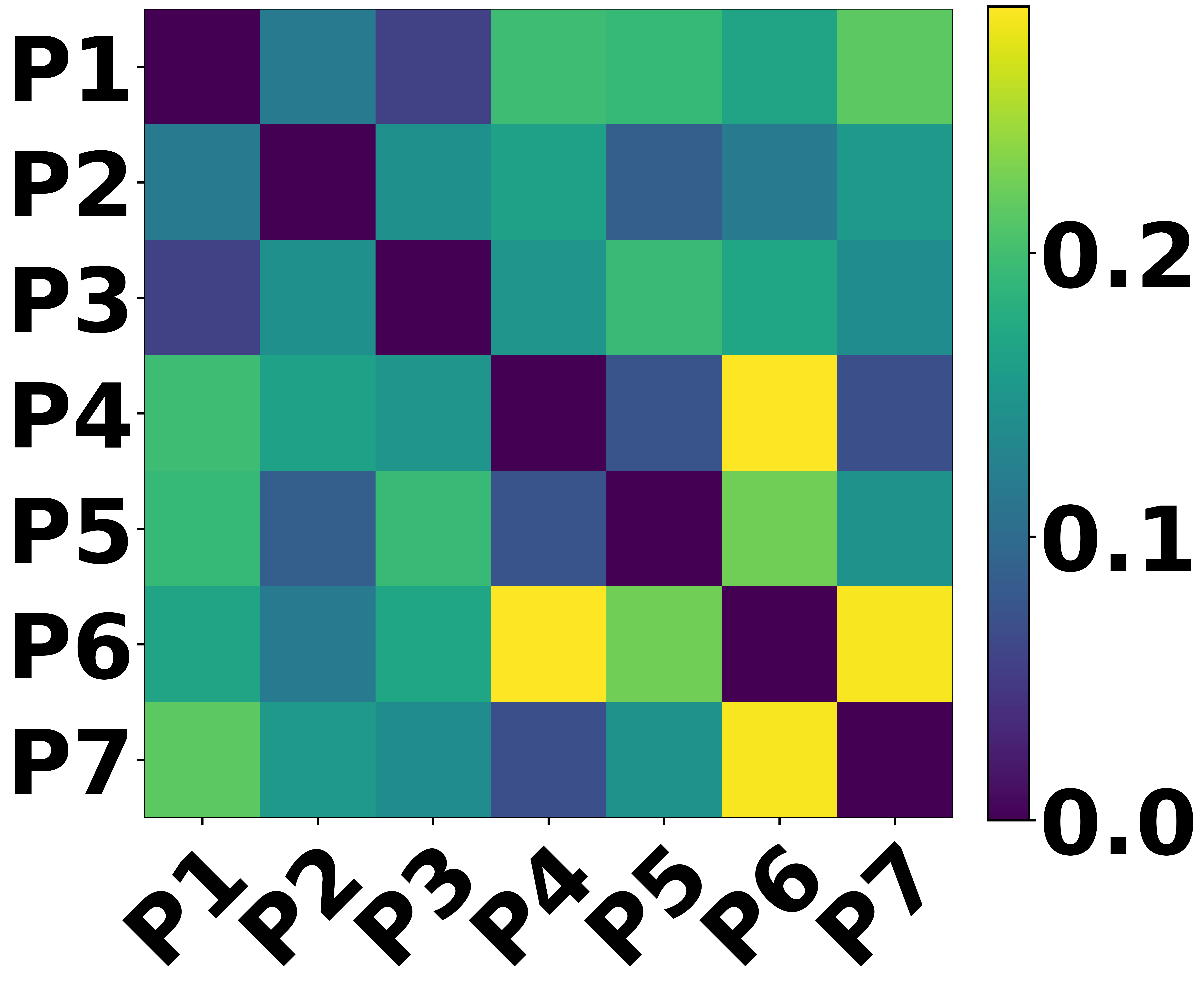}%
  \label{fig:hm_drink}%
}\hfill
\subfigure[Phone]{%
  \includegraphics[width=0.14\textwidth]{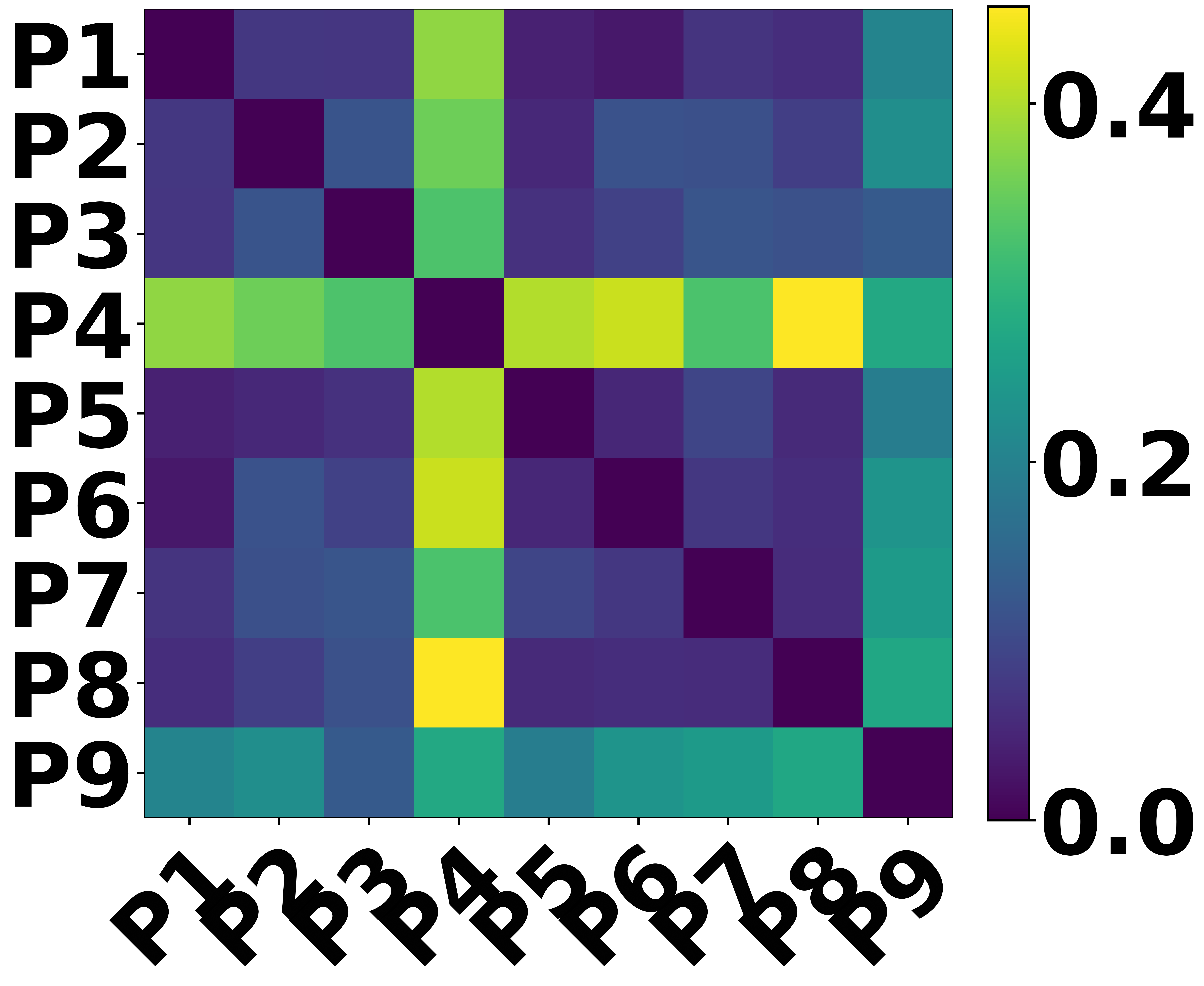}%
  \label{fig:hm_phone}%
}\hfill
\subfigure[Smoke]{%
  \includegraphics[width=0.14\textwidth]{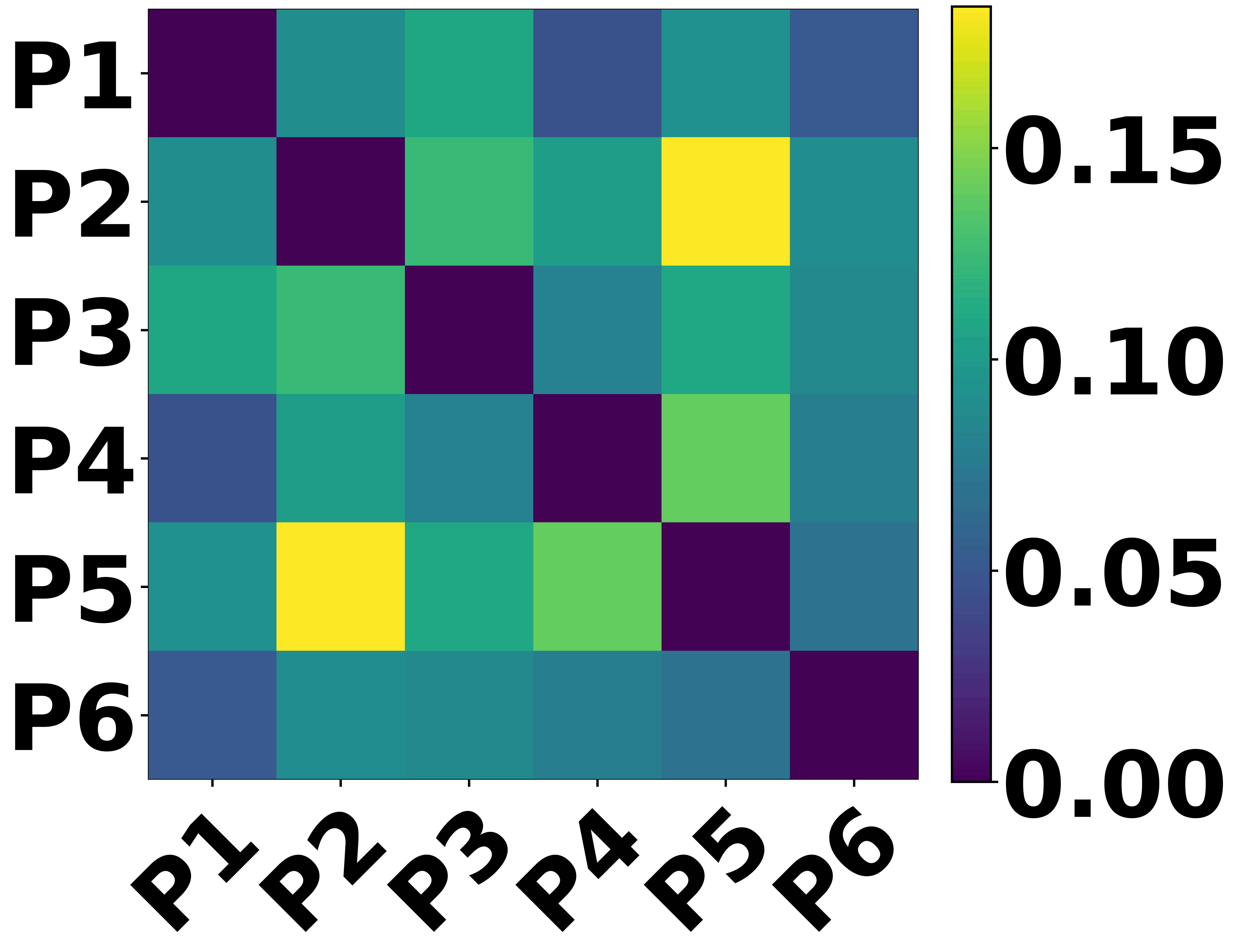}%
  \label{fig:hm_smoke}%
}\hfill
\subfigure[Panel]{%
  \includegraphics[width=0.14\textwidth]{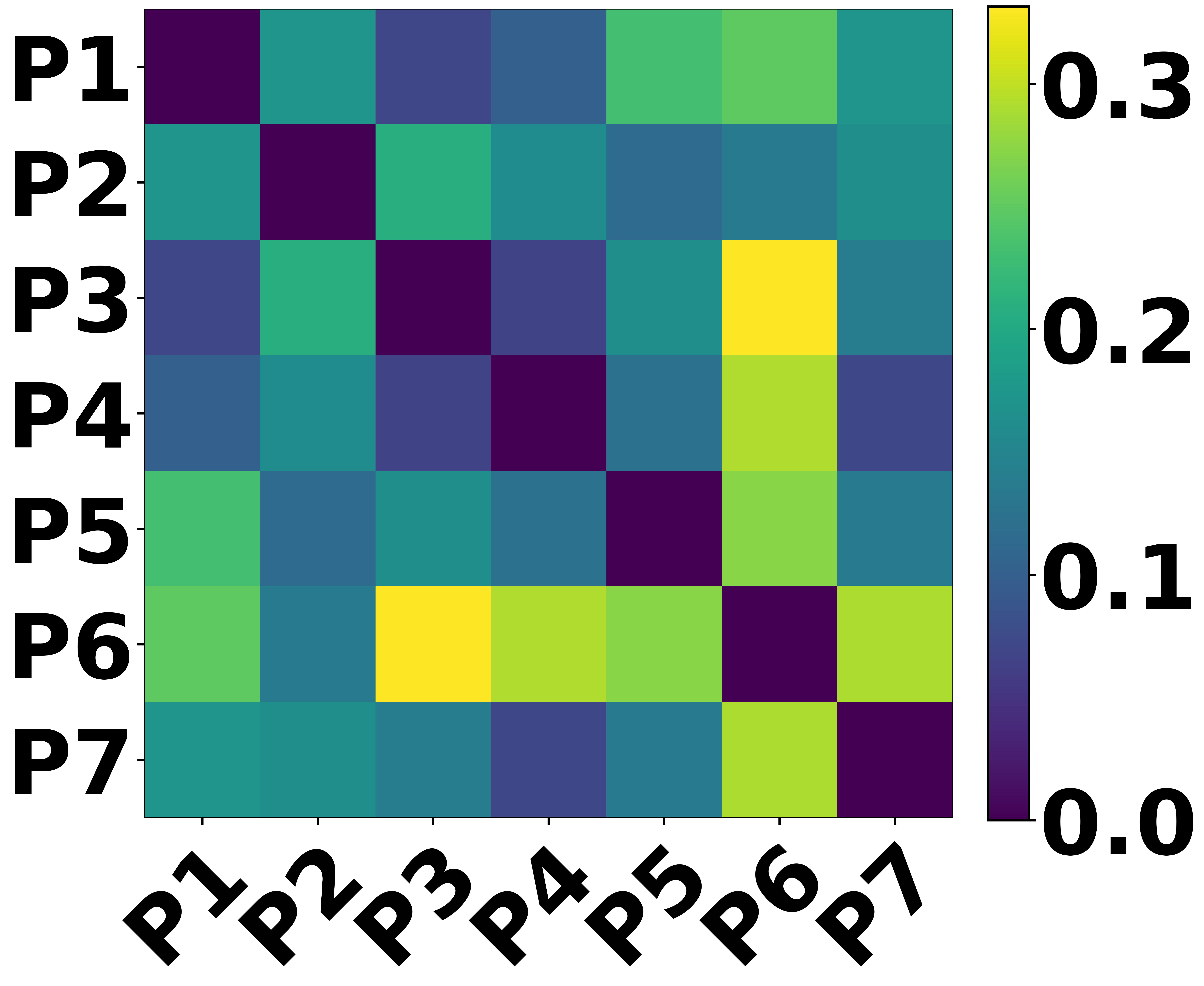}%
  \label{fig:hm_panel}%
}

\caption{\re{Inter-participant cosine-distance heatmaps for seven activities (Drive, Relax, Nod, Drink, Phone, Smoke, and Panel)}.}
\label{fig:heatmap_7behaviors}
\vspace{-10pt}
\end{figure*}

\re{To further verify that the collected signals are not trivially similar across participants even within the same behavior, we quantify inter-participant variability using a prototype-based distance analysis. 
For each behavior and each participant, we construct a participant-level prototype by averaging all available samples after applying sample-wise normalization to mitigate amplitude/offset differences. 
We then compute pairwise dissimilarity between participant ($\mathbf{p}$) prototypes using cosine distance, $d_{i,j}=1-\cos(\mathbf{p}_i,\mathbf{p}_j)$, which emphasizes similarity in spatiotemporal patterns rather than absolute magnitude. 
As shown in Fig.~\ref{fig:heatmap_7behaviors}, the resulting inter-participant distance heatmaps (one per behavior) exhibit diverse off-diagonal values across participant pairs, indicating noticeable participant-dependent variations and supporting that the ALERT dataset is not composed of near-duplicate signals. 
This complements the t-SNE feature visualization by providing a direct, quantitative view of cross-participant variability at the signal level.
}

\subsection{Evaluations of ISA-ViT}

\begin{figure*}[t]
\centering
    \subfigure[Overall accuracy]{
    \includegraphics[width=0.45\textwidth, height=4cm]{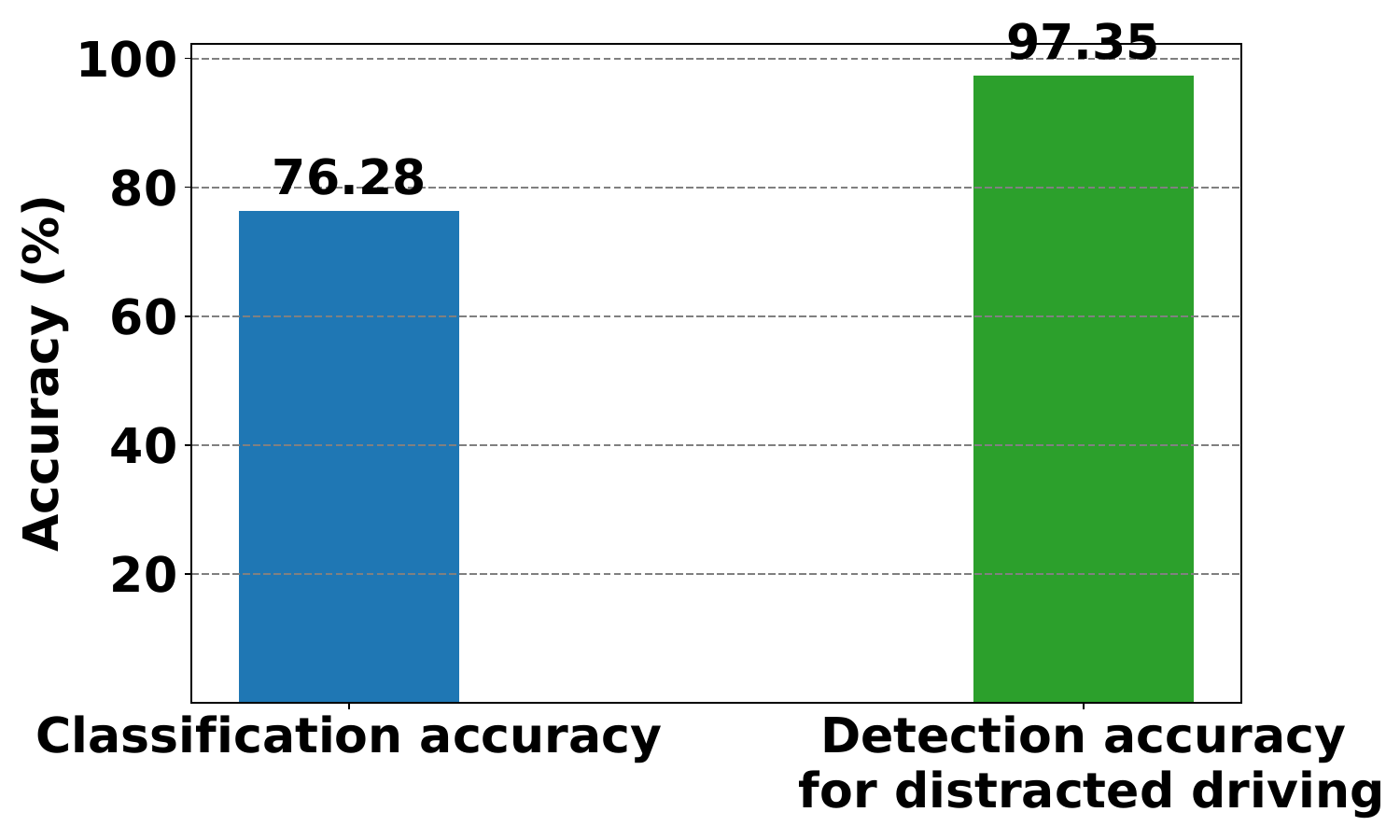}
    \label{fig:accuracy}
    }
    \subfigure[Confusion matrix of proposed scheme]{
    \includegraphics[width=0.45\textwidth, height=5cm]{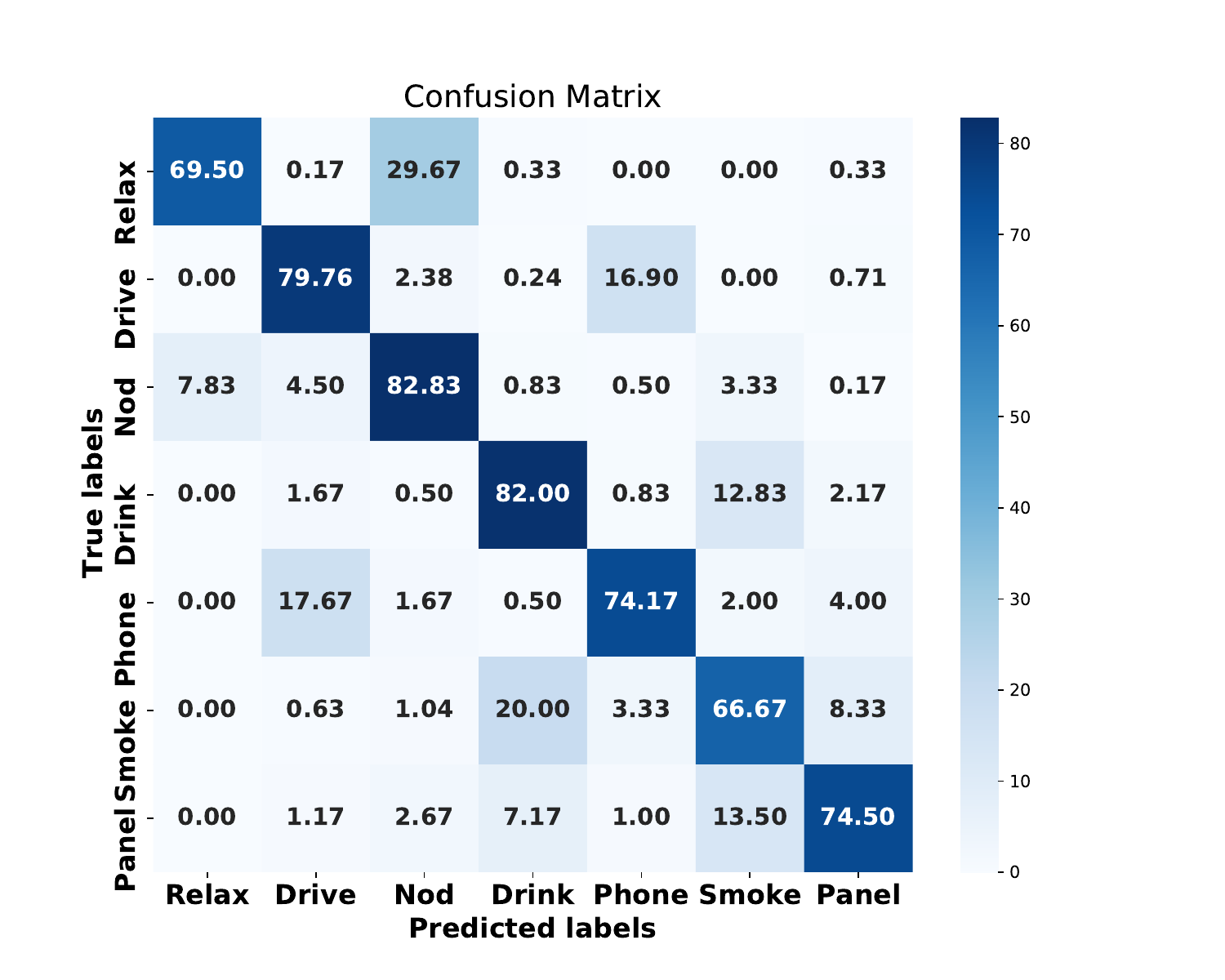}
    \label{fig:cm}
    }
    \caption{Experimental results for \mine performance.}
    \vspace{-10pt}
\label{fig:overall}
\end{figure*}

\subsubsection{\textbf{Overall performance}}
We evaluate the overall performance of DAR using UWB data. Our proposed model, \mine, is designed to handle various sizes of UWB data, making it adaptable to different data configurations. Furthermore, we incorporate a domain fusion algorithm to enhance the overall performance of DAR, enabling our system to leverage features from multiple domains for better classification results. The fusion of range-time and frequency-time domain features allows the model to capture more intricate patterns of driver activities, which are often missed by single-domain approaches. This multi-domain synergy significantly improves the accuracy and robustness of the system across diverse driving scenarios.

Fig.~\ref{fig:overall} shows the overall performance of the proposed scheme; Fig.~\ref{fig:accuracy} displays the average accuracy. We analyze accuracy from two perspectives; one considers both service-related aspects and safety driving, while the other focuses solely on accuracy from an extreme safety driving oriented perspective.

We refer to the classification accuracy (76.28\%), F1 score in Table~\ref{table2}, and confusion matrix in Fig.~\ref{fig:cm} to analyze the model’s accuracy from both safety and service perspectives. The classification accuracy reflects the model's ability to categorize all activities within our proposed framework. The precision and recall of each activity indicate the safety and service factors. If the recall for \emph{Drive} is low, the system may trigger false alarms, potentially interrupting the driving experience and reducing overall service quality. While false alarms may be frustrating from a service perspective, they do not pose significant safety risks. However, the precision for \emph{Drive} is directly linked to safety. Low precision could indicate that other potentially dangerous activities, such as using a smartphone, smoking, nodding are incorrectly classified as driving. Thus, we should refer to classification accuracy and F1 score together to analyze the accuracy in perspectives of safety and service.
 
\rev{Given that safe driving is more critical than general classification performance, \textbf{high accuracy in detecting distracted driving is essential}. In our case, the system should ensure that no activity other than \emph{Drive} is allowed while the vehicle is in motion. This requirement is crucial for maintaining safety, as identifying activities such as using a smartphone or smoking while driving is vital to preventing distractions that could lead to accidents. To this end, we focus on accuracy specifically related to the detection of distracted activities. While service-related aspects were not considered, achieving an accuracy of 97.35\% demonstrates a substantial contribution to enhancing road safety.}

\begin{table}[t]
    \centering
    \caption{Precision, recall, and F1 score for each label}
    \vspace{-10pt}
    \begin{tabular}{|p{1cm}|p{0.6cm}|p{0.6cm}|p{0.6cm}|p{0.6cm}|p{0.6cm}|p{0.6cm}|p{0.6cm}|}
        \hline
        & \emph{Drive} & Relax & Nod & Drink & Phone & Smoke & Panel \\ \hline
        Precision & \textbf{68.51} & 89.87 & 69.12 & 76.64 & 81.50 & 62.75 & 84.34 \\ \hline
        Recall    & \textbf{79.76} & 69.50 & 82.83 & 82.00 & 74.17 & 66.67 & 74.50 \\ \hline
        F1 score  & \textbf{73.71} & 78.38 & 75.36 & 79.23 & 77.66 & 64.65 & 79.12 \\ \hline
    \end{tabular}
    \label{table2}
    \vspace{-10pt}
\end{table}
\subsubsection{\textbf{Impact on resizing method}}
\begin{figure}[t]
\centering
        \subfigure[Using \emph{ALERT} dataset]{
        \includegraphics[width=0.45\textwidth, height=4cm]{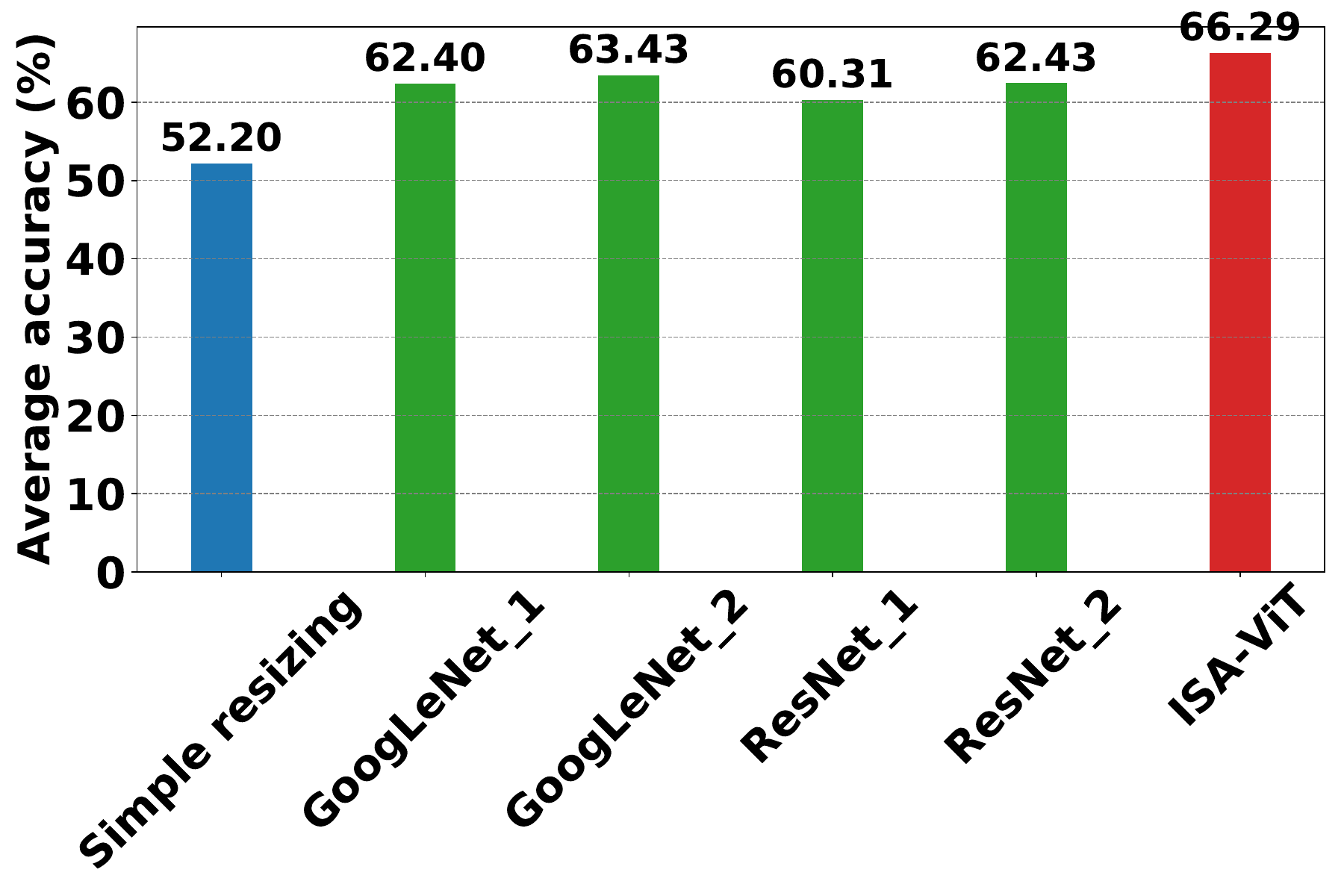}
        \label{fig:resizing_ALERT}
        }
        \hfill
        \subfigure[Using \emph{RaDA} dataset]{
        \includegraphics[width=0.45\textwidth, height=4cm]{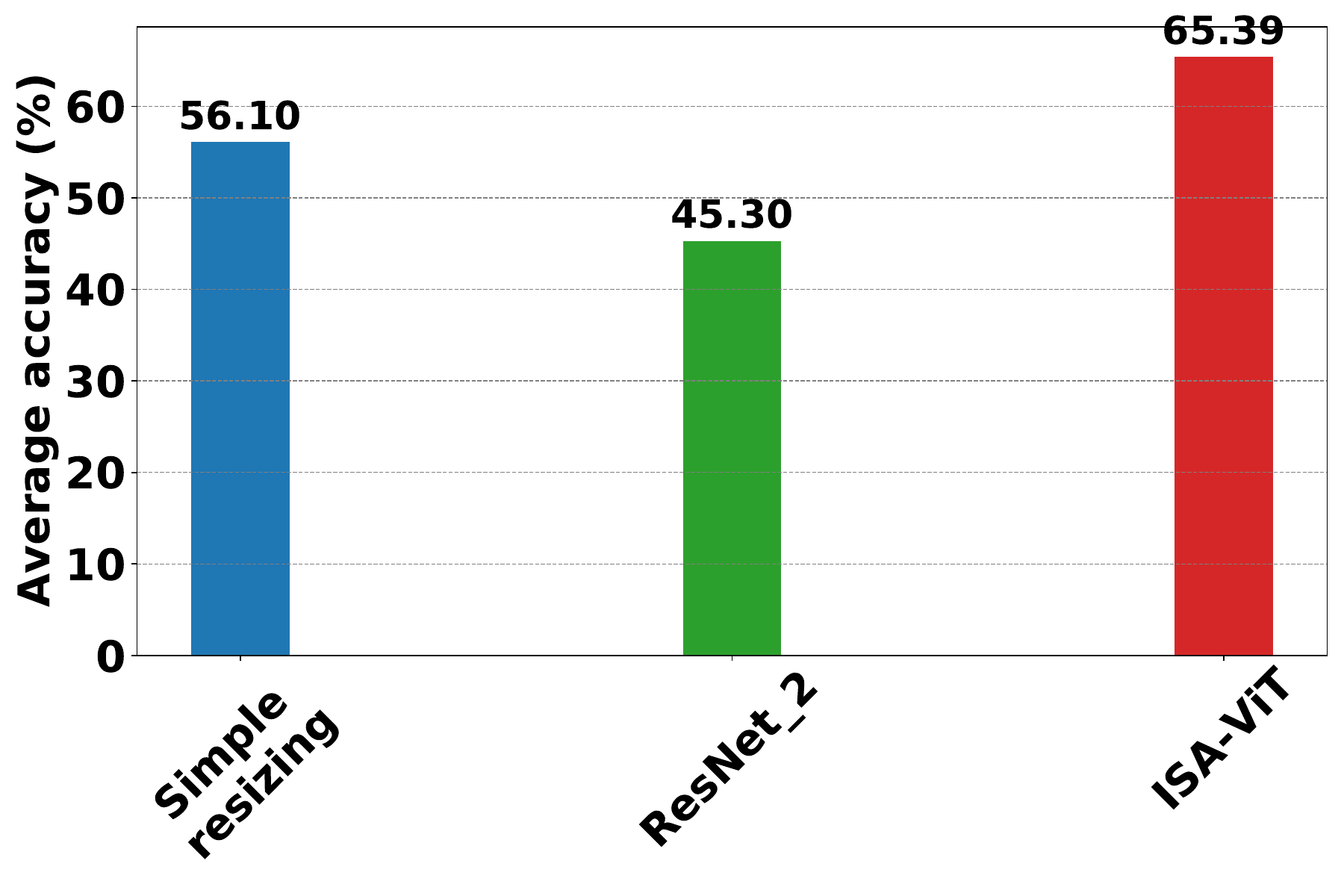}
        \label{fig:resizing_RaDA}
        }
    \caption{\rev{Accuracy comparison for resizing methods: simple resizing (using up- and down-sampling), GoogLeNet\_x (using xth inception modules and up- and down-sampling), ResNet\_x (using xth blocks and up- and down-sampling).}}
    \label{fig:resizing}
    \vspace{-10pt}
\end{figure}

\re{In the following \mine-specific evaluations (Figs.~\ref{fig:resizing}--\ref{fig:CNN_comp}, we further analyze the contributions of information-preserving resizing and complementary domain fusion under fair comparison settings.}
To evaluate \mine, we conduct experiments with several resizing methods for the \emph{ALERT} and \emph{RaDA} datasets. We have already verified that maintaining the 14$\times$14 sequence of pre-trained PEVs yields better performance than manipulating them in Section~\ref{sec:ISA-ViT}. Therefore, we only compare methods that maintain pre-trained PEVs.

\rev{Fig.~\ref{fig:resizing} compares the average accuracy of different resizing methods. The blue bar represents the simple resizing method, which applies basic up- and down-sampling to reshape the data to 224$\times$224. The green bars show CNN-based methods, which first use early CNN layers to extract local features from UWB data, then apply up- and down-sampling to adjust the feature size to 224$\times$224. \emph{GoogLeNet\_1} and \emph{GoogLeNet\_2} use the first and second inception modules of GoogLeNet, while \emph{ResNet\_1} and \emph{ResNet\_2} use the first and second blocks of ResNet.}

For the \emph{ALERT} dataset, as shown in Fig.~\ref{fig:resizing_ALERT}, using CNN-based resizing approaches outperforms simple resizing, as they minimize information loss by using feature extraction instead of simple down-sampling. However, \mine retains the raw UWB data information without any loss, leading it to outperform the other methods.

For the \emph{RaDA} dataset, we use only \emph{ResNet\_2} in the CNN-based method since its input size is 1024$\times$24. When using only a single ResNet block, the data size becomes 512$\times$12, resulting in approximately half down-sampling, which increases the likelihood of information loss. Therefore, passing the data through two blocks and then resizing it to fit 224 is necessary to minimize information loss. Additionally, in the case of GoogLeNet, using the inception module leads to a significant reduction in size, requiring additional padding, making it unsuitable for our experiments.

As shown in Fig.~\ref{fig:resizing_RaDA}, simple resizing outperforms the CNN-based method in the \emph{RaDA} dataset. Due to the dataset's characteristics, the key information is centered within the sample, so when CNN-based methods are applied, the features cover only a small portion of the sample. Thus, simple resizing performs better than the CNN-based method. However, \mine outperforms all methods due to its ability to resize without information loss.

Through this ablation study, we demonstrate that, for CNN-based approaches, the number of layers the input passes through depends on the input size. However, \mine can be applied to any input size without such considerations, offering greater flexibility and adaptability.

\subsubsection{\textbf{Impact on domain fusion}}
\begin{figure}[t]
\centering
    \includegraphics[width=0.45\textwidth, height=4cm]{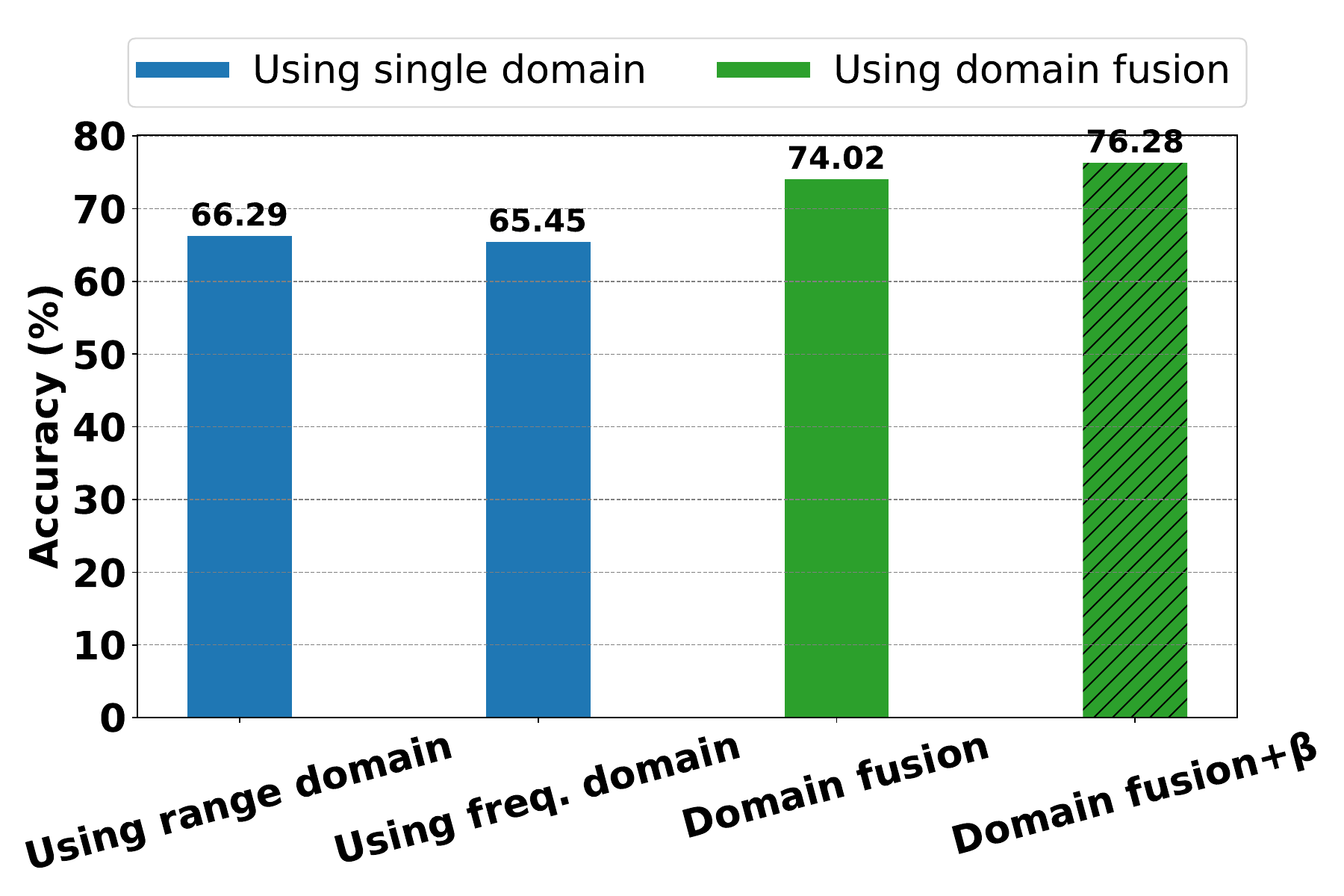}
    \caption{Accuracy results for domain fusion.}
    \label{fig:domain_fusion}
    \vspace{-10pt}
\end{figure}

\begin{table}[t]
    \centering
    \caption{Comparison of precision, recall, and F1 score before/after domain fusion}
    \begin{tabular}{|c|p{0.6cm}|p{0.6cm}|p{0.6cm}|p{0.6cm}|p{0.6cm}|p{0.6cm}|p{0.6cm}|}
        \hline
        \textbf{Before}& Relax & Drive & Nod & Drink & Phone & Smoke & Panel \\ \hline
        Precision & 87.56 & 57.19 & 58.73 & 61.64 & 69.24 & 76.54 & 78.69 \\ \hline
        Recall    & 56.33 & 85.24 & 75.67 & 77.67 & 70.17 & 51.67 & 56.00 \\ \hline
        F1 score  & 68.56 & 68.45 & 66.13 & 68.73 & 69.70 & 61.69 & 65.43 \\ \hline
        \hline
        \textbf{After}& Relax & \emph{Drive} & Nod & Drink & Phone & Smoke & Panel \\ \hline
        Precision & 89.87 & 68.51 & 69.12 & 76.64 & 81.50 & 62.75 & 84.34 \\ \hline
        Recall    & 69.50 & 79.76 & 82.83 & 82.00 & 74.17 & 66.67 & 74.50 \\ \hline
        F1 score  & \textbf{78.38} & \textbf{73.71} & \textbf{75.36} & \textbf{79.23} & \textbf{77.66} & \textbf{64.65} & \textbf{79.12} \\ \hline
    \end{tabular}
    \label{table3}
    \vspace{-10pt}
\end{table}

To enhance the performance of DAR using UWB, we jointly exploit the range domain and the frequency domain of UWB data. As shown in Fig.~\ref{fig:domain_fusion}, using only the range domain and the frequency domain achieves accuracy of 66.29\% and 65.45\%, respectively. The range domain performs slightly better than the frequency domain in DAR. However, since each domain classifies data based on its unique characteristics, they often provide different predictions for the same input. Therefore, combining and using the feature information from both domains jointly can lead to improved results. Consequently, the domain fusion algorithm achieves an accuracy of 76.28\%. 

Moreover, Fig.~\ref{fig:domain_fusion} illustrates the accuracy impact of the trained adjusting factor $\beta$ for frequency features. Before training the adjusting factor $\beta$, we achieve a performance of 74.02\% with range and frequency features at an equal ratio. However, after jointly training $\beta$ and applying it to domain fusion, we observe an accuracy of 76.28\%. This confirms that it is crucial not merely to combine range and frequency features, but to balance them appropriately for optimal fusion.
\re{This effect is also reflected at the activity level (Table.~\ref{table3}), where domain fusion improves the F1 score across all activities (e.g., Panel: +13.69, Drink: +10.50, Relax: +9.82, Nod: +9.23), indicating that complementary domain information is particularly helpful for distinguishing visually similar behaviors.}

The detailed performance enhancements are presented in Table~\ref{table3}, which shows the precision, recall, and F1 score for each activity before and after applying domain fusion. For all activities, the F1 scores improve after domain fusion is applied. These improvements demonstrate that by combining information from different domains, the model can more accurately distinguish similar activities that were previously easily confused. While precision and recall tendencies of some activities may vary, the overall improvement in F1 scores indicates a well-balanced trade-off between precision and recall, resulting in strong performance. The enhanced F1 scores for each activity after domain fusion confirm the model's improved ability to recognize a variety of activities more accurately. Through these experiments, we verify that using the domain fusion algorithm helps resolve conflicts between labels and enables more precise classification.

\subsubsection{\textbf{Comparison with CNN models}}
\begin{figure}[t]
    \subfigure[Using single domain]{
    \includegraphics[width=0.45\textwidth, height=3.5cm]{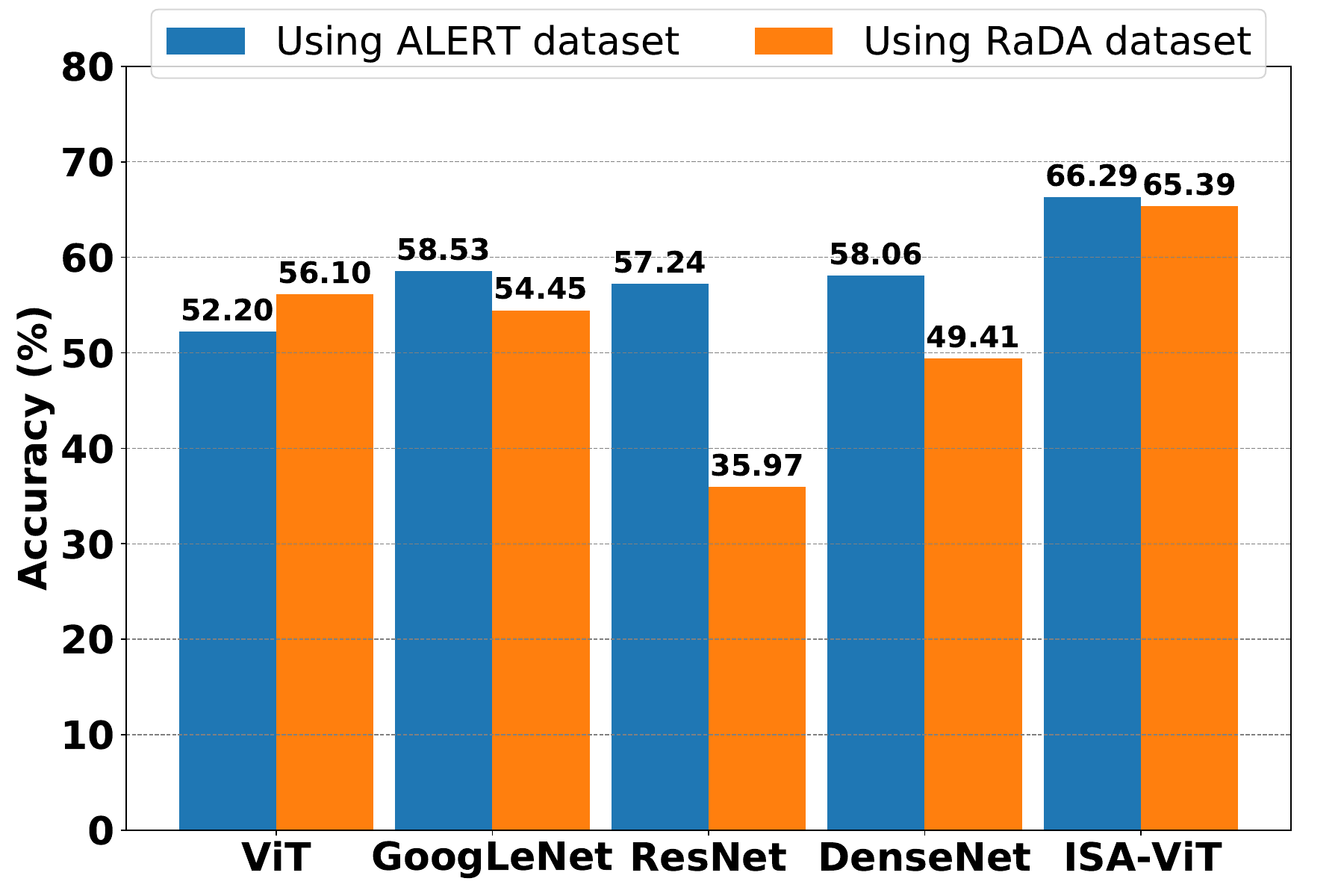}
    \label{fig:CNN_comparison}
    }
    \hfill
    \centering
    \subfigure[Using domain fusion]{
    \includegraphics[width=0.45\textwidth, height=3.5cm]{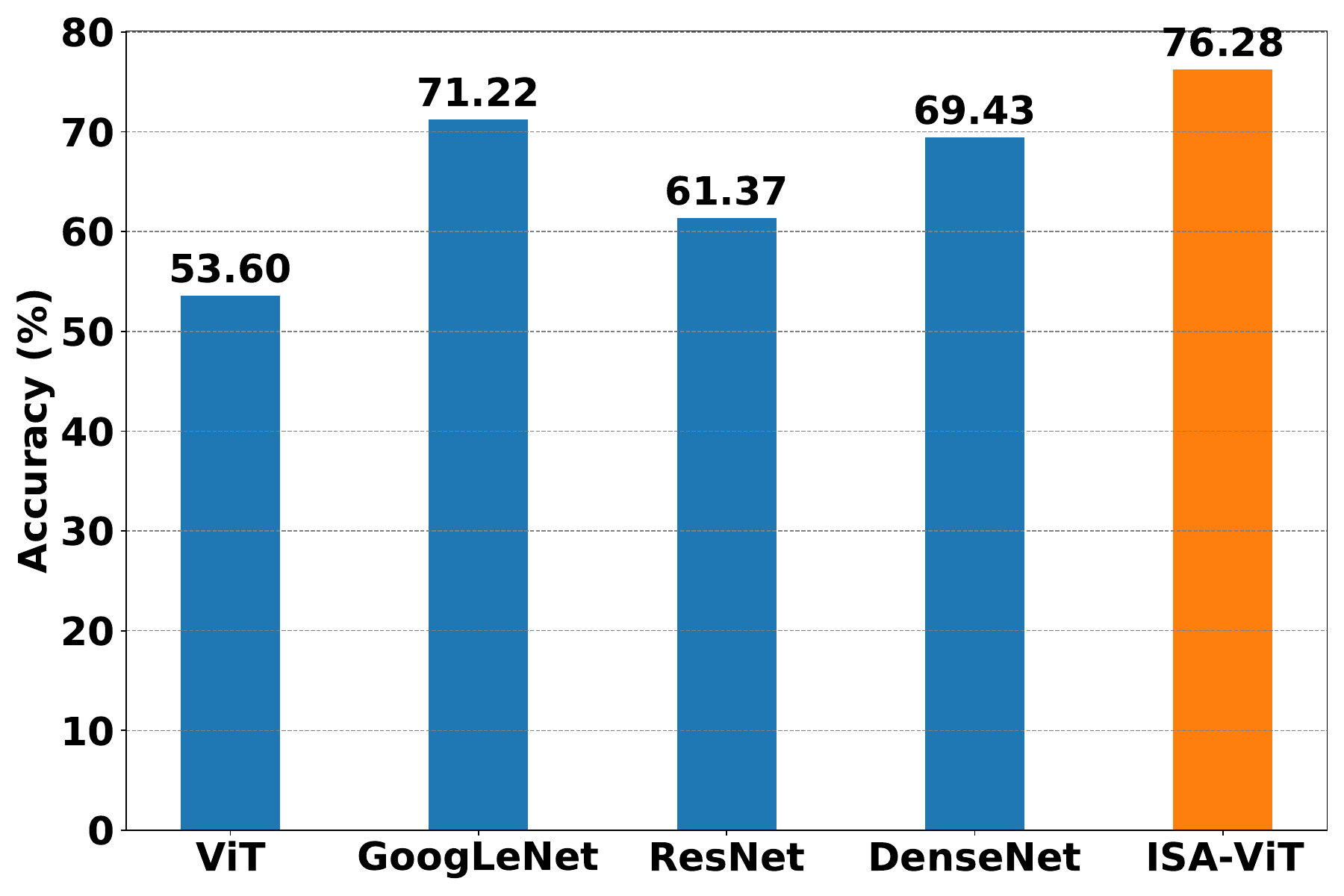}
    \label{fig:CNN_domain}
    }
    \caption{Accuracy comparison with CNN models.}
    \label{fig:CNN_comp}
    \vspace{-10pt}
\end{figure}
\begin{table*}[t!]
\caption{FLOPs of learning algorithms}
\centering  
    \begin{tabular}{|c|c|c|c|c|c|c|c|c|}  
    \hline  
    Algorithms & GoogLeNet & ResNet & DenseNet & MobileNet & RNN & DeiT & ViT & \mine\\ \hline
    FLOPs & 3.82~G&6.43~G&3.82~G&0.35~G&7.53~G&1.08~G&59.70~G&60.01~G\\ \hline
    \end{tabular}
    \label{flops}
    \vspace{-10pt}
\end{table*}
\rev{We compare the performance of ViT and CNN models with \mine. Although standard ViT is a state-of-the-art model for vision tasks, directly applying raw UWB data without resizing leads to poor recognition accuracy, as mentioned in the Section~\ref{sec:preliminary}. To enable fair comparison, we apply the simple resizing method to the UWB data before inputting it into ViT and compare its performance with CNN models and \mine.}

\rev{In the \ALERT dataset, as shown by the blue bars in Fig.~\ref{fig:CNN_comparison}, CNN models outperform ViT. This is because the simple resizing approach causes significant information loss, depending on the input size. For the \ALERT dataset, nearly half of the data information is lost during resizing. In contrast, \mine achieves higher performance than CNNs by allowing ViT to process data without information loss, regardless of input size.}

\rev{However, the \emph{RaDA} dataset, shown by the orange bars in Fig.~\ref{fig:CNN_comparison}, displays a different trend. This dataset has a size of 1024×24, forming a tall and narrow shape with most key features located near the center. In this case, simple resizing, which increases the horizontal dimension, helps ViT better capture structural patterns, leading to better performance than CNN-based methods\footnote{The better performance of CNN-based methods is due to the use of simple resizing for ViT. When we apply standard ViT to raw UWB data without resizing, the model fails to train properly on both the ALERT and RaDA datasets, resulting in less than 20\% accuracy.}. Nonetheless, \mine achieves the best performance overall, as it resizes data without information loss.}

\rev{These results show that performance varies depending on the data characteristics and model architecture. Still, \mine consistently provides the best performance across both datasets by preserving data information during resizing and effectively leveraging pre-trained PEVs.}

\rev{For the comparison with domain fusion of CNN models as shown in Fig.~\ref{fig:CNN_domain}, we only use the \emph{ALERT} dataset since \emph{RaDA} provides only the single domain data. We apply the same CNN models in parallel as feature extractors, passing both range and frequency domain data through each model. We then concatenate the features from each model and pass the concatenated features to the classifier. Here, we use the adjusting factor $\beta$ for the frequency domain, consistent with the domain fusion method of \mine. 
Therefore, we also train $\beta$ to achieve the best accuracy in the domain fusion of all CNN models. In a fair comparison, using the best performance of each method, our proposed scheme outperforms the CNN models.}
\re{Taken together, these studies provide component-level justification for \mine by showing that information-preserving resizing with preserved pre-trained PEVs and complementary domain fusion with a learned $\beta$ are key to robust performance across UWB input shapes.}

\section{Discussion}\label{sec:discussion}
\subsection{Points of Attention for \ALERT Dataset}
\re{The \emph{ALERT} dataset is the first UWB-based dataset that comprehensively captures distracted driving activities in a real-driving environment, and we expect it to support future DAR research. Nevertheless, we acknowledge several limitations.}

\re{First, \ALERT was collected using a single vehicle (a mid-sized sedan), as acquiring additional vehicle types (e.g., SUVs, compact cars, and large vehicles) was logistically challenging. To reduce potential bias from cabin geometry, we fixed the sensor mounting configuration and analyzed a driver-centric ROI by cropping range bins around the driver, which mitigates vehicle-dependent reflections. Moreover, as shown in Sec.~\ref{sec:evaluation}, long-delayed multipath components that often carry environmental information tend to act as noise for most algorithms, suggesting that region of interest~(ROI) cropping can further suppress environment/vehicle variations. While multi-vehicle experiments are left for future work, the radar-to-driver setup was kept consistent and the participant cohort still provides meaningful variability, so we expect the main findings to remain valid.}

\re{Second, only four of the nine participants collected data for all labels. This was mainly due to the smoking label: to avoid unrealistic samples, we excluded non-smokers mimicking smoking, and thus not all participants could contribute to that class. Nonetheless, each label was collected from a majority of participants, which we consider sufficient for model training and evaluation.}

\subsection{Implications and Further Explorations of \mine}
\re{\mine achieves substantially higher performance than other models with only a small additional cost over standard ViT, which is a key advantage. Built on the ViT architecture to transfer its representational power to UWB domains, \mine inevitably requires more computation than CNN/RNN baselines. While this trade-off is acceptable given the accuracy gain, low-latency response remains important in DAR scenarios.}

\re{Table~\ref{flops} summarizes the computational cost of each model. \mine improves accuracy by 22.68\% over ViT with only a 0.3~GFLOPs increase; however, real-time inference may still be challenging on resource-limited edge devices.}

\re{In future work, we will reduce computation via pruning, quantization, and knowledge distillation, and explore efficient transformer variants tailored for UWB data to retain high accuracy with lower cost for edge deployment.}

\section{Conclusion}\label{sec:conclusion}
\re{This study addresses distracted driving---a major cause of traffic accidents---by proposing an IR-UWB-radar-based Driver Activity Recognition (DAR) framework. IR-UWB mitigates key limitations of conventional approaches, such as privacy concerns and susceptibility to noise and interference. However, its adoption has been hindered by (i) the lack of comprehensive datasets collected under real-driving conditions and (ii) the difficulty of adapting state-of-the-art models, such as Vision Transformers (ViTs), to non-image UWB data.}

\re{To overcome these challenges, we develop the \ALERT dataset, comprising 10{,}220 samples of seven distracted driving activities recorded in real-driving environments. The dataset provides both range- and frequency-domain representations, enabling robust evaluations and realistic benchmarking. We also propose \mine, which supports variable-size UWB inputs by resizing the data without loss and adapting pre-trained positional embeddings. Furthermore, domain fusion improves recognition by combining complementary features from the range and frequency domains.}

\re{Extensive experiments investigate dataset-parameter effects across algorithms and validate the proposed framework. ISA-ViT achieves 76.28\% classification accuracy and 97.35\% distracted-driving detection accuracy, demonstrating the promise of UWB radar and transformer-based modeling for DAR.}

\re{By publicly releasing \ALERT with comprehensive benchmarks, this work establishes a foundation for future DAR research and aims to support safer driving by reducing distracted driving incidents.}

\bibliographystyle{IEEEtran}
\bibliography{reference}

\newpage
\vspace{-7ex}
\begin{IEEEbiography}[{\includegraphics[width=1in,height=1.25in,clip,keepaspectratio]{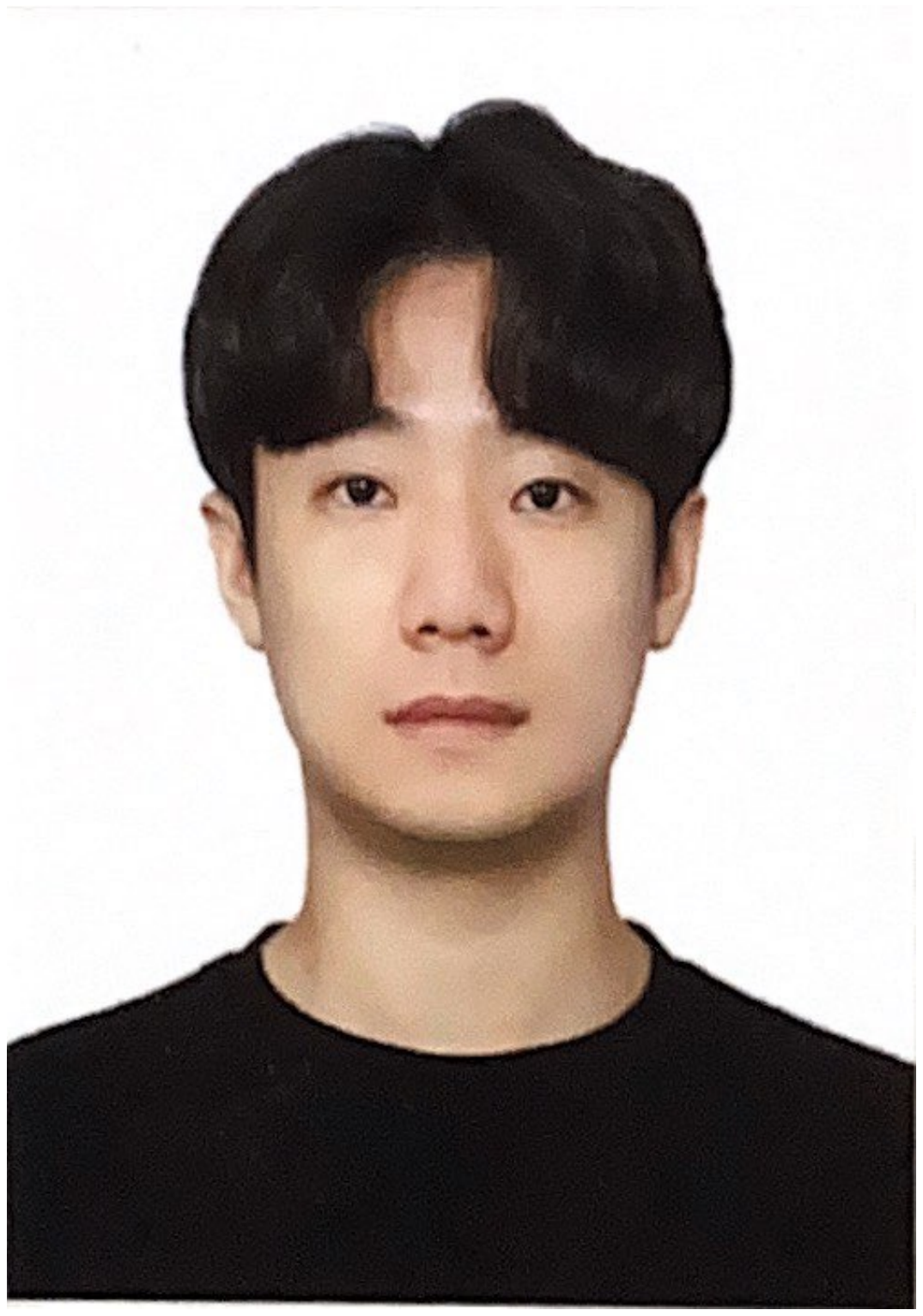}}]{Jeongjun Park}
received the B.S. degree from the Department of Information Communications Engineering, Hankuk University of Foreign Studies (HUFS), in 2018, and the M.S. and Ph.D. degree in Electrical and Computer Engineering from Seoul National University (SNU), Seoul, South Korea, in 2020 and 2025, respectively. He is currently with System LSI Business in Samsung Electronics. His research interests include the 5G/6G communication technologies, IoT, UWB, and AI.
\end{IEEEbiography}

\vspace{-5ex}
  
\begin{IEEEbiography}[{\includegraphics[width=1in,height=1.25in,clip,keepaspectratio]{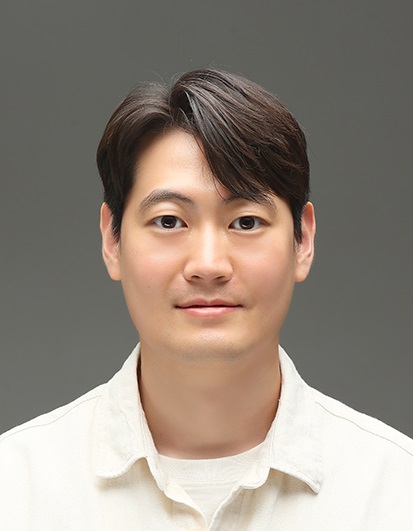}}]{Sunwook Hwang}
received the B.S. degree in Electrical Engineering from Pohang University of Science and Technology (POSTECH) in 2016 and the Ph.D. in Electrical and Computer Engineering from Seoul National University (SNU), Seoul, Korea, in August 2023. He served as a Post-Doctoral Researcher at SNU from September 2023 to August 2024.
He is currently with Samsung Electronics, where his work focuses on Neural Processing Unit (NPU) design. His research interests include autonomous driving and computer vision.
\end{IEEEbiography}

\begin{IEEEbiography}[{\includegraphics[width=1in,height=1.25in,clip,keepaspectratio]{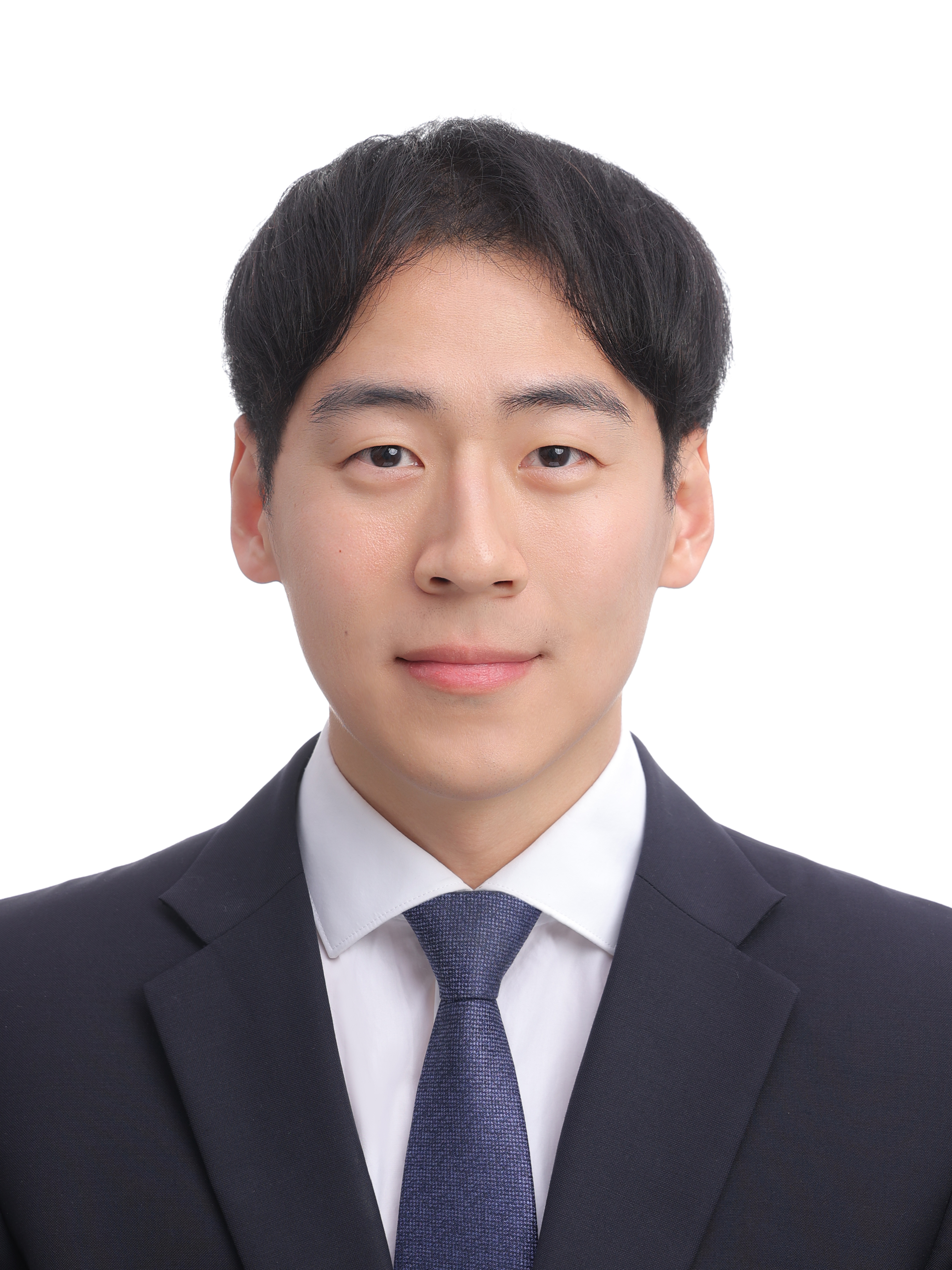}}]{Hyeonho Noh}
received the B.S. and the M.S. degrees in electrical engineering from Ulsan National Institute of Science and Technology (UNIST), Ulsan, South Korea, in 2018 and 2021, respectively, and received the Ph.D. degree in electrical engineering with the Pohang University of Science and Technology (POSTECH), Pohang, South Korea, in 2024. From Oct. 2024 to Jun. 2025, he worked as a post-doctoral researcher in the Department of Electrical Engineering, Seoul National University (SNU), Seoul, South Korea. He is currently an Assistant Professor with the Department of Information and Communication Engineering, Hanbat National University, Daejeon, South Korea. His research interests include integrated sensing and communication and future wireless systems with deep neural networks.
\end{IEEEbiography}

\begin{IEEEbiography}[{\includegraphics[width=1in,height=1.25in,clip,keepaspectratio]{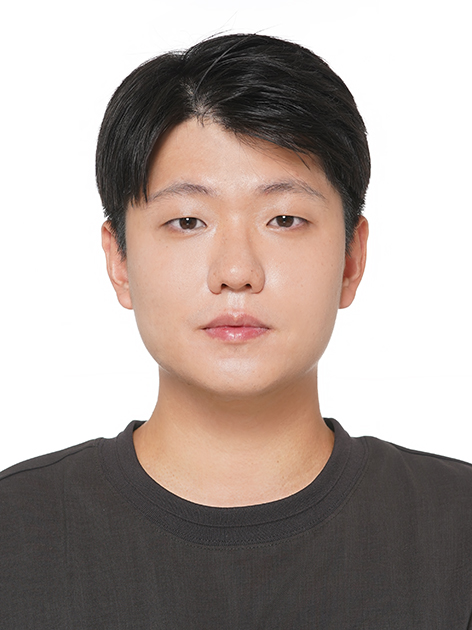}}]{Jin Mo Yang}
received the B.S. degree in electronic engineering from Hanyang University in 2018. He is currently pursuing the Ph.D. degree in electrical and computer engineering with Seoul National University. His research interests include video analytics, edge computing, and test-time adaptation (TTA) for evolving vision domains.
\end{IEEEbiography}

\begin{IEEEbiography}[{\includegraphics[width=1in,height=1.25in,clip,keepaspectratio]{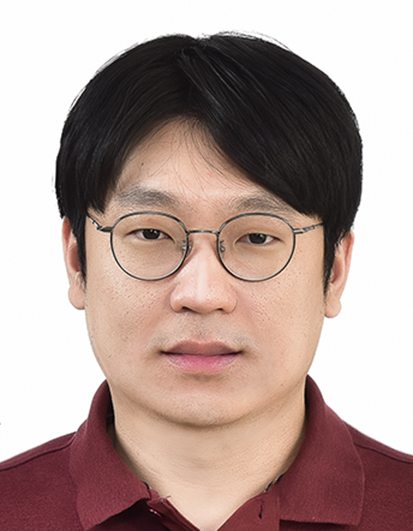}}]{Hyun Jong Yang}
is an associate professor in the department of electrical and computer engineering, Seoul National University (SNU), Seoul, Korea. He received the B.S. degree in electrical engineering from Korea Advanced Institute of Science and Technology (KAIST), Korea, in 2004, and the M.S. and Ph.D. degrees in electrical engineering from KAIST, in 2006 and 2010, respectively. From Aug. 2010 to Aug. 2011, he was a research fellow at Korea Research Institute of Ships \& Ocean Engineering (KRISO), Korea. From Oct. 2011 to Oct. 2012, he worked as a post-doctoral researcher in the Electrical Engineering Department, Stanford University, Stanford, CA. From Oct. 2012 to Aug. 2013, he was a Staff II Systems Design Engineer, Broadcom Corporation, Sunnyvale, CA, where he developed physical-layer algorithms for LTE-A MIMO receivers. In addition, he was a delegate of Broadcom in 3GPP RAN1 standard meetings. From Sept. 2013 to July 2020, he was an assistant/associate professor in the School of Electrical and Computer Engineering, UNIST, Korea. From July 2020 to Aug. 2024, he was an associate professor in the department of electrical engineering, Pohang University of Science and Technology (POSTECH), Pohang, Korea. Since Sept. 2024, he has been an associate professor in the department of electrical and computer engineering, Seoul National University (SNU), Seoul, Korea. His fields of interests are signal processing, wireless communications, and machine learning. Dr. Yang is an Editor for IEEE Wireless Communications Letters (WCL) and IEEE Internet of Things Journal (IoTJ), an Associate Editor-in-Chief for the Journal of Korean Institute of Communications and Information Sciences, and a Program Track Chair for the IEEE Consumer Communications \& Networking Conference (CCNC) 2023~2025. 
\end{IEEEbiography}

\vspace{-10ex}

\begin{IEEEbiography}[{\includegraphics[width=1in,height=1.25in,clip,keepaspectratio]{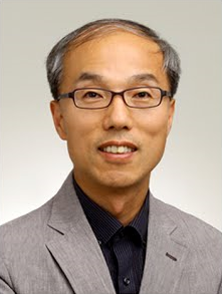}}]{Saewoong Bahk}
(Senior Member, IEEE) received the B.S. and M.S. degrees in electrical engineering from Seoul National University (SNU), in 1984 and 1986, respectively, and the Ph.D. degree from the University of Pennsylvania, in 1991. He was with AT\&T Bell Laboratories as a Member of Technical Staff, from 1991 to 1994, where he had worked on network management. From 2009 to 2011, he served as the Director of the Institute of New Media and Communications. He is currently a Professor at SNU. He has been leading many industrial projects on 3G through 6G and IoT connectivity supported by the Korean industry. He has published more than 300 technical articles and holds more than 100 patents. He was a recipient of the KICS Haedong Scholar Award, in 2012. He was President of the Korean Institute of Communications and Information Sciences (KICS). He served as Chief Information Officer (CIO) of SNU. He was General Chair of the IEEE WCNC 2020 (Wireless Communication and Networking Conference), IEEE ICCE 2020 (International Conference on Communications and Electronics), and IEEE DySPAN 2018 (Dynamic Spectrum Access and Networks). He was TPC Chair of the IEEE VTC-Spring 2014, and General Chair of JCCI 2015. He was Director of the Asia–Pacific Region of the IEEE ComSoc. He is an Editor of the IEEE Network Magazine and IEEE Transactions on Vehicular Technology. He was Co-Editor-in-Chief of the Journal of Communications and Networks (JCN), and an editor of Computer Networks Journal and IEEE Tran. on Wireless Communications. He is a member of the National Academy of Engineering of Korea (NAEK).
\end{IEEEbiography}

\end{document}